%% file: iclr2027_conference.tex
\documentclass{article} % For LaTeX2e

\PassOptionsToPackage{dvipsnames,table}{xcolor}
\usepackage{iclr2027_conference,times}

\setcitestyle{numbers,square,comma}

\usepackage[utf8]{inputenc} % allow utf-8 input
\usepackage[T1]{fontenc}    % use 8-bit T1 fonts
\usepackage{hyperref}       % hyperlinks
\usepackage{url}            % simple URL typesetting
\usepackage{booktabs}       % professional-quality tables
\usepackage{amsfonts}       % blackboard math symbols
\usepackage{amsmath}
\usepackage{amssymb}
\usepackage{nicefrac}       % compact symbols for 1/2, etc.
\usepackage{microtype}      % microtypography
\usepackage{xcolor}    % options passed above, before iclr2027_conference
\usepackage{graphicx}
\usepackage{wrapfig}
\usepackage{enumitem}
\usepackage[ruled,vlined]{algorithm2e}
\usepackage{cleveref}
\usepackage{caption}
\usepackage{tabularx}
\newcolumntype{Y}{>{\centering\arraybackslash}X}
\usepackage{multirow}
\usepackage{amsthm}
\usepackage{thmtools}
\usepackage{thm-restate}
\usepackage{paracol}
\usepackage{multicol}
\usepackage{mdframed}
\usepackage{bbm}

\columnratio{0.40}
\theoremstyle{plain}

\newtheorem{lemma}{Lemma}
\newtheorem{proposition}{Proposition}
\theoremstyle{definition}

\theoremstyle{remark}
\newtheorem{remark}{Remark}
\providecommand{\addb}[1]{{#1}}

\newcommand{\methodname}{\texttt{S}eed\texttt{F}lood}

\hypersetup{
    colorlinks = true,
    linkcolor = red,
    anchorcolor = red,
    urlcolor = red,
    citecolor = Blue,
}

\title{\methodname{}: A Step Toward Scalable\\Decentralized Fine-Tuning of LLMs}

\author{Jihun Kim\textsuperscript{1}, Dongyeop Lee\textsuperscript{1}, Namhoon Lee\textsuperscript{1} \\
\textsuperscript{1}POSTECH \\
\texttt{\{jihun.kim,dongyeop.lee2,namhoon.lee\}@postech.ac.kr}
}

\iclrfinalcopy % Uncomment for camera-ready version, but NOT for submission.

\begin{document}

\maketitle
\lhead{Preprint}

\input{sections/0-Abstract}
\input{sections/1-Introduction}

\input{sections/2-Background}

\input{sections/4-Method}

\input{sections/4.5-Theory}
\input{sections/5-Experiments}

\input{sections/6-Conclusion}

%%%%%%%%%%%%%%%%%%%%%%%%%%%%%%%%%%%%%%%%%%%%%%%%%%%%%%%%%%%%
% ICLR 2027 required / recommended statements.
% These go at the end of the main text, before the references, and do NOT
% count toward the 9-page limit (each should not exceed 1 page).
%%%%%%%%%%%%%%%%%%%%%%%%%%%%%%%%%%%%%%%%%%%%%%%%%%%%%%%%%%%%

\subsection*{AI use statement}
% REQUIRED by ICLR 2027. See the ICLR 2027 AI Policy for Authors.
% TODO: replace the bracketed placeholders with the actual disclosure.
In this work, we used generative AI tools to assist with the proofs of our theoretical analysis, specifically to check the completed proofs for errors post hoc and to identify loose steps in the
convergence bounds; and to assist with translation.
We have not used generative AI tools to provide critical ingredients for proving mathematical claims, develop theoretical models or conceptual frameworks, formulate mathematical claims, propose or refine hypotheses, design or provide feedback on research methodology or experiments, implement the core framework of the method, or interpret results. Generating synthetic datasets, cleaning or reformatting datasets, and conducting qualitative or thematic data analysis were not applicable to this work.
Additionally, we used generative AI tools for editing software code (organizing variables, cleaning code, and
helping interpret code written by co-authors), but not for implementing the
method; for editing the paper to improve readability; for identifying relevant
literature; and for creating or modifying Figures 1 and 2. We have reviewed all AI-assisted work. 
All theorem statements were formulated by the authors, every
AI-assisted proof step, including the Gr\"onwall-based argument, was verified
line by line by the authors, all cited works were checked against the original
sources, and all code edited with the aid of generative AI was reviewed by the
authors.
We take responsibility for the final content of this work, including text, claims or
artifacts produced with the aid of generative AI.

\subsection*{Ethics statement}
% RECOMMENDED by ICLR 2027.
This work does not involve human subjects, the collection or release of personal or sensitive data, or deployment in safety-critical settings. We use publicly available models and benchmark datasets and are not aware of specific ethical concerns beyond those generally associated with large language model research.

\subsection*{Reproducibility statement}
We provide the SeedFlood algorithm in Section 3 and Appendix A. Experimental configurations, including datasets, hyperparameters, LoRA settings, iteration-budget protocols, and compute resources, are detailed in Section 5.1 and Appendix G, with implementation and runtime details in Appendix D and task-level numerical results in Appendix H. The assumptions and complete proofs for our theoretical results are provided in Appendices E and F. The source code will be made publicly available upon acceptance.
% RECOMMENDED by ICLR 2027. Reference the parts of the paper/appendix that
% support reproducibility rather than restating the details here.
% TODO: fill in, pointing at sections/appendix-expdetail and the released code.

\subsubsection*{Acknowledgments}
This work was partly supported by Institute of Information \& communications Technology Planning
\& Evaluation (IITP) grant (No.2019-0-01906, Artificial Intelligence Graduate School Program(POSTECH), No.RS-2026-25507427, Development of Efficient Architectures and Training Techniques for High-Performance Lightweight
AI Models), 
the National Research Foundation of Korea (NRF) grant (RS-2023-00210466, RS-2026-25612350), 
and the Korea Basic Science Institute(National research Facilities and Equipment Center) grant (RS-2026-25500419), all funded by the Korea government(MSIT).

% Do NOT include acknowledgments in the anonymized submission -- only in the
% camera-ready version.

%%%%%%%%%%%%%%%%%%%%%%%%%%%%%%%%%%%%%%%%%%%%%%%%%%%%%%%%%%%%

\bibliography{iclr2027_conference}
\bibliographystyle{iclr2027_conference}

%%%%%%%%%%%%%%%%%%%%%%%%%%%%%%%%%%%%%%%%%%%%%%%%%%%%%%%%%%%%

\newpage
\tableofcontents
\newpage
\appendix
\input{sections/appendix-Algorithm}
\input{sections/appendix-additional_experiment}
\input{sections/appendix-Discussion}
\input{sections/appendix-Runtime_analysis}
\input{sections/appendix-theory}
\input{sections/appendix-expdetail}
\input{sections/appendix-SupplementaryResults}

\end{document}

%% file: sections/0-abstract.tex
\begin{abstract}
This work presents \methodname{}, a new approach to decentralized LLM fine-tuning designed to scale across large models, large collaborations, and complex network topologies while achieving global consensus with negligible communication overhead.
Traditional methods suffer from high communication costs that grow with model size, while information decay over network hops renders global consensus inefficient.
\methodname{} takes a significant departure from these practices by exploiting the \underline{seed}-reconstructible structure of zeroth-order gradients and effectively making the messages to transmit near-zero in size, allowing them to be \underline{flood}ed to every client in the network, and thereby enhancing scalability of decentralized training.
% Our Theoretical result formalizes that \methodname{} matches centralized aggregation of zeroth-order updates, which removes topology-dependent consensus terms from the bound.
Consequently, \methodname{} enables training in regimes previously considered impractical, such as billion-parameter scale models or distributed across hundred of clients.
Our experiments on decentralized LLM fine-tuning demonstrate that \methodname{} consistently outperforms the standard zeroth-order baselines in both communication efficiency and generalization performance, and even achieves results comparable to first-order gossip-based methods in large-scale settings, while requiring orders-of-magnitude less communication cost.
% Despite the intrinsic limitations of zeroth-order estimation, \methodname{} can match or surpass first-order gossip baselines at larger collaboration scales by avoiding consensus errors, while requiring orders-of-magnitude less communication.
We also provide theoretical analysis to formalize that \methodname{} avoids topology-dependent consensus terms in the convergence bound while retaining the acceleration enabled by increased client participation.

% Also, \methodname{} does not suffer from performance degradation as the network scales; rather, performance improves with more participating clients, as supported both theoretically and empirically.
\end{abstract}

%% file: sections/1-Introduction.tex
\section{Introduction}
\label{sec:intro}

Decentralized collaborative fine-tuning has recently emerged as a promising approach for adapting large language models under limited local compute and memory resources, enabling participants to collectively build a single model beyond what they could obtain individually \citep{JustOneByte,FedKseed,DecomFL,LoRA-Gossip,wang2025adf}. 
To further advance this paradigm, recent efforts have begun to exploit the low intrinsic dimensionality of fine-tuning through techniques such as LoRA \citep{lora} and zeroth-order optimization \citep{MeZO,LOZO}, and to explore various combinations that balance communication and optimization costs to identify configurations that better support scalable decentralized collaboration \citep{LoRA-Gossip,wang2025adf}, demonstrating promising communication--performance trade-offs.
% To further advance this paradigm, recent efforts have begun to exploit the low intrinsic dimensionality of fine-tuning through techniques such as LoRA \citep{lora} and zeroth-order optimization \citep{MeZO,LOZO}, reducing communication costs in large-scale decentralized collaboration \citep{LoRA-Gossip,wang2025adf} and demonstrating promising communication--performance trade-offs.

However, while reduced communication loads have broadened the scalability of decentralized training, comparatively little attention has been paid to the deeper interplay beyond simply combining existing components: how optimization structure should shape communication design, and how communication strategies should, in turn, reshape optimization.
This gap is particularly consequential given the longstanding limitations imposed by conventional treatments of communication constraints when designing decentralized optimization methods that scale to large and complex collaborations \citep{choco-sgd,vogels2021relaysum,consensus_control}.
Moreover, just as advances in fine-tuning have broadened our understanding of what can be achieved through more efficient optimization, progress in decentralized fine-tuning may similarly reveal new principles for how decentralized training itself can be designed and optimized.

Motivated by this, we develop \methodname{} around a simple yet powerful recent idea: zeroth-order gradient estimates can be represented using only two scalars---a seed that reconstructs the random perturbation and a scaling coefficient---reducing high-dimensional gradient information to a nearly negligible payload independent of model scale \citep{JustOneByte,FedKseed,DecomFL}.
Realizing this as a consistent end-to-end protocol calls for a new decentralized training framework that naturally accommodates this representation, since forcing it into the usual gossip-averaging paradigm generally destroys the very simplicity that makes seed--scalar communication attractive in the first place.
% This opens new design opportunities about the communication protocol. %and large-scale update reconstruction.
Instead, we argue that the resulting compact payload makes flooding \citep{kshemkalyani2011distributed} a natural fit, while also enabling exact network-wide consensus by preserving individual contributions intact rather than diluting them through successive gossip averages \citep{Unified_theory_gossip}, a major source of slow consensus at large collaboration scales in traditional decentralized frameworks.
% This opens new design opportunities about the communication protocol. %and large-scale update reconstruction.
% Precisely, the resulting compact payload makes flooding \citep{kshemkalyani2011distributed} a viable communication and consensus mechanism, achieving exact consensus throughout the network without having their individual contributions diluted through successive gossip averages \citep{Unified_theory_gossip}.

Additionally, for the overall framework to scale effectively, local computation must remain balanced with the now-reduced communication burden: reconstructing high-dimensional perturbations from all other participants causes the per-node workload to grow linearly with the number of clients, eventually creating a severe computational bottleneck at large collaboration scales, particularly on modern accelerators.
% However, naive flooding creates a severe computational bottleneck: each node must reconstruct high-dimensional perturbation vectors from every other participant, which grows linearly with network size, causing substantial local computations at large collaboration scale.
To resolve this, \methodname{} couples flooding with a novel aggregation-aware perturbation structure, constraining optimization to a periodically refreshed shared low-rank subspace whose size is independent of collaboration scale, while remaining sufficiently expressive in practice and allowing each client to aggregate updates directly in the compressed space before reconstructing them in the original parameter space.

Together, these designs enable \methodname{} to scale with both model size and collaboration scale, addressing three distinct challenges: model-size-dependent communication overhead (\Cref{sec:shared-randomness}), consensus degradation in larger decentralized collaborations (\Cref{sec:Flooding}), and the increasing local computation required to process flooded updates (\Cref{sec:aggregation}).
% Empirically, \methodname{} delivers favorable efficiency--performance trade-offs, consistently outperforming conventional framework designs---including a wide range of combinations spanning different communication--optimization trade-offs---in both communication efficiency and generalization performance across large-scale networks.
Empirically, \methodname{} delivers favorable efficiency--performance trade-offs and, at larger collaboration scales, consistently outperforms conventional framework designs---including diverse communication--optimization trade-offs---in both communication efficiency and generalization performance.
Furthermore, our theoretical analysis establishes convergence bounds without topology-dependent consensus terms, while retaining the convergence acceleration enabled by increased client participation.

We highlight our contributions in the following:
\begin{itemize}[leftmargin=2ex]
    \item \underline{\methodname{} framework}: A new decentralized training foundation that achieves topology-invariant consensus with minimal, constant-size, model-independent communication overhead.
    \item \underline{Scalable Aggregator}: A scalable gradient aggregator that operates within a shared low-rank subspace, enabling an extreme speedup in aggregating large volumes of flooded updates. 
    \item \underline{Theoretical guarantees}:
    Topology-independent convergence bounds that retain the benefits of increased client participation through exact gradient aggregation.
    \item \underline{Empirical validation}: Extensive evaluations scaling up to billion-parameter LLMs and networks of 128 clients, demonstrating a strong performance-cost trade-off against first-order gossip baselines, particularly as collaboration scale increases.
    \end{itemize}

%% file: sections/2-Background.tex
\section{Background}
\label{sec:background}
\subsection{Decentralized training}

Decentralized training aims to learn a common model from resources and data distributed across clients that communicate only with connected neighbors, without directly sharing private raw data \citep{lian2017can,DFedAvg}.
Formally, clients indexed by $\mathcal{V}=\{1,\ldots,n\}$ communicate over a connected, undirected, static network $\mathcal{G}=(\mathcal{V},\mathcal{E})$ and update local parameters $\theta_i\in\mathbb{R}^d$ to solve
\begin{equation}
\min_{\theta_1,\ldots,\theta_n} \, \sum_{i\in\mathcal{V}} f_i(\theta_i)
\quad \text{s.t.} \quad
\theta_i=\theta_j, \;\; \forall (i,j)\in\mathcal{E},
\end{equation}
where $f_i:\mathbb{R}^d\to\mathbb{R}$ is the local objective of client $i$. The core challenge is therefore to jointly design what information to exchange, how to exchange it, and how to perform local computation so that the local models rapidly reach consensus on a model that performs well across all clients as the collaboration scale $n$ grows.

The existing paradigm has largely settled on a combination of gradient compression and gossip averaging \citep{gabrielli2023survey}: techniques such as quantization \citep{QSGD}, sparsification \citep{sparseSGD}, and low-rank compression \citep{powerGossip}, often coupled with error-feedback or communication-aware mechanisms \citep{choco-sgd,DeepSqueeze,BEER,MoTEF,AdaGossip}, reduce each round's payload.
% However, due to still large payloads, typical decentralized implementations employ a partial gossip averaging over the model at each step, \emph{e.g.}, $\theta_i \leftarrow \frac{1}{|\mathcal{N}(i)|}\sum_{j\in\mathcal{N}(i)} \theta_j$, so that the whole process occurs without communication propagating information through repeated exchanges of fixed-size messages while avoiding the need to forward an increasingly large collection of raw messages \citep{lian2017can,Unified_theory_gossip,DFedAvg,koneputugodage2026factored}.
% Meanwhile, gossip averaging for decentralized learning typically interleave local gradient updates with neighbor-wise model averaging, \emph{e.g.}, $\theta_i \leftarrow \frac{1}{|\mathcal{N}(i)|}\sum_{j\in\mathcal{N}(i)} \theta_j$, 
% propagating information through repeated exchanges of fixed-size messages while avoiding the need to forward an increasingly large collection of raw messages \citep{lian2017can,Unified_theory_gossip,DFedAvg,koneputugodage2026factored}.
This information is usually exchanged via gossip averaging \citep{lian2017can,Unified_theory_gossip,DFedAvg,koneputugodage2026factored}, which recursively mixes neighboring updates into a single fixed-size message, \emph{e.g.}, $m_i \leftarrow \frac{1}{|\mathcal{N}(i)|}\sum_{j\in\mathcal{N}(i)} m_j$, allowing information to propagate across the network without forwarding an ever-growing collection of raw payloads.
Although this substantially reduces communication, it still offers limited relief from payloads that scale with model dimensionality as model size grows; hence, the usual practice is to perform only a few rounds of gossip mixing over parameters $\theta$ per local optimization step, trading communication cost against the degree of consensus attained for global optimization.

However, local updates continually introduce model disagreement, while limited gossip mixing leaves residual consensus errors that can impair optimization \citep{consensus_control}.
This becomes particularly problematic in poorly connected or large network graphs, where achieving consensus requires more communication \citep{vogels2021relaysum,Unified_theory_gossip,consensus_control}.
At LLM scale, exchanging high-dimensional models remains costly, while compression can further hinder consensus by introducing additional errors \citep{lian2017can,choco-sgd,powerGossip}.
Thus, despite recent progress in collaborative LLM training \citep{Petals,DeDLOC,SwarmP}, robust network-wide consensus at massive scale remains difficult to achieve without prohibitive communication \citep{JustOneByte,LoRA-Gossip,wang2025adf}.

\subsection{Zeroth-order optimization}

Zeroth-order optimization provides a gradient-free alternative to backpropagation and has shown strong effectiveness for modern language model fine-tuning \citep{MeZO,LOZO}.
It estimates descent directions using only function evaluations \citep{Spall_ZO}, typically through a symmetric two-point estimator along a Gaussian perturbation $z\sim\mathcal{N}(0,I_d)$:
\begin{equation}
    g = \alpha z =
    \frac{
        f(\theta + \epsilon z)
        -
        f(\theta - \epsilon z)
    }{2\epsilon} z,
    \label{eqn:nspsa}
\end{equation}
where $\epsilon>0$ is the perturbation scale, with averaging over multiple perturbations progressively improving the gradient estimate \citep{ghadimi2013stochastic,nesterov2017random}.
Prior work has substituted such zeroth-order estimates for local gradients within standard decentralized templates \citep{hajinezhad2019zone,tang2020distributed,yi2022zeroth,hu2025stability}, but consequently inherits the consensus limitations of their first-order counterparts. 
Moreover, zeroth-order decentralized fine-tuning of LLMs over arbitrary network topologies remains largely unexplored.

Taken together, this motivates us to ask whether decentralized fine-tuning can instead achieve global consensus without making collaboration scale, network topology, or model dimensionality the limiting factor.

%% file: sections/4-Method.tex
\section{Method}

% We propose a novel method that supports exact global consensus while substantially alleviating the dependence on model dimensionality and collaboration scale in decentralized fine-tuning.

We propose \methodname{} through three components that address communication overhead, consensus degradation, and the increasing local computation. The complete algorithm is provided in Appendix~\ref{app:algorithm}.
% Specifically, we exploit the fact that random perturbations for zeroth-order estimation can be fully determined by a single seed, and design around this property to propagate each client’s estimate intact across the network with minimal communication and computation overhead, as described below.

\label{sec:method}
% \subsection{Shared randomness as a communication primitive}
% \subsection{Exchanging zeroth-order estimations via two scalars}
\subsection{Seed-scalar based dimension-free communication}
\label{sec:shared-randomness}
Naively, communicating a zeroth-order estimate requires transmitting either the scalar directional derivative $\alpha$ together with the $d$-dimensional perturbation $z$, or simply the resulting $d$-dimensional estimate $\alpha z$.
In practice, however, $z$ is typically generated deterministically from a compact seed $s$ using a pseudorandom number generator $\texttt{PRNG}:\mathbb{R}\to\mathbb{R}^d$:
\begin{equation*}
    z = \texttt{PRNG}(s) \sim \mathcal{N}(0,I_d),
\end{equation*}
a standard approach for generating sufficiently diverse perturbations in computational statistics and machine learning \citep{MeZO,FedKseed}. Consequently, any client using the same \texttt{PRNG} can exactly regenerate another client's perturbation from its seed:
\begin{equation*}
    \forall z_i := \texttt{PRNG}_i(s),\qquad
    \texttt{PRNG}_j=\texttt{PRNG}_i
    \;\Longrightarrow\;
    \texttt{PRNG}_j(s)=z_i.
\end{equation*}
Thus, once the \texttt{PRNG} is shared in advance, the full zeroth-order estimate can be reconstructed from only the seed--scalar pair $(s,\alpha)$, effectively removing model dimensionality as the dominant communication bottleneck in centralized federated learning \citep{FedKseed}.

%, an approach exploited in memory-efficient zeroth-order optimization \citep{malladi2023fine}.

% The corresponding scalar directional derivative is
% \begin{equation}
%     \alpha_{i,t}
%     =
%     \frac{
%         F(\theta^t + \epsilon z_{i,t};\, B_{i,t})
%         -
%         F(\theta^t - \epsilon z_{i,t};\, B_{i,t})
%     }{2\epsilon} .
% \end{equation}
% Each zeroth-order update is thus fully specified by the seed--scalar pair $m_{i,t} = (s_{i,t}, \alpha_{i,t})$: upon receiving $m_{i,t}$, any client can deterministically reconstruct $z_{i,t}$ and apply the update locally.
% Consequently, the size of each communicated update is independent of the model dimension~$d$, removing dimensionality as the communication bottleneck in decentralized training.

% \subsection{\methodname{}: Flooding based decentralized fine-tuning}
\subsection{Flooding for Scalable Global Consensus}

\label{sec:Flooding}
\begin{figure}
    \centering
    \includegraphics[width=1\linewidth]{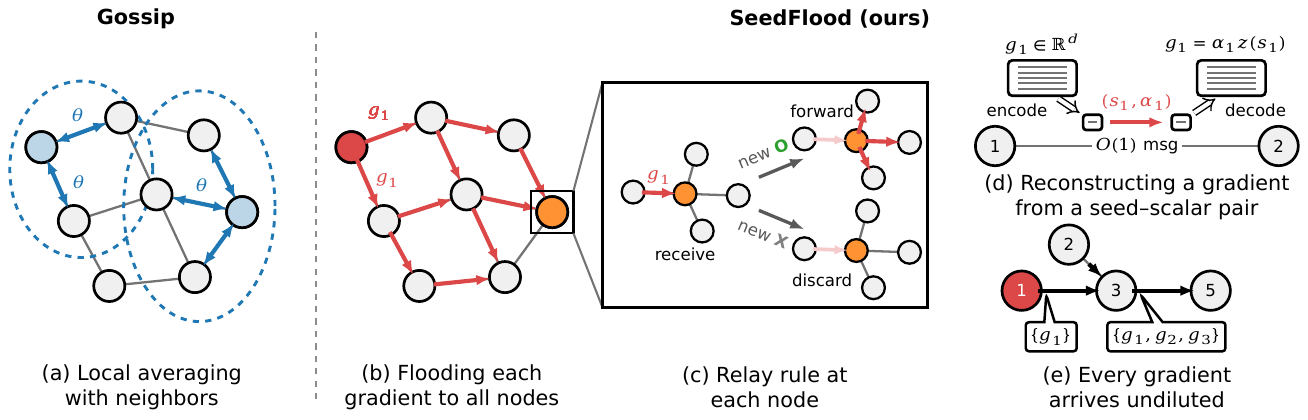}
\caption{Comparison of gossip averaging and \methodname{}.
(a) Gossip reaches consensus through repeated neighbor-wise model averaging.
(b) \methodname{} instead floods every client's gradient to the entire network, so that each client updates with the average of all of them.
(c) Flooding remains duplicate-free because each client forwards a message exactly once, upon its first reception.
(d) Each zeroth-order gradient can be communicated using only a random seed and a scalar coefficient.
(e) This lossless flooding enables rapid multi-hop propagation and exact consensus without the information dilution inherent in gossip averaging.}
\label{fig:floodingoverview}
\end{figure}

However, conventional gossip averaging is not well suited to seed--scalar communication for two reasons.
First, an arbitrary message---such as gossip averaged---models generally cannot be represented by a single seed--scalar pair without solving an impractical inverse-search problem over the chosen \texttt{PRNG}, \emph{i.e.},
$\arg\min_{s,\alpha} \| \alpha \cdot \texttt{PRNG}(s) - \frac{1}{|\mathcal{N}(i)|}\sum_{j\in\mathcal{N}(i)}\theta_j \|$.
Alternative approaches also compromise the simplicity or efficiency of seed--scalar communication (Appendix~\ref{app:seed_gossip}).
Second, even if such encoding were feasible, finite-round gossip mixing would still leave topology-dependent consensus errors that can degrade optimization, making its limitations difficult to justify when individual updates can already be communicated compactly \citep{vogels2021relaysum}.

We therefore propose adopting flooding \citep{kshemkalyani2011distributed} as the mechanism for seed-based communication in decentralized training, allowing compact seed--scalar representations to be disseminated intact across the network while naturally enabling exact global consensus. 
As illustrated in \Cref{fig:floodingoverview}b, each communication round consists of multiple \emph{flooding iterations}, in which clients relay messages received in the preceding iteration. 
Upon receiving a previously unseen gradient, a client applies the corresponding update and forwards it to each neighbor only once~(\Cref{fig:floodingoverview}c). 
Repeating this process for at least $D$ flooding iterations, where $D$ is the network diameter, ensures that every zeroth-order gradient $g_i$ reaches every client. 
Since each gradient is transmitted only as a seed--scalar pair~(\Cref{fig:floodingoverview}d), all clients receive the complete set of updates without the hop-wise dilution of gossip~(\Cref{fig:floodingoverview}e), thereby applying identical updates and reaching exact consensus with communication that remains independent of model dimensionality.

\subsection{Decoupling gradient aggregation from numbers of clients}
\label{sec:aggregation}

While \methodname{} enables exact global consensus, local computation can still become a scalability bottleneck as collaboration scale grows.

To elaborate, naively aggregating $n$ seed-encoded gradients into a model update requires regenerating and accumulating a dense perturbation $Z_i\in\mathbb{R}^{d_1\times d_2}$, with $d=d_1d_2$, for each message, incurring $O(nd)$ computation that couples number of clients $n$ with model dimensionality $d$.
Beyond this linear complexity, how well the computation maps to modern accelerators is also an important practical consideration that can become a dominant factor under the large-scale workloads typical of modern LLMs.
Here, the \texttt{axpy}-style accumulation of fully materialized perturbations is relatively low in arithmetic intensity and typically memory-bandwidth-bound compared to compute-dense matrix multiplication such as \texttt{GEMM}, adding an additional penalty on top of the linear $O(nd)$ scaling.
% Also, the raw computational complexity is not the only factor that may hinder practical scalability: its low arithmetic intensity can itself limit throughput on modern accelerators, becoming particularly relevant at the model and network scales that \methodname{} is intended to target.
% Precisely, fully materializing perturbations and streaming them through memory for scale-and-add operations, \emph{i.e.}, \texttt{axpy}, yields a representative memory-bandwidth-bound workload that makes less effective use of modern GPU compute throughput than compute-dense matrix multiplication such as \texttt{GEMM}, thereby imposing an additional practical scaling penalty beyond the already linear $O(nd)$ complexity.

% Although a classical coordinate-wise gradient estimation (CGE)~\citep{kiefer1952stochastic,DeepZero} would partially eliviate this by reducing the total amount of entries, this does not change the overall arithmetic-intensity, while coordinate zeroth-order optimization are not that effectiuve than dense ones in practice.

% To these reduce aggregation overhead, we design our perturbation scheme to satisfy two requirements: \textbf{(R1)} applying a received update must not require fully materializing $n$ number of a $d$-dimensional random vector, and \textbf{(R2)} applying many updates must reduce to dense matrix arithmetic rather than repeated per-message vector updates. 

\begin{figure}[t]
    \centering
    \includegraphics[width=1\linewidth]{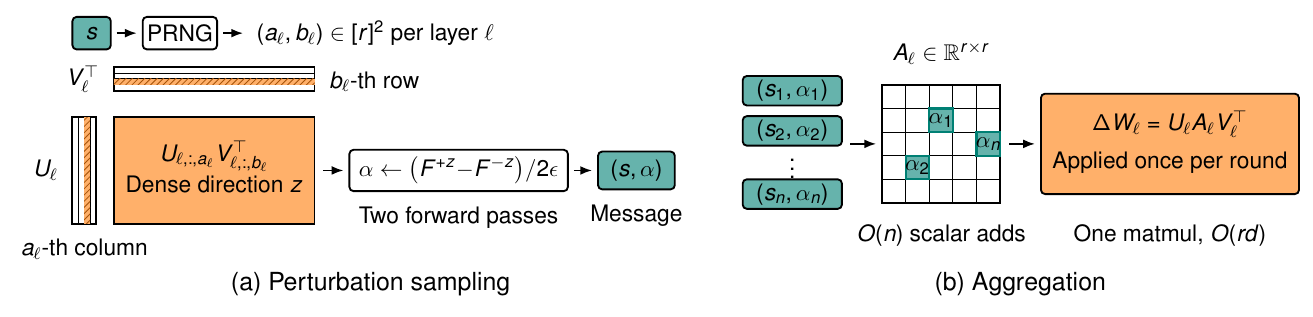}
    \caption{Subspace coordinate-wise perturbation of \methodname{}. (a)~Instead of drawing perturbations from Gaussian distribution, \methodname{} prepares shared bases $U_l,V_l$, samples a random coordinate pair $(a_l,b_l)$ at each iteration, which selects the dense rank-one direction $U_{\ell,:,a_\ell} V_{\ell,:,b_\ell}^{\top}$ as the perturbation. (b)~The $n$ received scalars are accumulated into $A_\ell$ and applied by a single matrix product.}
    \label{fig:subspace_estimator}
\end{figure}

To address both aspects, we propose a particular solution that restricts perturbations to a finite set of rank-one directions $\{\, u_a v_b = U \mathbbm{1}_{a,b} V^\top \mid (a, b) \in \mathbb{N}_{\leq r}^2 \,\}$ where each direction is specified by a pair of low-dimensional indices $(a,b)$, equivalently by a one-hot indexing matrix $\mathbbm{1}_{a,b}\in\mathbb{R}^{r\times r}$, over periodically refreshed, globally shared random matrices $U\in\mathbb{R}^{d_1\times r}$ and $V\in\mathbb{R}^{d_2\times r}$. 
Across $l$ layers, this yields ${r^2}$ candidate directions per layer, as illustrated in \Cref{fig:subspace_estimator}, and hence $r^{2l}$ possible model-wide perturbations $z$.

This construction does not fully materialize a $d$-dimensional perturbation for each message; instead, it first accumulates the $n$ received scalars into their corresponding entries of a coefficient matrix $A = \sum_{i=1}^{n}\alpha_{i}\mathbbm{1}_{a_i,b_i} \in \mathbb{R}^{r \times r}$, and only afterward materializes the aggregate update in parameter space through a single matrix product $U A V^\top$ (\Cref{fig:subspace_estimator}b):
\begin{equation*}
    \underbrace{\sum_{i=1}^{n} \alpha_i Z_i}_{\text{Full matrix aggregation: O($nd$)}}
    \qquad \Longrightarrow \qquad
    \sum_{i=1}^{n}
    \alpha_{i} U \mathbbm{1}_{a_i, b_i} V^\top
    =
    U
    \underbrace{
        \left(
        \sum_{i=1}^{n}
        \alpha_{i}\mathbbm{1}_{a_i,b_i}
        \right)
    }_{\text{Scalar aggregation: } O(n)}
    V^\top.
\end{equation*}
This reduces the aggregation cost from $O(nd)$ to $O(n+rd)$, decoupling collaboration scale from model dimensionality, while leaving a single dense matrix multiplication as the dominant operation---a workload substantially better suited to modern GPUs than repeated \texttt{axpy}-style updates.
In our profiling run, applying thousands of received messages is up to ${\sim}2000\times$ faster than naive reconstruction (\Cref{fig:MA_times}; implementation details in Appendix~\ref{appendix:Subcgeruntime}).

Two practical considerations remain.
First, the additional indices and shared matrices need not introduce substantial communication overhead: the $2l$ indices at each iteration and the periodically refreshed per-layer $U,V$, are likewise generated by a \texttt{PRNG} and can therefore be determined, communicated, and reconstructed from a single seed~(\Cref{fig:subspace_estimator}a).
Second, confining to $lr^2$ directions within a refresh period may restrict the optimization dynamics despite periodic refreshing.
We analyze this effect formally in \Cref{sec:theory}, including the roles of the subspace size $r$ and refresh period $\tau$.
Appendix~\ref{app:char_sub_est} empirically validates the runtime efficiency and task performance, including its match to unrestricted Gaussian estimation (\Cref{tab:subcge_mezo_single}).

%% file: sections/4.5-Theory.tex
\section{Theoretical Analysis}
\label{sec:theory}

In this section, we provide the convergence analysis of \methodname{}.
The convergence of the underlying subspace coordinate-wise estimator is established separately in Proposition~\ref{thm:lowrank-estimator} in Appendix~\ref{appendix:subspace_converge}.
Full proofs, together with the formal statements of all assumptions (Assumptions~\ref{assump:smoothness}--\ref{assump:client}), are provided in Appendices~\ref{appendix:subspace_converge} and~\ref{appendix:seedflood-proof}.

\paragraph{Setup.}
Consider $n$ clients with local objectives
$f_i(X):=\mathbb{E}_{\xi\sim\mathcal{D}_i}[F_i(X;\xi)]$
and global objective $f(X):=\frac{1}{n}\sum_{i=1}^n f_i(X)$,
where $X\in\mathbb{R}^{d_1\times d_2}$.
At period $k$, all clients share
$U_k\in\mathbb{R}^{d_1\times r_1}$ and
$V_k\in\mathbb{R}^{d_2\times r_2}$ for $\tau$ inner steps.
Let $X^{(k)}:=X_{k,0}=X_{k\tau}$, $B_{k,0}=0$, and $q:=r_1r_2$.
At inner step $s$, each client $i$ independently samples
$J_{i,k,s}\sim\mathrm{Unif}\{1,\ldots,q\}$,
sets $C_{i,k,s}=E_{J_{i,k,s}}$, and computes

\begin{equation}
    \begin{aligned}
a_{i,k,s}
&:=
\frac{
F_i(X^{(k)}+U_k(B_{k,s}+\epsilon C_{i,k,s})V_k^\top;\xi_{i,k,s})
-
F_i(X^{(k)}+U_k(B_{k,s}-\epsilon C_{i,k,s})V_k^\top;\xi_{i,k,s})
}{2\epsilon},\\
B_{k,s+1}
&=
B_{k,s}
-\frac{\eta}{n}\sum_{i=1}^n a_{i,k,s}C_{i,k,s},
\qquad
X_{k,s}
=
X^{(k)}+U_kB_{k,s}V_k^\top .
\end{aligned}
\label{eq:seedflood-update}
\end{equation}

Under exact flooding, decentralized dynamics coincide with centralized data-parallel zeroth-order optimization over the shared subspace.

Under Assumption~\ref{assump:client}, each local objective satisfies the same smoothness and noise conditions, with minibatch variance at most $\sigma^2/b$, and
$\frac{1}{n}\sum_i\|\nabla f_i(X)-\nabla f(X)\|_F^2\le\zeta^2$.
We further let
$T:=K\tau$,
$\widetilde{L}:=L\rho_U^2\rho_V^2$, and
$\widetilde{\zeta}^2:=\rho_U^2\rho_V^2\zeta^2$.

  \begin{restatable}[Convergence of \methodname{} under exact flooding]{theorem}{ThmSeedflood}
  \label{thm:seedflood}
  Under Assumptions~\ref{assump:subspace}, \ref{assump:coordinate}, and \ref{assump:client} with exact flooding, let $\Lambda_{q,n}(\tau):=\tau^2/q^2+\tau/(nq)$. If $\eta\le\bigl(\widetilde{L}{\sqrt{32\,\Lambda_{q,n}(\tau)}}\bigr)^{-1}$ and $c_2\eta\widetilde{L}(\tau/q+1/n)\le c_1/2$, then
  \[
      \frac{1}{K}\sum_{k=0}^{K-1}\mathbb{E}\|\nabla f(X_{k\tau})\|_F^2
      \;\le\;
      \frac{2\Delta_0}{c_1\eta T}
      + a_\sigma\widetilde{L}\eta{\frac{\sigma^2}{nb}}
      + a_\zeta\widetilde{L}\eta\frac{\widetilde{\zeta}^2}{qn}
      + a_\epsilon\frac{c_\delta^2}{q}\!\left[1+\widetilde{L}\eta\!\left(\frac{\tau}{q}+\frac{1}{n}\right)\right]\!\epsilon^2,
  \]
  where $\widetilde{\zeta}^2:=\rho_U^2\rho_V^2\zeta^2$, $c_\delta:=\sqrt{q}\,\widetilde{L}/2$, and $a_\sigma,a_\zeta,a_\epsilon>0$ are universal constants.
  \end{restatable}

The theorem highlights two key scaling properties of \methodname{}.
First, the bound contains no topology-dependent factor such as a spectral gap: exact flooding makes the decentralized dynamics identical to centralized data-parallel optimization.
Second, larger $n$ reduces both the stochastic-noise term $\sigma^2/(nb)$ and the heterogeneity term $\widetilde{\zeta}^2/(qn)$, while relaxing the stepsize condition through $\Lambda_{q,n}(\tau)$.
The remaining trade-off arises from subspace reuse: large $\tau$ relative to $q$ amplifies the finite-difference error, making the refresh period a practically important hyperparameter.

%% file: sections/5-Experiments.tex
\section{Experiments}
\label{sec:expermients}
Our experiments examine two consequences of \methodname{}'s design: whether dimension-free exact aggregation improves the communication--performance trade-off at LLM scale, and whether its advantage persists or grows with increasing client participation.

\subsection{Experimental setup}
\begin{figure*}[t]
\centering
\includegraphics[width=\textwidth]{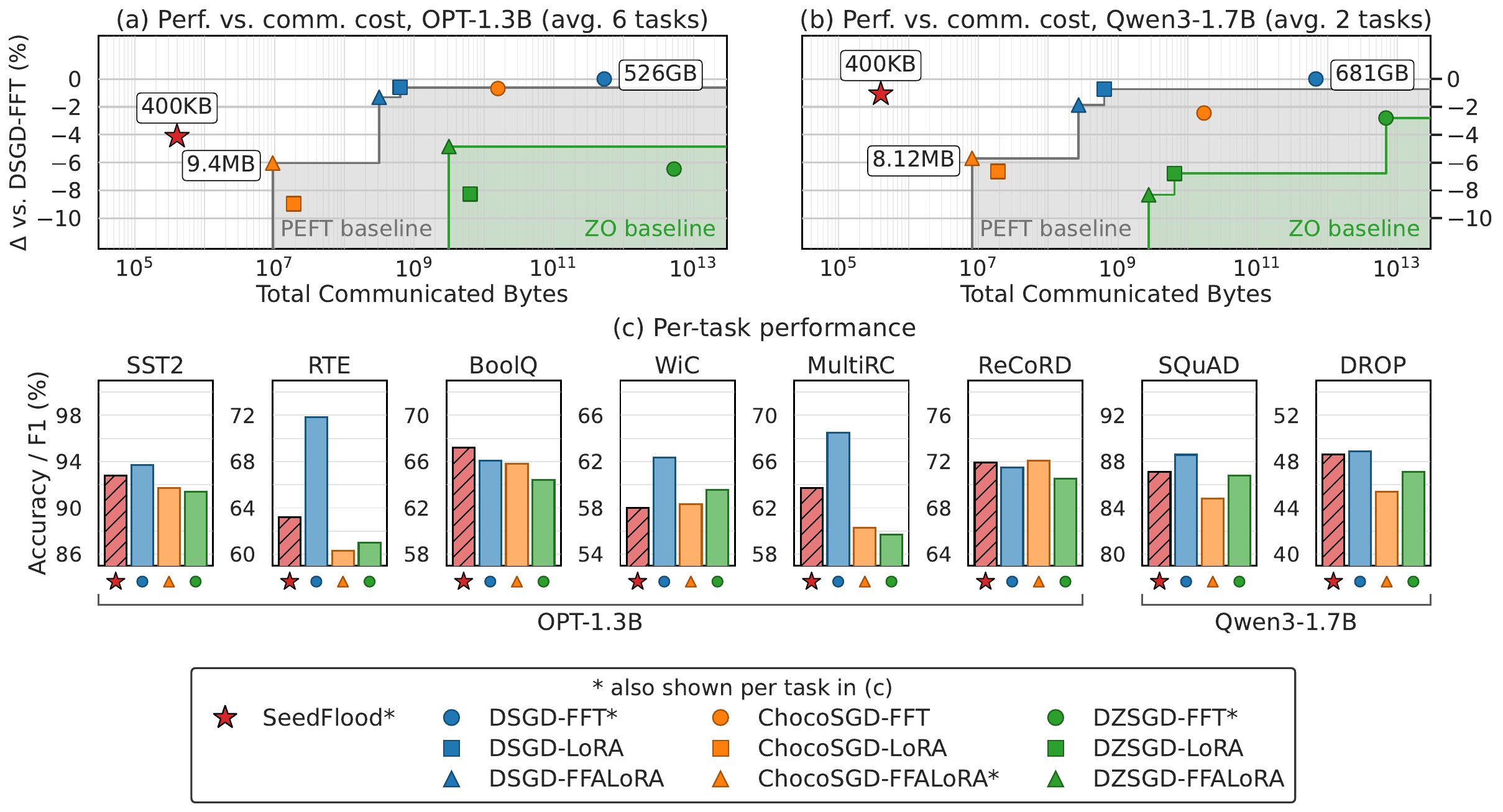}
\caption{Performance--communication trade-off on a 16-client ring network.
(a,b) \methodname{} versus 9 decentralized fine-tuning baselines on OPT-1.3B and Qwen3-1.7B-Base; performance is averaged over tasks relative to DSGD-FFT.
(c) Per-task performance of \methodname{} and representative baselines at the communication-cost extremes: DSGD-FFT, DZSGD-FFT, and ChocoSGD-FFALoRA.
\methodname{} outperforms zeroth-order and the most communication-efficient first-order PEFT baselines with model-size-independent communication and $10$--$10^7\times$ fewer communicated bytes.
}
\label{fig:overall}
\vspace{-1.5em}
\end{figure*}

We compare \methodname{} against nine baselines spanning three decentralized optimization methods---DSGD \citep{lian2017can}, communication-efficient ChocoSGD \citep{choco-sgd}, and zeroth-order DZSGD \citep{tang2020distributed}, which isolates the benefit of flooding-based aggregation over naive gossip averaging under the same zeroth-order setting---each combined with FFT, LoRA \citep{lora}, and FFA-LoRA \citep{sun2024improving}, where only trainable parameters are communicated.
To evaluate across diverse tasks and model families, we use a subset of SuperGLUE \citep{wang2019superglue} and SST-2 \citep{socher2013recursive} with OPT-1.3B \citep{opt-model}, and additionally SQuAD \citep{rajpurkar2016squad} and DROP \citep{dua2019drop} with Qwen3-1.7B \citep{yang2025qwen3} to cover generative tasks and a more recent backbone.
We report task performance using accuracy or F1 score, together with total communication volume per edge over training.
To avoid overstating the communication cost of faster-converging first-order methods, we train first-order baselines for $500$ local iterations and zeroth-order methods for $5,000$ which we observed to be sufficient for convergence without overfitting; full configurations and dataset details are provided in Appendix~\ref{app:exp_setup}.

\subsection{Communication--performance trade-off at LLM scale}
\label{sec:main-results}
% We evaluate decentralized fine-tuning over a 16-client ring network with 8 task-model pair.
% Figure~\ref{fig:overall} summarizes the communication--performance trade-off.
% Within zeroth-order training, \methodname{} delivers a markedly better communication--performance trade-off.
Across eight task--model pairs on a 16-client ring network, \methodname{} delivers a markedly better communication--performance trade-off within zeroth-order training (\Cref{fig:overall}). 
Compared with the zeroth-order gossip baselines (e.g., DZSGD and its LoRA/FFA-LoRA variants), \methodname{} transmits $10^4$--$10^7\times$ fewer bytes, yet achieves higher task performance on nearly all tasks.
By replacing multi-hop gossip with flooding and communicating only seeds and scalars, \methodname{} reduces the total communication to just 400~KB over the entire training run.

Compared with the first-order baselines, DSGD-FFT provides the strongest but most communication-intensive reference, achieving the best performance in most settings. 
However, exchanging dense gradients costs 526~GB per edge for OPT-1.3B, corresponding to roughly half a day of transmission over a 100~Mbps link. Against this unconstrained baseline, \methodname{} incurs a 4\% performance gap on OPT-1.3B and as little as 1.1\% on Qwen3-1.7B, while reducing communication by $10^6\times$. In contrast, when FO methods are pushed toward a communication-efficient regime, their performance falls below \methodname{}: Choco-FFA, the most communication-efficient FO baseline at 9.4~MB, underperforms \methodname{} by 2--3\%p.

Overall, \methodname{} combines near-zero, model-size-independent communication with performance close to unconstrained first-order training, while outperforming every baseline operating under a comparable communication budget.
As we show next, this trade-off is specific to small networks: as the number of clients grows, gossip-based methods degrade while SeedFlood does not, turning the trade-off in SeedFlood's favor.
Additional meshgrid topology results are provided in Appendix~\ref{app:Meshgrid}.

\subsection{Performance Scaling with Client Participation}
\label{sec:scaling}
\label{subsec:large_network}

\begin{figure}[t]
    \centering
\includegraphics[width=\textwidth]{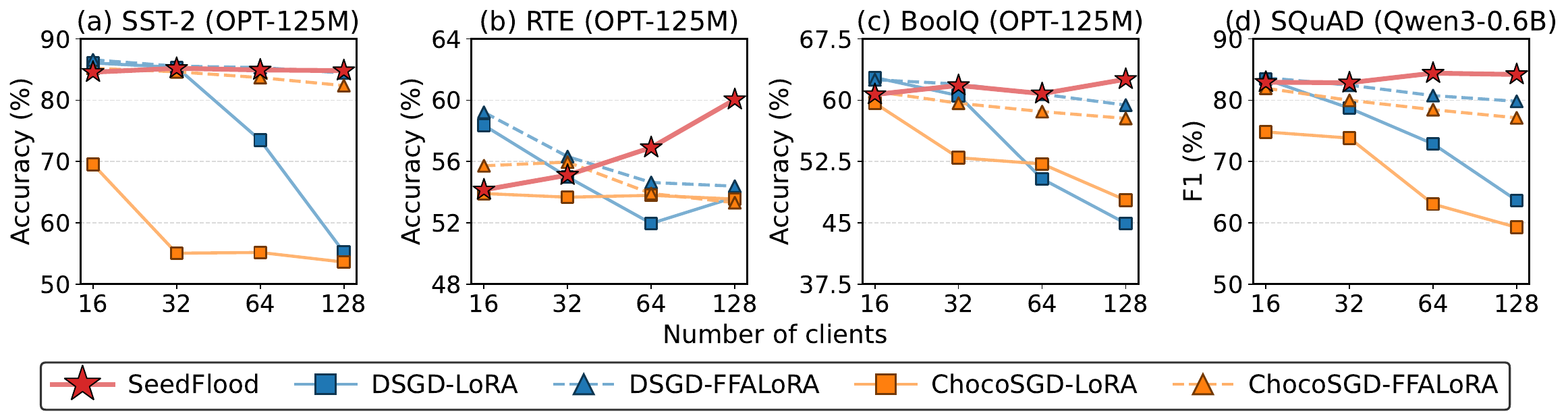}
    \caption{
    Performance scaling with increasing network size in a ring topology. Unlike first-order gossip baselines, \methodname{} maintains stable or improving performance as the number of clients increases, and surpasses the best-performing gossip baseline at larger client counts.
    }
    \label{fig:large_overall_reg}
\end{figure}

Increasing client participation creates opposing effects in conventional decentralized training. More clients provide additional optimization signal, but make global consensus increasingly difficult for gossip-based methods. We examine how \methodname{} behaves under this scaling regime.

% We next study the scalability of \methodname{} under large and highly distributed networks.
% By achieving perfect consensus with negligible communication overhead, \methodname{} remains insensitive to network size, making it well suited for large-scale decentralized training.
We scale the network from 16 to 128 clients while keeping the total training set fixed and evenly partitioning it across clients. For computational feasibility, we evaluate four representative tasks with PEFT baselines.

% The relevant performance which is normalized by the 16-client DSGD is summarized in Table~\ref{tab:largenetwork_overall}.

Figure~\ref{fig:large_overall_reg} reveals contrasting scaling behaviors. While gossip-based baselines degrade as the number of clients increases, \methodname{} maintains or even improves its performance. 
This suggests that exact aggregation not only removes topology-dependent consensus errors but also allows larger collaborations to benefit from the reduced variance of zeroth-order estimates.

\subsection{Iteration Scaling with Client Participation}
\label{sec:iter_budget}

\begin{figure}[t]
    \centering
    \includegraphics[width=1\linewidth]{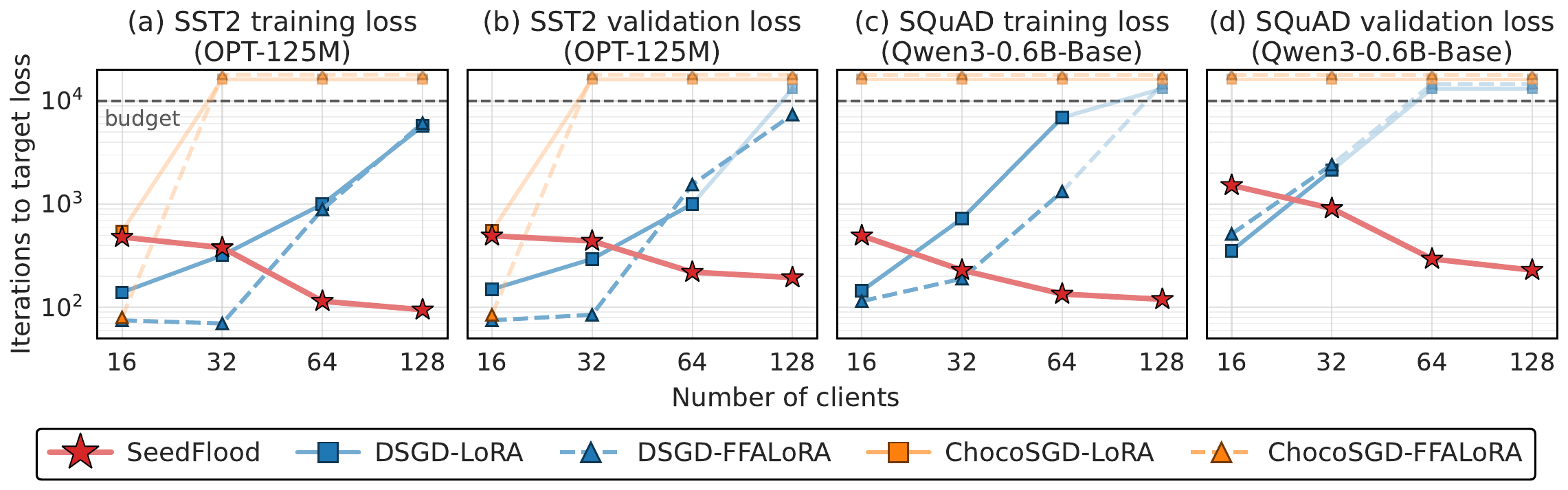}
\caption{Iterations required to reach the target loss across
16--128 clients, measured on training (a, c) and validation
(b, d) sets. While \methodname{} initially requires more
iterations than convergent first-order gossip baselines, its iteration
count decreases with increasing client participation.
In contrast, first-order gossip baselines require more iterations
or fail to reach the target within the 10,000-iteration budget
(dashed horizontal line).}
\label{fig:iter_budget}
\vspace{-1.5em}
\end{figure}

At small scale, \methodname{} requires more iterations than first-order methods, raising the question of whether its communication advantage comes with a persistent optimization cost.
We therefore examine how this iteration gap evolves with increasing client participation by measuring the number of iterations required to reach a target loss under the same setup as Section~\ref{sec:scaling}.
The target is set to the best loss attained by \methodname{} in the 16-client setting; runs are capped at 10,000 iterations, and further protocol details are given in Appendix~\ref{app:iter_budget}.

The iteration narrows sharply as client participation increases (Figure~\ref{fig:iter_budget}).
At 16 clients, \methodname{} requires roughly 3--5$\times$ more iterations than the first-order PEFT baselines to reach the same loss, confirming the expected convergence disadvantage of zeroth-order optimization.
However, this gap narrows as the number of clients increases.
For \methodname{}, the required iteration count decreases with more clients, as aggregating more perturbations in parallel reduces the variance of the zeroth-order estimate, consistent with the $1/n$ noise reduction in Theorem~\ref{thm:seedflood}.
In contrast, the first-order gossip baselines require increasingly more iterations as the number of clients grows and several configurations fail to reach the target.
Consequently, the initial iteration gap between \methodname{} and the gossip baselines narrows substantially as the number of clients increases, and a crossover is observed under the target loss and tasks considered here.

%% file: sections/6-Conclusion.tex
\section{Limitation}

\paragraph{\methodname{} for general decentralized learning.}
While \methodname{} is not inherently restricted to fine-tuning and could, in principle, extend to more general decentralized learning tasks, several considerations make us cautious about drawing concrete conclusions in such settings.
Notably, it would largely inherit the limitations of zeroth-order optimization, whose extension to broader regimes such as pretraining remains an active area of research \citep{DeepZero,allaire2025zeroth}, despite substantial progress in optimization efficiency \citep{ZO-SVRG,MeZO-SVRG,shu2025refining,LOZO,yu2025zeroth,zhao2026pazo,dang2026fzoo}.
However, decentralized training can be limited more directly by slow consensus than by the choice of zeroth- versus first-order local optimization, especially at large collaboration scales \citep{consensus_control,Unified_theory_gossip,hashemi2021benefits}; even with first-order updates, gossip-averaged models may require substantially more iterations to improve, extending into regimes where fully synchronized training has already passed its best test performance and begun to overfit \citep{wang2026controlled,zhu2026surprising}.
This suggests that the exact global consensus enabled by \methodname{} may touch on a critical bottleneck in decentralized learning, providing a basis for broader analysis and potentially inspiring approaches to narrow the gap with centralized learning in more general settings (see Appendix~\ref{app:small_scale_pretraining} for additional small-scale pretraining results).

\paragraph{Dimension-free exact consensus beyond zeroth-order optimization.}
Establishing such a representation beyond zeroth-order optimization would be highly impactful: it could decouple seed--scalar communication from the optimization procedure itself, enabling more effective local optimization methods while preserving exact, dimension-free global consensus.
Although the difficulty discussed in \Cref{sec:Flooding} suggests that such a representation is hard to retain for arbitrary updates, a natural compromise is to restrict optimization to a sufficiently expressive family of seeded updates and project first-order updates onto this family, as recently explored through shared-randomness projection \citep{Ferret}.
Related work further suggests that the seeded family itself can be made more optimization-aware through selective seed dictionaries \citep{FedKseed} or structured low-dimensional perturbations that better capture first-order gradients \citep{LOZO,yu2025zeroth}.
% Apart from this, another direction may be to build on \texttt{PRNG}s constructed from efficiently computable bijections, such as Threefry and Philox \citet{salmon2011parallel}, and investigate whether their reversible structure can be adapted to permit efficient inverse mapping or search for optimization-relevant updates.

% This raises a broader question of whether pseudorandom or learned generators can be designed to provide both tractable encoding and sufficient expressiveness for optimization updates, potentially opening an interesting connection between decentralized learning and the design of pseudorandom representations.

\paragraph{Obtaining further scalability}
% No single method can be expected to perform best across all configurations, and \methodname{} is no exception.
While flooding can reach exact consensus in finite time on any connected topology by propagating updates without dilution, its latency still scales with graph diameter and may become limiting on sufficiently large networks as each update must traverse more hops to reach all clients.
One direct way to mitigate this cost is delayed flooding, where updates traverse only a limited number of hops per round rather than the full graph diameter, reducing per-round propagation latency while allowing global propagation to occur over multiple rounds.
Beyond delayed flooding, system-level mechanisms such as topology-aware relay structures, hierarchical or tree-based routing, message batching, and pipelining across hops or local computation could further reduce effective propagation latency \citep{kshemkalyani2011distributed,vogels2021relaysum,takezawa2023beyond}.
These design choices inevitably trade communication latency against the frequency and freshness of globally incorporated updates, and their impact on end-to-end learning efficiency warrants further study.
We provide partial empirical results for delayed flooding in Appendix~\ref{app:delayed_flooding}, showing that moderate propagation delays can still preserve performance comparable to full flooding.

\section{Conclusion}
\label{sec:conclusion}
Decentralized training has long been hindered by two fundamental barriers: prohibitive communication overhead and the fragility of global consensus.
\methodname{} bypasses these bottlenecks by leveraging the seed-reconstructible structure of zeroth-order updates alongside a robust flooding mechanism, and substantially reduces the communication overhead of decentralized LLM fine-tuning, while maintaining robust performance even in large and sparsely connected peer-to-peer networks. 
Our results further demonstrate that decentralized training and zeroth-order optimization can complement each other's weaknesses: communication-efficient zeroth-order updates with flooding alleviate the communication and consensus bottleneck of decentralized training, while large-scale collaboration helps offset the optimization inefficiency of zeroth-order optimization. These results suggest that \methodname{} provides a practical path toward decentralized LLM fine-tuning in low-bandwidth collaborative environments.

% Nevertheless, \methodname{} does not aim to improve the underlying zeroth-order optimizer itself.
% For tasks with unfavorable optimization landscapes, such as those with high effective dimensionality, eliminating consensus errors alone may not be sufficient to close the optimization gap between zeroth-order and first-order methods.
% Integrating recent advances in zeroth-order optimization \citep{MeZO-SVRG,shu2025refining, zhao2026pazo, dang2026fzoo} with \methodname{} to further improve optimization efficiency remains a promising direction for future work.

%% file: sections/appendix-Algorithm.tex
\section{Algorithm overview}
\label{app:algorithm}
\input{algorithim/SeedFlood}

Algorithm~\ref{alg:seedflood} assembles the three primitives.
{(A)~Subspace setup:} every $\tau$ iterations, all clients re-derive the shared basis $(\mathbf{U}, \mathbf{V})$ from the global seed.
{(B)~Local gradient estimation:} each client samples a minibatch and a local seed; the seed selects one coordinate of the shared basis per layer(\textsc{RNG\_SF}), and the two-point finite difference along the induced rank-one direction yields the seed--scalar message, with the step size and the $1/n$ averaging factor folded into its scalar.
{(C)~Flooding and aggregation:} the message is flooded for $D$ rounds following the relay rule of \Cref{sec:Flooding}, and each client applies newly received messages through the coordinate accumulation of
\Cref{sec:aggregation}.

 \begin{wraptable}[7]{r}{0.40\linewidth}
    \vspace{-1.3\baselineskip}
    \centering
    \scriptsize
    \caption{
    Communication overhead and consensus properties.
        }
    \label{tab:comm_comp_compare}
    \renewcommand\arraystretch{0.9}

    \resizebox{\linewidth}{!}{%
    \begin{tabular}{lccc}
        \toprule
        {Approach} &
        \begin{tabular}[c]{@{}c@{}}{Communicated}\\{Bytes}\end{tabular} &
        \begin{tabular}[c]{@{}c@{}}{Applying}\\{Computation}\end{tabular} &
        \begin{tabular}[c]{@{}c@{}}{Perfect}\\{Consensus}\end{tabular} \\
        \midrule
        Traditional Gossip    & $\mathcal{O}(d)$   & $\mathcal{O}(d)$    & -- \\
        % Gossip with SR\textsuperscript{$\dagger$}  & $\mathcal{O}(tn)$  & $\mathcal{O}(tnd)^{\ddagger}$  & -- \\
        \rowcolor{red!8}\methodname{} (ours) & $\mathcal{O}(n)$   & $\mathcal{O}(n + rd)$ & $\checkmark$ \\
        \bottomrule
    \end{tabular}%
    }
\end{wraptable}

Table~\ref{tab:comm_comp_compare} contrasts the resulting costs, where $n$ is the number of clients, $d$ the parameter dimension, and $r$ the subspace rank, with $d \gg n \geq r$. Traditional gossip pays $O(d)$ communication per edge without reaching exact consensus, whereas \methodname{} attains exact consensus with $O(n)$ communication---one compact message per client---and $O(n + rd)$ aggregation cost.

%% file: algorithim/SeedFlood.tex
\SetKwFunction{RNGS}{RNG\_SF}
\SetKwFunction{ZOEstimate}{ZOEstimate}
\SetKwFunction{SetupSubspace}{SetupSubspace}

\SetKwProg{Fn}{Function}{:}{}
\SetKwInOut{KwInput}{Input}
\SetKwInOut{KwOutput}{Output}

% \begin{wrapfigure}{r}{0.55\textwidth}
% \vspace{-\intextsep}
% \begin{minipage}{\linewidth}
\begin{algorithm}[H]
\setlength{\columnsep}{2em}         
\setlength{\columnseprule}{0.8pt}
\begin{multicols}{2}
\DontPrintSemicolon
\scriptsize

\caption{\methodname}
\label{alg:seedflood}

\KwInput{
Initial models $\{\theta_i^0\}_{i=1}^n$ are identical;
clients $n$; local data $\mathcal{D}_i$; diameter $D$;
learning rate $\eta_t$; total steps $T$;
perturbation scale $\epsilon$; period $\tau$; rank $r$;
stochastic loss $F(\theta;\mathcal{B})$;
$\mathcal{S}_i=\emptyset$, $\mathcal{R}_i=\emptyset$;
global seed $s^{\rm glob}$.
}

\KwOutput{
Global model
$\bar{\theta}=\frac{1}{n}\sum_{i=1}^n\theta_i^T=\theta_i^T$.
}
\vspace{0.3em}
\For{$t=0,\ldots,T-1$ in parallel for all $i\in[n]$}{

    \tcc{\textbf{(A) Shared subspace setup}}
    \If{$t \bmod \tau = 0$}{
        $(\mathcal{U},\mathcal{V})
        \gets
        \texttt{SetupSubspace}(s^{\rm glob},t,r)$\;
    }
    \tcc{\textbf{(B) Local ZO gradient estimation}}
    $(\alpha_{i,t}, z_{i,t}, s_{i,t})
    \gets
    \texttt{ZOEstimate}
    (\theta_i^t,\mathcal{D}_i,\mathcal{U},\mathcal{V},\epsilon)$\;
    $\theta_i^{t,0} = \theta_i^{t} - \eta_t \frac{1}{n}\alpha_{i,t} z_{i,t}$\;
    \tcc{\textbf{(C) Flooding iter \& aggregation}}
    $\mathcal{R}_i
    \gets
    \mathcal{R}_i\cup
    \{(s_{i,t},\eta_t\alpha_{i,t}/n)\}$\;
    $\mathcal{S}_i
    \gets
    \mathcal{S}_i\cup
    \{(s_{i,t},\eta_t\alpha_{i,t}/n)\}$\;
    \For{$d=0,\ldots,D-1$}{
        Send $\mathcal{R}_i$ to all neighbors $j\in\mathcal{N}(i)\setminus\{i\}$\;

        $\mathcal{R}_i\gets$ messages received from neighbors\;

        $\mathcal{R}_i\gets\mathcal{R}_i\setminus\mathcal{S}_i$\;

        $\mathcal{S}_i\gets\mathcal{S}_i\cup\mathcal{R}_i$\;

        $\Delta\gets
        \sum_{(s_k,v_k)\in\mathcal{R}_i}
        v_k\,\texttt{RNG\_SF}(s_k,\mathcal{U},\mathcal{V})$\;

        $\theta_i^{t,d+1}\gets\theta_i^{t,d}-\Delta$\;
    }

    $\theta_i^{t+1}\gets\theta_i^{t,D}$\;
}

\Fn{\SetupSubspace{$s^{\rm glob},t,r$}}{
    Initialize RNG with seed $s^{\rm glob}+t$\;

    \ForEach{2D layer $\theta_\ell\in\mathbb{R}^{n_\ell\times m_\ell}$}{
        $U_\ell \sim \mathcal{N}(0,1)^{n_\ell\times r}, \quad V_\ell \sim \mathcal{N}(0,1)^{m_\ell\times r}$\;
    }

    $\mathcal{U}\gets\{U_\ell\}_{\ell=1}^L$,
    $\quad \mathcal{V}\gets\{V_\ell\}_{\ell=1}^L$\;

    \Return{$(\mathcal{U},\mathcal{V})$}\;
}

\Fn{\ZOEstimate{$\theta,\mathcal{D},\mathcal{U},\mathcal{V},\epsilon$}}{
    Sample minibatch $\mathcal{B}\sim\mathcal{D}$ and random seed $s$\;

    $z\gets\texttt{RNG\_SF}(s,\mathcal{U},\mathcal{V})$\;

    $F^+\gets F(\theta+\epsilon z;\mathcal{B}) ,\quad F^-\gets F(\theta-\epsilon z;\mathcal{B})$\;

    $\alpha\gets(F^+-F^-)/(2\epsilon)$\;

    \Return{$(\alpha,z,s)$}\;
}

\Fn{\RNGS{$s,\mathcal{U},\mathcal{V}$}}{
    Initialize RNG with seed $s$\;

    \ForEach{layer $\theta_\ell$}{
        \If{$\theta_\ell$ is a 2D matrix}{
            Sample $i_\ell,j_\ell\sim\mathrm{Unif}\{0,\ldots,r-1\}$\;

            $z_\ell\gets
            \operatorname{vec}
            \left(
            U_\ell[:,i_\ell]V_\ell[:,j_\ell]^\top
            \right)$\;
        }
        \Else{
            Sample
            $z_\ell\sim
            \mathcal{N}(0,1)^{\operatorname{shape}(\theta_\ell)}$\;
        }
    }
    \Return{$z=[z_1,z_2,\ldots,z_L]$}\;
}
\end{multicols}
\vspace{0.7em}
\end{algorithm}

% \end{minipage}
% \vspace{-\intextsep}
% \end{wrapfigure}

%% file: sections/appendix-additional_experiment.tex
\section{Additional Experimental Results}
\label{app:additional_experiments}

In this section, we provide additional experimental results that complement the empirical evaluation in the main paper.

\subsection{Additional Meshgrid Topology Results}
\label{app:Meshgrid}

\begin{figure}
    \centering
    \includegraphics[width=1\linewidth]{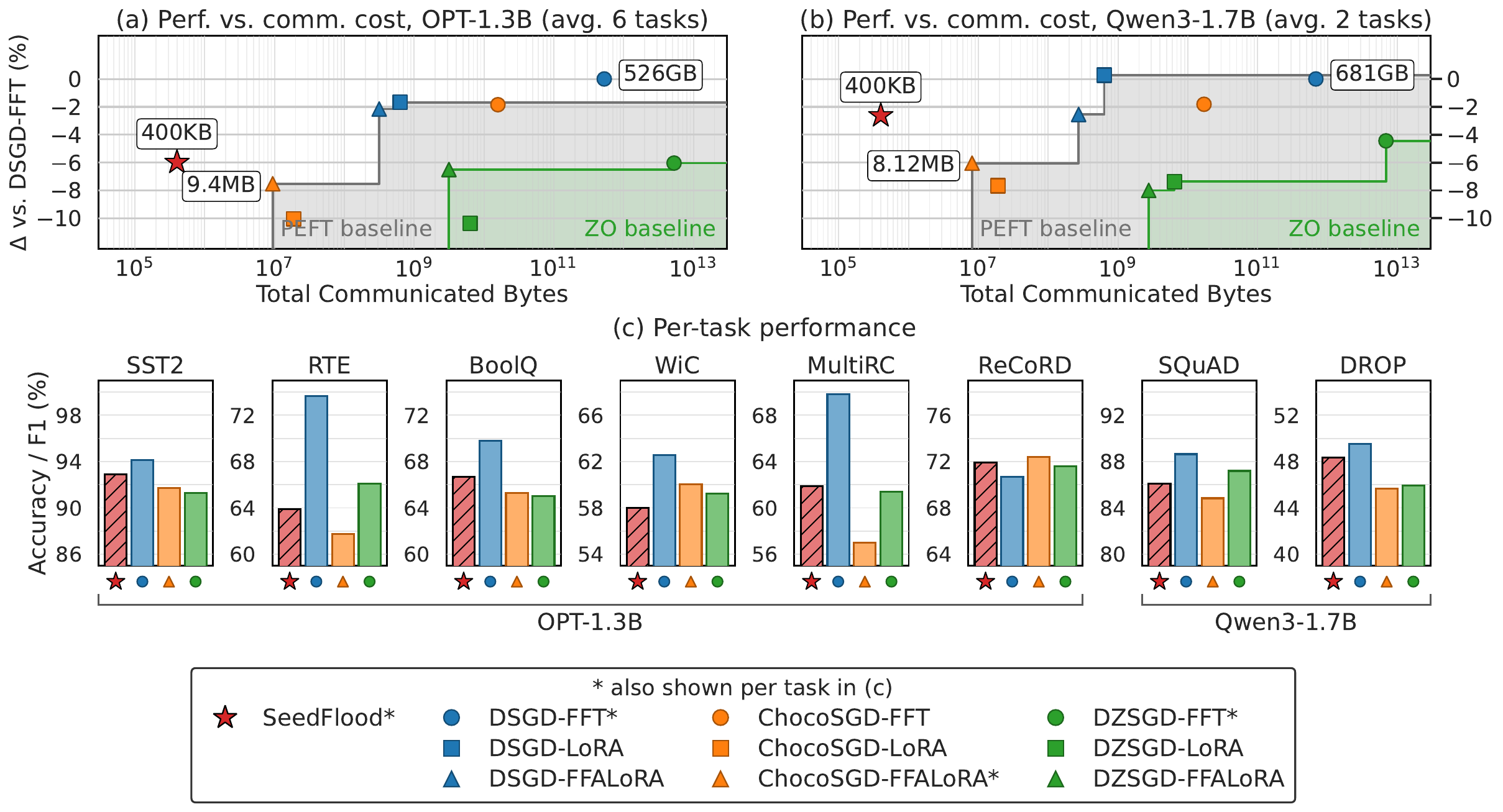}
    \caption{Performance–communication trade-off on a 16-client meshgrid network.}
    \label{fig:meshgrid_main}
\end{figure}
We additionally repeat the experiment in \Cref{sec:main-results} under a denser meshgrid topology (\Cref{fig:meshgrid_main}). As expected, several gossip-based baselines show modest performance improvements with increased connectivity, while \methodname{} remains competitive despite being largely insensitive to the underlying topology.

\subsection{Characterization of the subspace gradient estimator} \label{app:char_sub_est}
% \begin{wrapfigure}[18]{r}{1\linewidth}

\begin{figure}[ht]
\centering
\begin{minipage}{0.48\linewidth}
    \centering

    % ---------------------------------------------------------
    % Figure 1: Runtime
    % ---------------------------------------------------------
    \includegraphics[width=0.85\linewidth]{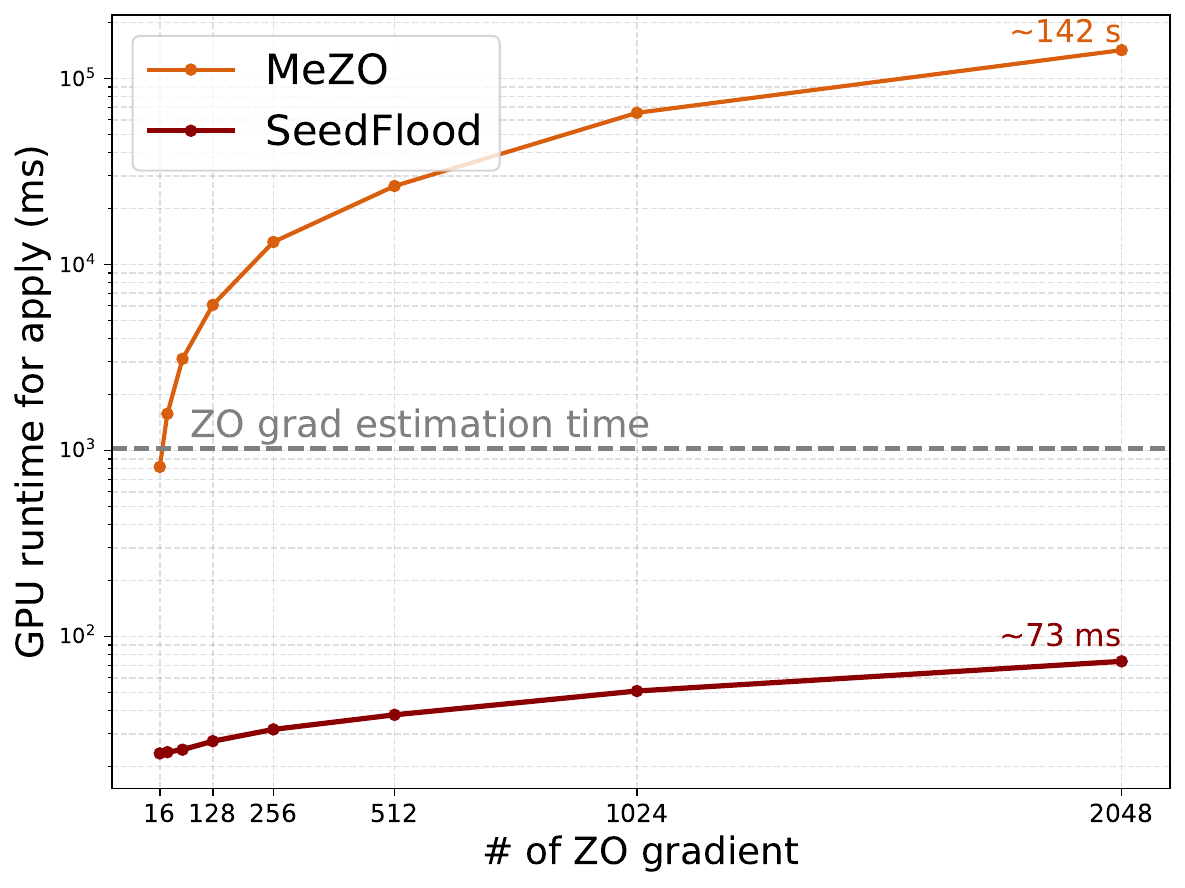}
    \captionof{figure}{
    Runtime for applying multiple ZO messages to OPT-2.7B on a single A100 GPU with subspace size $R=50$.
    }
    \label{fig:MA_times}
    \vspace{0.5em}
\end{minipage}
\hspace{0.5em}
\begin{minipage}[r]{0.48\linewidth}
    % ---------------------------------------------------------
    % Table: Single-client performance
    % ---------------------------------------------------------
    \footnotesize
    \captionof{table}{
    Single-client finetuning of OPT-2.7B on SuperGLUE + SST-2 (\%). ``Avg'' is  normalized by MeZO and averaged.
    }
    \label{tab:subcge_mezo_single}

    \renewcommand\arraystretch{0.85}
    \resizebox{\linewidth}{!}{
    \begin{tabular}{lccccccc}
        \toprule
        Method & SST2 & RTE & BoolQ & WiC & MultiRC & ReCoRD & Avg \\
        \midrule
        MeZO   & 93.58 & 62.46 & 64.70 & 56.27 & 56.90 & 76.50 & 0.00\% \\
        \methodname{} & 92.89 & 63.18 & 64.90 & 56.11 & 58.90 & 76.30 & +0.62\% \\
        \bottomrule
    \end{tabular}
    }

    % ---------------------------------------------------------
    % Figure 2: Hyperparameter sensitivity
    % ---------------------------------------------------------
    \includegraphics[width=\linewidth]{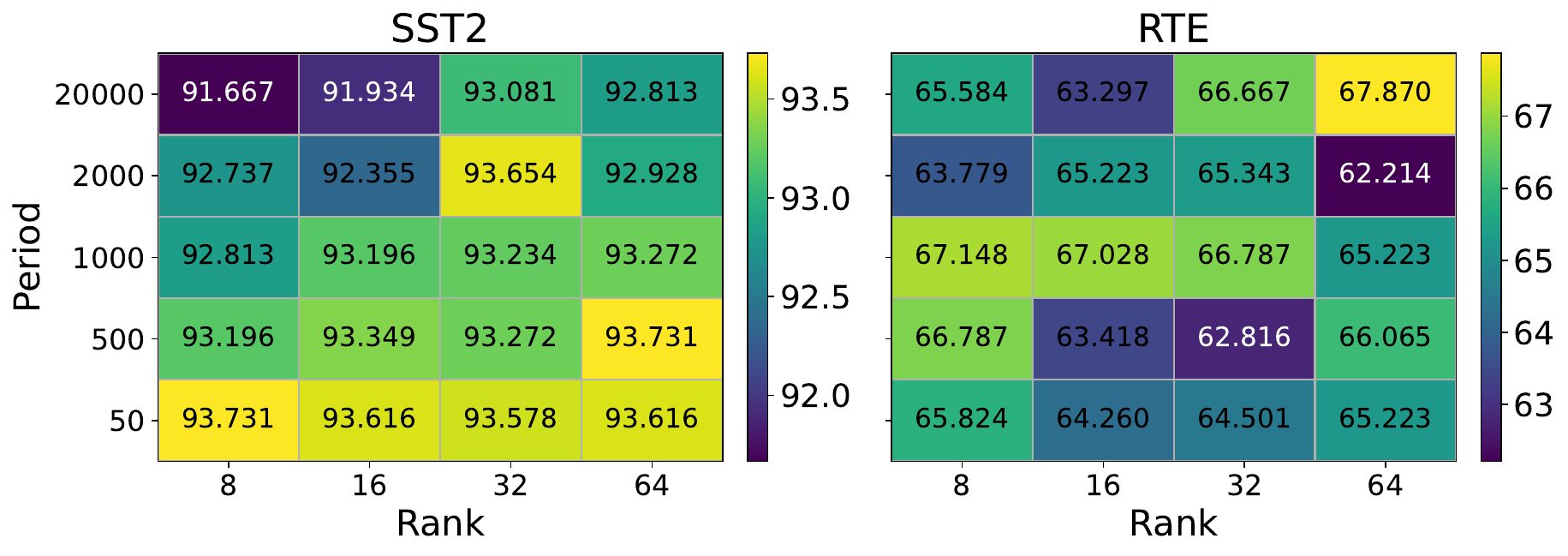}
    \captionof{figure}{
    Sensitivity of \methodname{} to the subspace rank and refresh period on SST-2 and RTE.
    % Performance degrades when the search space remains overly small throughout training.
    }
    \label{fig:hp_sens}
    \vspace{0.5em}
\end{minipage}
\vspace{-2.0em}
\end{figure}
% \end{wrapfigure}

\paragraph{Setup.}
For the gradient-estimator ablation study, we conduct 20K training iterations on a single OPT-2.7B client. 
In the overall performance experiments, the subspace rank and update interval are fixed to 32 and 500, respectively, without additional search.
For the hyperparameter sensitivity analysis, we evaluate subspace ranks in 
{8,16,32,64} and update intervals in {50,500,1000,2000,20000}, where an interval of 20000 corresponds to no subspace update during training.
All other settings, including prompts, hyperparameters, and learning rate search procedures, follow the same experimental setup.

As shown in Figure~\ref{fig:MA_times}, \methodname{}'s gradient estimator maintains low and scalable aggregation overhead, remaining orders of magnitude faster than MeZO even with thousands of gradient updates. We provide implementation details and runtime analysis in Appendix~\ref{appendix:Subcgeruntime}.

We further verify that this computational gain does not come from an ineffective estimator. In a single-client OPT-2.7B fine-tuning experiment, \methodname{} 's estimator matches MeZO-level performance on average 
as seen in Table~\ref{tab:subcge_mezo_single}. Hyperparameter ablation shows that larger rank and more frequent subspace refreshes generally improve performance. However, on RTE, excessively frequent refreshes at large ranks can degrade performance, likely because the optimizer moves to a new subspace before sufficiently exploiting the current one (see Figure~\ref{fig:hp_sens}).

\subsection{Delayed flooding} \label{app:delayed_flooding}

\begin{figure}
    \vspace{-1.0\baselineskip}
    \centering
    \includegraphics[width=0.6\linewidth]{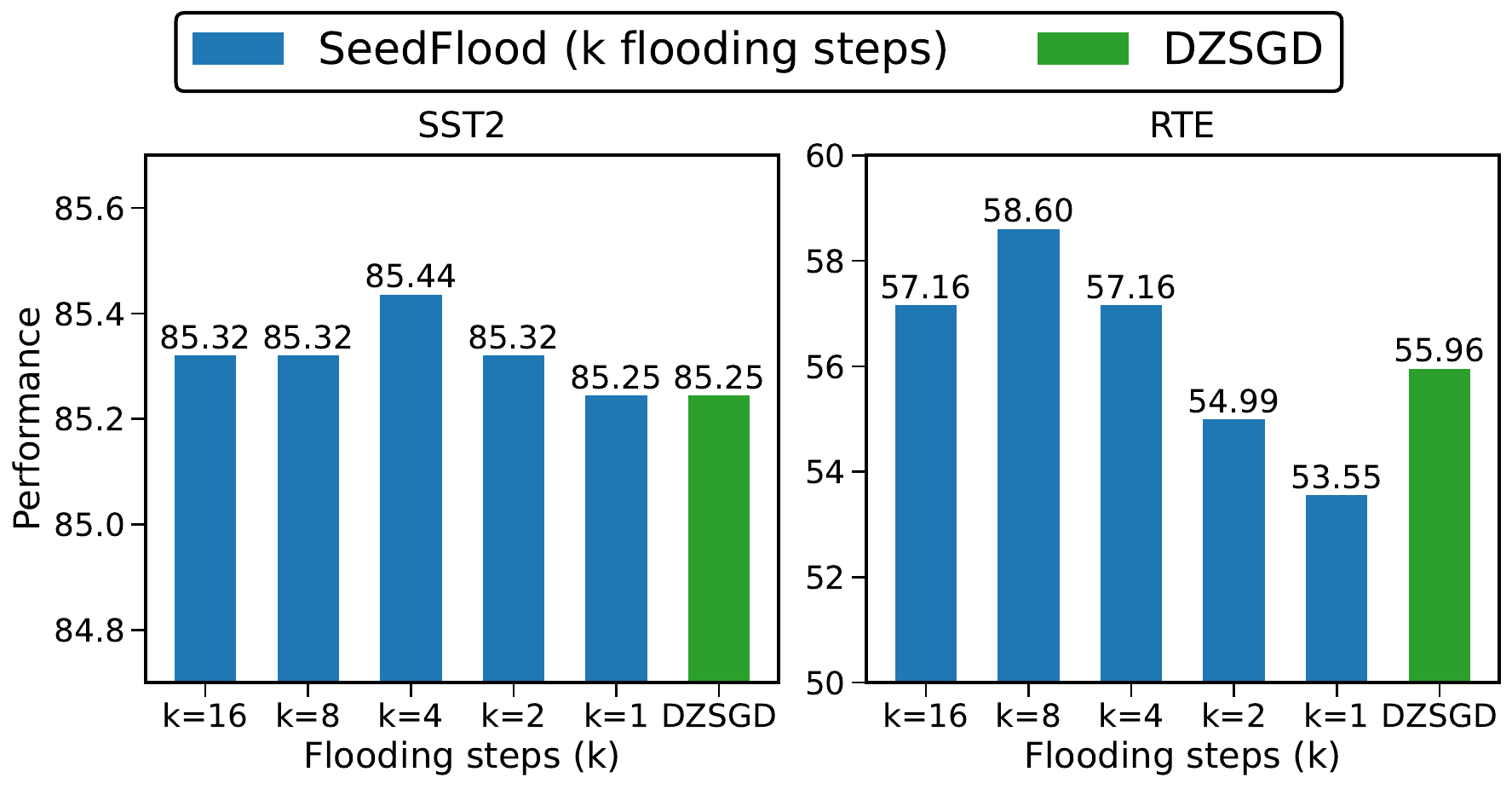}
    \caption{
    Task performances on SST2 and RTE under partial flooding with varying number of flooding steps.
    }
    \label{fig:partial_flood}
\end{figure}
In practice, the exact network diameter may be unknown or full propagation before each update may be too restrictive. We thus evaluate a relaxed variant performing only k flooding steps per round.
We evaluate delayed flooding by training OPT-125M on a ring topology with 32 clients, using $k \in \{1, 2, 4, 8, 16\}$.
As shown in Figure~\ref{fig:partial_flood}, \methodname{} exhibits no observable performance degradation for moderate values of $k$ ($k \ge 4$), maintaining performance comparable to full flooding and consistently outperforming DZSGD.
This indicates that the benefit of flooding-based global aggregation {does not rely on strictly enforcing perfect consensus at every iteration,}
but rather on sufficiently fast global propagation of update information.
Overall, \methodname{} remains effective without exact full flooding, as long as update propagation delay is sufficiently bounded.

\subsection{Small-Scale Pretraining from Scratch}
\label{app:small_scale_pretraining}

Although the main scope of this paper is decentralized LLM fine-tuning, this choice is intended to avoid overstating our claims as large-scale LLM pretraining.
The underlying framework of \methodname{}, however, is not restricted to fine-tuning.
When the model size is small enough for zeroth-order optimization to be practical, \methodname{} can also be applied to training from scratch.

We therefore consider a small-scale training-from-scratch setup previously studied in ~\citet{DecomFL}.
To adapt this setup to decentralized learning, we replace the centralized communication structure with a 32-client ring topology and compare against decentralized baseline algorithms.
Specifically, we train a 3-layer CNN from scratch on MNIST under both IID and non-IID data partitions.
The results are summarized in Table~\ref{tab:mnist_pretraining}.

\begin{table}[t]
    \footnotesize
    \centering
    \caption{
    Training from scratch on MNIST with a 3-layer CNN under a 32-client ring topology.
    \methodname{} achieves competitive accuracy while requiring substantially less communication than dense decentralized baselines.
    }
    \label{tab:mnist_pretraining}
    \renewcommand{\arraystretch}{1.05}
    \begin{tabular}{lcccc}
        \toprule
        Method &
        IID Acc. &
        non-IID Acc. &
        Total Comm. &
        Comm. / Iter. \\
        \midrule
        DSGD      & 97.57 & 96.08 & 260 MB  & 173 KB \\
        ChocoSGD  & 90.47 & 69.01 & 5.24 MB & 3.5 KB \\
        \methodname{} & 94.17 & 92.82 & 2.78 MB & 0.3 KB \\
        DZSGD     & 94.10 & 84.45 & 1.3 GB  & 173 KB \\
        \bottomrule
    \end{tabular}
\end{table}

The results show that \methodname{} is also effective beyond LLM fine-tuning.
Compared with dense communication baselines such as DSGD and DZSGD, \methodname{} reduces communication by orders of magnitude while maintaining competitive accuracy.
In particular, under the non-IID setting, \methodname{} substantially outperforms ChocoSGD and DZSGD, despite using the smallest per-iteration communication among all methods.
These results suggest that the proposed seed-based flooding framework is applicable not only to communication-efficient LLM fine-tuning, but also to small-scale decentralized training from scratch.

\subsection{Supervised Instruction Tuning}
\label{app:sft}
 
The main experiments evaluate SeedFlood on downstream fine-tuning tasks. 
This appendix asks whether the same estimator is usable for a supervised instruction tuning, under a same iteration budget. 
we report one model on one instruction-following corpus.
 
\paragraph{Setup.}
We instruction-tune Qwen3-8B-Base~\citep{yang2025qwen3} on
{tulu-3-sft-personas-instruction-following}~\citep{lambert2024tulu}, 29980
single-turn examples. We split 90/5/5 into 26982 / 1499 /
1499 for train / validation / test and format every example with the
model's own ChatML template. All decentralized methods use FFA-LoRA~\citep{sun2024improving}
($r{=}8$ on $\mathbf{q}_{\text{proj}}, \mathbf{v}_{\text{proj}}$; $A$ frozen
and shared, only $B$ trained and communicated) on a 32-client ring. The token budget and iteration is
identical for every method: 3 epochs, 159 iterations. For each method we sweep the learning rate over a three-point grid
We evaluate on IFEval (541 generative
prompts; prompt- and instruction-level, strict and loose) \citep{ifeval}, BBH (18 subtasks,
0-shot, log-likelihood) \citep{BBH} and MMLU (57 subjects, 0-shot, log-likelihood)\citep{MMLU}; the
last two check that instruction tuning does not degrade knowledge or
reasoning.

\providecommand{\ms}[2]{#1{\tiny\,$\pm$#2}}
\begin{table}[t]
\centering
\small
\renewcommand{\arraystretch}{1.05}
\setlength{\tabcolsep}{2.8pt}
\caption{%
Instruction tuning Qwen3-8B-Base on Tulu-3 personas-IF on a 32-client ring
under a matched budget of 3 epoch. Mean $\pm$ std
over three seeds; all numbers in \%. The first-order and DZO baselines train
an FFA-LoRA adapter ($r{=}8$); SeedFlood trains the full parameter set through
seed-encoded perturbations. 
Comm.\ is the total payload one client sends over the 159-iteration run.
Learning rates were selected by validation loss from a per-method grid. Best result in
bold.}
\label{tab:sft}
\begin{tabular}{@{}lrcccccc@{}}
\toprule
 & & \multicolumn{4}{c}{IFEval} & & \\
\cmidrule(lr){3-6}
 & Comm. & Prompt & Inst. & Prompt & Inst. & & \\
Method & (MB total) & strict & strict & loose & loose & BBH & MMLU \\
\midrule
% --- optional: untuned base ------------------------------------------------
Qwen3-8B-Base (untuned) & -- & 46.4 & 58.5 & 52.1 & 63.2 & 47.4 & 69.1 \\
\midrule
% ---------------------------------------------------------------------------
DSGD-FFALoRA     & 182.9 & \ms{51.1}{0.3} & \ms{61.5}{0.5} & \ms{55.3}{0.3} & \ms{65.3}{0.5} & \ms{\textbf{49.0}}{0.1} & \ms{71.5}{0.2} \\
ChocoSGD-FFALoRA & 5.5   & \ms{50.3}{1.7} & \ms{61.0}{1.6} & \ms{55.0}{1.2} & \ms{64.9}{1.2} & \ms{48.9}{0.3} & \ms{\textbf{71.7}}{0.2} \\
DZO-FFALoRA      & 182.9 & \ms{46.7}{0.7} & \ms{58.4}{0.2} & \ms{52.1}{0.7} & \ms{62.7}{0.3} & \ms{47.1}{0.1} & \ms{69.0}{0.2} \\
SeedFlood (ours) & \textbf{0.04} & \ms{\textbf{51.8}}{0.8} & \ms{\textbf{62.6}}{0.4} & \ms{\textbf{56.1}}{1.0} & \ms{\textbf{66.1}}{0.6} & \ms{48.5}{0.2} & \ms{71.2}{1.8} \\
%% --- optional: single-client references ------------------------------------
% \midrule
% \multicolumn{8}{@{}l}{\textit{Single-client SGD reference$^\dagger$ (no communication constraint; not comparable)}} \\
% SGD-FFALoRA, 1 client$^\dagger$ & -- & \ms{49.7}{1.2} & \ms{60.6}{1.1} & \ms{54.4}{1.2} & \ms{64.6}{1.0} & \ms{48.9}{0.2} & \ms{71.4}{0.1} \\
% SGD-Full, 1 client$^\dagger$    & -- & \ms{69.0}{0.4} & \ms{76.7}{0.6} & \ms{72.7}{0.6} & \ms{80.1}{0.6} & \ms{52.7}{1.0} & \ms{74.4}{0.4} \\
%% ---------------------------------------------------------------------------
\bottomrule
\end{tabular}
\end{table} 

\paragraph{Result.}
SeedFlood, DSGD and ChocoSGD finish in one tier (Table~\ref{tab:sft}): on
IFEval prompt-level strict accuracy they reach 51.8\,$\pm$\,0.8,
51.1\,$\pm$\,0.3 and 50.3\,$\pm$\,1.7 from an untuned 46.4, and every gap
among the three is inside one seed's spread. What separates them is
communication: SeedFlood sends 0.04\,MB per client over the whole run against
182.9\,MB for DSGD and 5.5\,MB for ChocoSGD, while updating the full 8.2B
parameters rather than an $r{=}8$ adapter. DZO does not train: its validation
loss moves by less than $5\times10^{-4}$ at every learning rate and every
benchmark stays within one point of the untuned model.

%% file: sections/appendix-Discussion.tex
\section{Alternatives for Integrating Seed–Scalar Communication with Gossip}
% \section{Why not combine seed-based communication with gossip?}
\label{app:seed_gossip}

A natural alternative to \methodname{} is to combine seed-based communication with existing
gossip-based decentralized training frameworks.
Since shared randomness allows a gradient update to be represented by a seed--scalar pair,
one may try to avoid exchanging dense model parameters by representing each local model as
an accumulation of past seed-reconstructible updates.
Formally, let $\mathcal{M}_i^t$ denote the set of update messages known to client $i$ at iteration $t$.
Then the local model can be written as
\begin{equation}
\theta_i^t
= \theta^0 - \sum_{m\in\mathcal{M}_i^t}
c_{i,t}(m)\, \alpha(m)\, \mathrm{RNG}\!\bigl(s(m)\bigr),
\label{eq:seed_gossip_model}
\end{equation}
where $s(m)$ and $\alpha(m)$ denote the seed and scalar coefficient of update message $m$,
and $c_{i,t}(m)$ denotes the current gossip-induced mixing coefficient assigned to that update.

Under gossip-based model averaging, clients repeatedly mix their local models with neighbors.
In the seed-based representation above, this corresponds to exchanging the update histories
and averaging the coefficients of all known messages:
\begin{equation}
c_{i,t+1}(m)
= \sum_{j\in\mathcal{N}(i)} w_{ij}\, c_{j,t}(m),
\qquad
\forall m \in \bigcup_{j\in\mathcal{N}(i)} \mathcal{M}_j^t .
\label{eq:seed_gossip_coeff}
\end{equation}

\begin{figure}
    \centering
    \vspace{-\baselineskip}
    \includegraphics[width=0.8\linewidth]{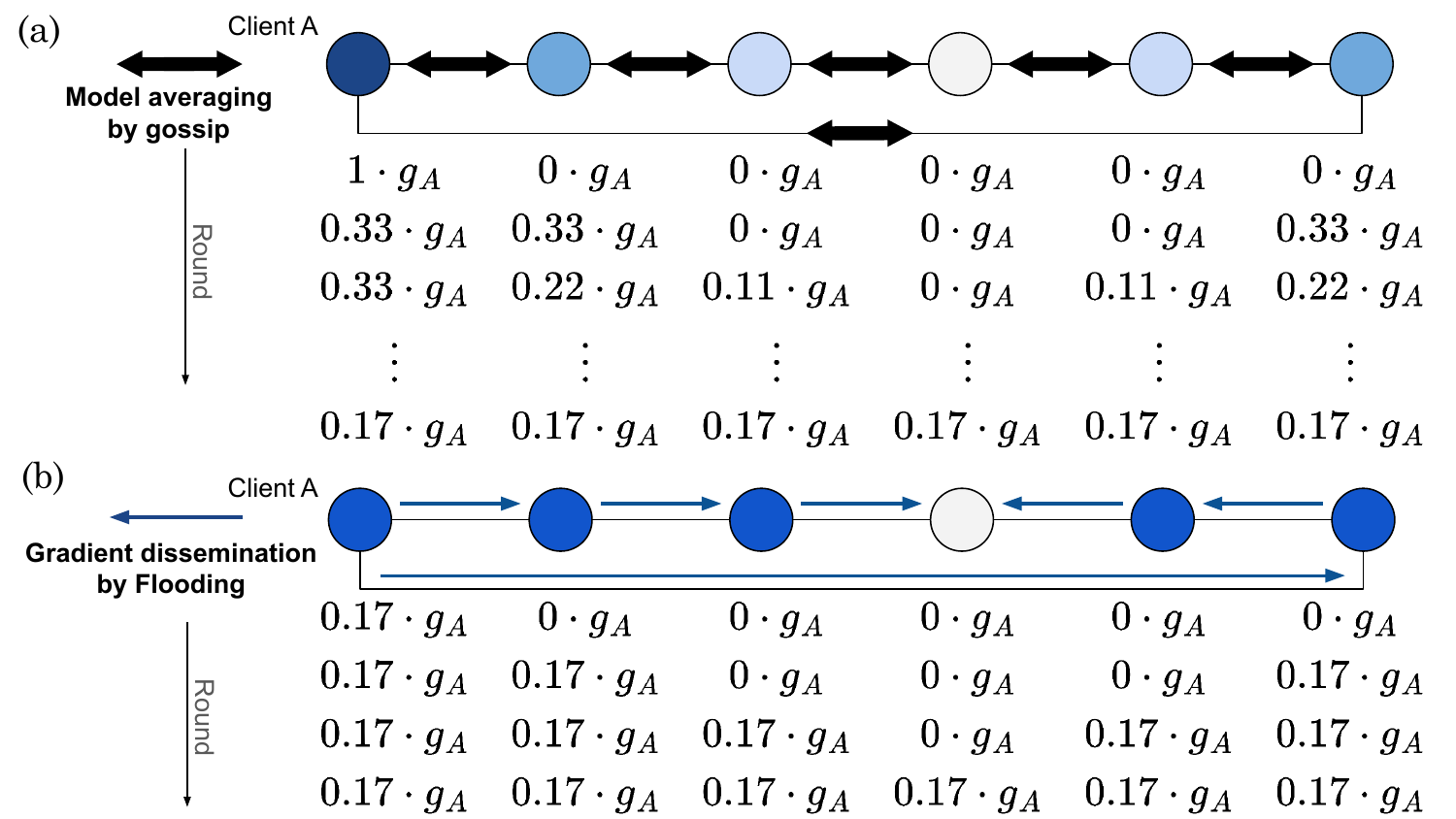}
    \caption{Consensus dynamics of a single gradient under gossip-based model averaging (a) and flooding-based gradient dissemination (b).}
    \label{fig:gossipvsflooding}
\end{figure}

This idea is closely related in spirit to prior centralized federated learning methods that
represent model updates through seed-based histories \citep{FedKseed, DecomFL}.
Because each message is only a constant-size seed--scalar pair, communicating all updates
generated up to iteration $t$ requires $O(tn)$ communication, which is much smaller than
transmitting dense model parameters of dimension $d$.

However, this communication saving does not translate into a scalable decentralized algorithm
when combined with gossip.
The key issue is that gossip does not merely disseminate an update; it repeatedly changes the
effective coefficient of every past update as models are mixed over the network.
Figure~\ref{fig:gossipvsflooding}(a) illustrates this behavior.
Even a single update generated by client $A$ is gradually diffused through repeated local averaging,
and its contribution to each client's model changes at every gossip step.

This creates a computational bottleneck under seed-based communication.
Whenever the coefficient $c_{i,t}(m)$ of a past update changes, client $i$ must realize the
corresponding change in the dense model space.
Equivalently, it must reconstruct the perturbation $\mathrm{RNG}(s(m))$ from its seed and apply
the coefficient difference to the model.
Since the number of stored update messages grows as $O(tn)$ and reconstructing each update costs
$O(d)$, maintaining the gossip-averaged model can require $O(tnd)$ computation over time.
Thus, the bottleneck is shifted from communication to reconstruction.

This limitation is structural rather than merely implementation-specific.
Gossip is designed for iterative averaging of dense model states, where repeated reweighting is
the mechanism by which consensus is reached.
In contrast, seed-based communication is most effective when each transmitted object is an
individual update that can be reconstructed and applied with a fixed coefficient.
Repeatedly reweighting the same seed-reconstructible updates undermines this advantage, because
it forces clients to revisit and reconstruct an ever-growing history of past updates.

This observation motivates the flooding-based design of \methodname{}.
Instead of gradually diffusing model states through repeated averaging, flooding disseminates
each update message itself across the network.
Once a client receives a previously unseen update, it applies the update exactly once with its
fixed coefficient and forwards the message to its neighbors.
As illustrated in Figure~\ref{fig:gossipvsflooding}(b), the role of communication is no longer to
iteratively reweight local models, but to ensure that all clients eventually receive the same set
of update messages.
Therefore, under seed-based communication, flooding provides a more compatible consensus primitive:
it preserves the communication benefit of shared randomness while avoiding the repeated
reconstruction cost induced by gossip.

%% file: sections/appendix-Runtime_analysis.tex
\section{Implementation details and runtime analysis of gradient estimator}
\label{appendix:Subcgeruntime}
In this section, we provide implementation details of our subspace coordinate-wise gradient estimator and a detailed breakdown of its runtime components.
In our implementation, each two-dimensional weight matrix maintains an additional buffer of the form \(U_\ell A_\ell V_\ell^\top\).
During training, all coordinate-wise updates are accumulated into \(A_\ell\), and the full update \(U_\ell A_\ell V_\ell^\top\) is applied to the base weight on-the-fly during the forward pass.
This design reduces the cost of applying received messages to \(O(n)\) and shifts the \(O(rd)\) matrix multiplication cost to the forward pass.
Moreover, operations required for generating and updating perturbation vectors—which account for approximately 25\% of the gradient estimation cost in MeZO—are replaced with lightweight coordinate updates in our estimator.

Table~\ref{table:subcge_runtime} presents a detailed runtime comparison between our estimator and MeZO.
For each zeroth-order method, the total runtime is decomposed into gradient estimation and message application phases.
The cost of applying multiple messages further consists of random number generation and arithmetic operations for parameter updates.

Our estimator follows MeZO's procedure only for a small number of one-dimensional tensors, while for two-dimensional tensors it performs coordinate updates on the buffer \(A_\ell\).
In the gradient estimation phase, the additional \(O(rd)\) matrix multiplication required to inject buffer values into the model slightly increases the forward-pass time; however, this overhead is offset by the substantially reduced perturbation cost.

Overall, although our estimator has a theoretical computation complexity of \(O(n + rd)\), our implementation effectively eliminates the computation overhead of message application in practice, with the additional \(O(rd)\) computation largely hidden by the efficiency gains in perturbation.
\begin{table}[h]
    \footnotesize
    \centering
    \caption{Detailed wall clock time per iteration of the \methodname{} framework with
MeZO versus our subspace coordinate-wise gradient estimator. Results are averaged over 5 steps on OPT-2.7B with
batch size~16 and 16 clients (i.e., 16 ZO-gradient messages generated
per iteration). ``GE'' denotes the gradient estimation phase, and ``MA''
denotes the message-applying phase.}
    \label{table:subcge_runtime}
\begin{tabular}{lccc}
\toprule
\multicolumn{4}{c}{\textbf{1. Overall runtime (ms)}} \\
\midrule
Method & GE & MA & Total time \\
\midrule
MeZO   & 1077 & 1432 & 2509 \\
Ours   & \;914 & \;\;28 & \;942 \\
\midrule
\multicolumn{4}{c}{\vspace{-0.5em}} \\
\toprule
\multicolumn{4}{c}{\textbf{2. Gradient estimation phase (ms)}} \\
\midrule
Method & Forward pass & Perturbation & Local update \\
\midrule
MeZO   & 757 & 239 & 80 \\
Ours   & 870 & \;26 & \;2 \\
\midrule
\multicolumn{4}{c}{\vspace{-0.5em}} \\
\toprule
\multicolumn{4}{c}{\textbf{3. Message apply phase (ms)}} \\
\midrule
Method & RV generation & Param update & Coordinate update \\
\midrule
MeZO   & 223.2 & 1189 & -- \\
Ours   & \;2.4 & \;4.8 & 10.48 \\
\bottomrule
\end{tabular}

\vspace{-1em}
\end{table}

%% file: sections/appendix-theory.tex
\providecommand{\add}[1]{{#1}}
\providecommand{\addb}[1]{{#1}}

\newcommand{\R}{\mathbb{R}}
\newcommand{\E}{\mathbb{E}}
\newcommand{\fnorm}[1]{\|#1\|_F}
\newcommand{\norm}[1]{\|#1\|}
\newcommand{\ip}[2]{\langle #1,\, #2\rangle}

% (The paragraph that follows, "For convenience, we restate this in the
%  explicit form used in the proofs: ...", stays unchanged.)

\section{Convergence Analysis of the Subspace Coordinate-Wise Estimator}
\label{appendix:subspace_converge}

In this section, we analyze subspace coordinate wise gradient estimator through a period-wise subspace optimization view.
The main idea is to reinterpret this estimator as solving, over each outer period, a lower-dimensional stochastic zeroth-order subproblem induced by a fixed low-rank subspace.
This avoids handling directly the dependence between the iterates and the reused subspace.

\paragraph{Actual update used in the analysis.}
We consider the following concrete form of the subspace coordinate-wise estimator.
At the beginning of outer period $k$, a pair of low-rank basis matrices
\[
U_k \in \R^{d_1 \times r_1}, \qquad V_k \in \R^{d_2 \times r_2}
\]
is sampled according to Assumption~\ref{assump:subspace} and fixed for the next $\tau$ inner steps.
For each inner step $s = 0,1,\dots,\tau-1$, we sample a coefficient-space random direction
\[
C_{k,s} \in \R^{r_1 \times r_2},
\]
form the full-space perturbation direction
\begin{equation}
D_{k,s} := U_k C_{k,s} V_k^{\top},
\label{eq:full_direction}
\end{equation}
and compute the two-point zeroth-order coefficient
\begin{equation}
a_{k,s}
:=
\frac{F(X_{k,s} + \epsilon D_{k,s};\xi_{k,s}) - F(X_{k,s} - \epsilon D_{k,s};\xi_{k,s})}{2\epsilon}.
\label{eq:coefficient}
\end{equation}
Then the actual update is
\begin{equation}
X_{k,s+1} = X_{k,s} - \eta \, a_{k,s} D_{k,s},
\qquad s=0,1,\dots,\tau-1,
\label{eq:actual_update}
\end{equation}
where $X_{k,0} = X_{k\tau}$ and $X_{k+1,0} = X_{k,\tau}$.

\paragraph{Objective.}
We consider the stochastic optimization problem
\begin{equation}
\min_{X \in \R^{d_1 \times d_2}} f(X) := \E_{\xi}[F(X;\xi)].
\label{eq:problem}
\end{equation}
We present the proof in matrix form, using the Frobenius norm throughout.

\subsection{Period-wise subspace formulation}

Let
\begin{equation}
X^{(k)} := X_{k,0} = X_{k\tau}
\label{eq:period_start}
\end{equation}
denote the starting point of period $k$.
Define the subspace objective
\begin{equation}
g_k(B) := f\bigl(X^{(k)} + U_k B V_k^{\top}\bigr),
\qquad B \in \R^{r_1 \times r_2}.
\label{eq:subproblem}
\end{equation}
The point of this definition is that, once $U_k$ and $V_k$ are fixed, the algorithm only moves inside the affine subspace
\[
\{X^{(k)} + U_k B V_k^{\top} : B \in \R^{r_1 \times r_2}\}.
\]

We now define the coefficient-space iterates by
\begin{equation}
B_{k,0} := 0,
\label{eq:B_init}
\end{equation}
and
\begin{equation}
B_{k,s+1} = B_{k,s} - \eta \, a_{k,s} C_{k,s},
\qquad s=0,1,\dots,\tau-1.
\label{eq:B_update}
\end{equation}

The next proposition shows that the full-space recursion \eqref{eq:actual_update} is exactly equivalent to the lower-dimensional recursion \eqref{eq:B_update}.

\begin{proposition}[Period-wise subspace representation]
\label{prop:equivalence}
Suppose that $U_k$ and $V_k$ are fixed during period $k$.
Let $B_{k,s}$ be generated by \eqref{eq:B_init}--\eqref{eq:B_update}.
Then for every $s=0,1,\dots,\tau$,
\begin{equation}
X_{k,s} = X^{(k)} + U_k B_{k,s} V_k^{\top}.
\label{eq:X_B_relation}
\end{equation}
\end{proposition}

\begin{proof}
We prove the claim by induction on $s$.
For $s=0$, the claim follows from the definitions $X_{k,0}=X^{(k)}$ and $B_{k,0}=0$.
Now suppose that
\[
X_{k,s} = X^{(k)} + U_k B_{k,s} V_k^{\top}
\]
holds for some $s \in \{0,1,\dots,\tau-1\}$.
Using the actual update \eqref{eq:actual_update} and the definition of $D_{k,s}$ in \eqref{eq:full_direction}, we obtain
\[
X_{k,s+1}
=
X_{k,s} - \eta a_{k,s} U_k C_{k,s} V_k^{\top}.
\]
Substituting the induction hypothesis gives
\[
X_{k,s+1}
=
X^{(k)} + U_k B_{k,s} V_k^{\top} - \eta a_{k,s} U_k C_{k,s} V_k^{\top}.
\]
Factoring out $U_k$ and $V_k^{\top}$ yields
\[
X_{k,s+1}
=
X^{(k)} + U_k\bigl(B_{k,s} - \eta a_{k,s} C_{k,s}\bigr)V_k^{\top}.
\]
By the coefficient-space update \eqref{eq:B_update}, the term in parentheses is precisely $B_{k,s+1}$.
Hence
\[
X_{k,s+1} = X^{(k)} + U_k B_{k,s+1} V_k^{\top}.
\]
This closes the induction.
\end{proof}

By Proposition~\ref{prop:equivalence}, the period-$k$ analysis reduces to the analysis of a stochastic ZO method on the lower-dimensional problem \eqref{eq:subproblem}.

\subsection{Basic identities for the subspace problem}

We first record the gradient of the subspace objective.

\begin{lemma}[Gradient of the subspace objective]
\label{lem:grad_g}
For the function $g_k(B) = f(X^{(k)} + U_k B V_k^{\top})$, the gradient is
\begin{equation}
\nabla g_k(B) = U_k^{\top} \nabla f\bigl(X^{(k)} + U_k B V_k^{\top}\bigr) V_k.
\label{eq:grad_g}
\end{equation}
In particular,
\begin{equation}
\nabla g_k(0) = U_k^{\top} \nabla f(X^{(k)}) V_k.
\label{eq:grad_g_zero}
\end{equation}
\end{lemma}

\begin{proof}
Let $H \in \R^{r_1 \times r_2}$ be arbitrary.
By the chain rule,
\begin{align*}
\ip{\nabla g_k(B)}{H}
&=
\left.\frac{d}{dt} g_k(B+tH)\right|_{t=0}
\\
&=
\left.\frac{d}{dt} f\bigl(X^{(k)} + U_k(B+tH)V_k^{\top}\bigr)\right|_{t=0}
\\
&=
\ip{\nabla f(X^{(k)} + U_k B V_k^{\top})}{U_k H V_k^{\top}}.
\end{align*}
Using the Frobenius inner product identity
\[
\ip{G}{UHV^{\top}} = \ip{U^{\top} G V}{H},
\]
we obtain
\[
\ip{\nabla g_k(B)}{H}
=
\ip{U_k^{\top} \nabla f(X^{(k)} + U_k B V_k^{\top}) V_k}{H}.
\]
Since this holds for all $H$, equation \eqref{eq:grad_g} follows.
Setting $B=0$ yields \eqref{eq:grad_g_zero}.
\end{proof}

We next show that the smoothness of $f$ induces smoothness of $g_k$.

\begin{restatable}[Smoothness and bounded minibatch noise]{assumption}{AssumpSmoothness}
\label{assump:smoothness}
\addb{$f$ is bounded below by $f^\star$, and every realization $F(\cdot;\xi)$ is
$L$-smooth: $\|\nabla F(X;\xi)-\nabla F(Y;\xi)\|_F \le L\|X-Y\|_F$ (hence
$f$ is $L$-smooth). Minibatch stochastic gradients of size $b$ are
unbiased with variance at most $\sigma^2/b$.}
\end{restatable}

For convenience, we restate this in the explicit form used in the proofs:
$L$-smoothness reads
\begin{equation}
\fnorm{\nabla f(X) - \nabla f(Y)} \le L \fnorm{X-Y},
\qquad \forall X,Y,
\label{eq:L_smooth}
\end{equation}
and unbiased mini-batch stochastic gradients with bounded variance read
\begin{equation}
\E_{\xi}[\nabla F(X;\xi)] = \nabla f(X),
\qquad
\E_{\xi}\fnorm{\nabla F(X;\xi) - \nabla f(X)}^2 \le \frac{\sigma^2}{b},
\qquad \forall X,
\label{eq:variance_assump}
\end{equation}
where $\xi$ denotes a mini-batch of size $b$.
% ======================= [B1] BEGIN ========================================
\addb{In addition, Assumption~\ref{assump:smoothness} imposes
\emph{per-realization} smoothness: for every mini-batch $\xi$, the map
$X\mapsto F(X;\xi)$ is differentiable and}
\begin{equation}
{
\fnorm{\nabla F(X;\xi)-\nabla F(Y;\xi)} \le L\fnorm{X-Y},
\qquad \forall X,Y,}
\label{eq:per_realization_smooth}
\end{equation}
\addb{which implies \eqref{eq:L_smooth} by Jensen's inequality.
Per-realization smoothness is standard in convergence analyses of zeroth-order
methods
% TODO: add bib entry for Malladi et al. (MeZO).
\cite{MeZO}.}
% ======================= [B1] END ==========================================

\begin{restatable}[Sphere-normalized subspace sampling]{assumption}{AssumpSubspace}
\label{assump:subspace}
\addb{Columns of $U_k$ and $V_k$ are drawn i.i.d.\ from
$\mathrm{Unif}(\sqrt{d_1}\,\mathbb{S}^{d_1-1})$ and
$\mathrm{Unif}(\sqrt{d_2}\,\mathbb{S}^{d_2-1})$, independently across
periods and of all inner-loop randomness, and satisfy
$\|U_k\|_2\le\rho_U:=2\sqrt{d_1}$ and $\|V_k\|_2\le\rho_V:=2\sqrt{d_2}$;
the discussion below explains why this operator-norm bound is benign.}
\end{restatable}
% ======================= [E1'] BEGIN =======================================
% Formal interpretation of the operator-norm bound assumed in Assumption 2.

\add{We first make precise the ensemble underlying
Assumption~\ref{assump:subspace}: the columns of
$U_k=[u_{k,1},\ldots,u_{k,r_1}]$ and $V_k=[v_{k,1},\ldots,v_{k,r_2}]$ are
sampled independently as
$u_{k,a}\sim \mathrm{Unif}(\sqrt{d_1}\mathbb{S}^{d_1-1})$ and
$v_{k,b}\sim \mathrm{Unif}(\sqrt{d_2}\mathbb{S}^{d_2-1})$,
independently across periods and of $X^{(k)}$ and inner-loop randomness, and
the joint draw $(U_k,V_k)$ is retained only on the spectral event}
\begin{equation}
{
\mathcal{E}_k
:=
\bigl\{\norm{U_k}_2 \le 2\sqrt{d_1}\bigr\}
\cap
\bigl\{\norm{V_k}_2 \le 2\sqrt{d_2}\bigr\}}
\label{eq:spectral_event}
\end{equation}
\add{and redrawn otherwise, so that the operator-norm bounds
$\norm{U_k}_2\le\rho_U=2\sqrt{d_1}$ and $\norm{V_k}_2\le\rho_V=2\sqrt{d_2}$
of Assumption~\ref{assump:subspace} hold deterministically.
This conditioning is benign: whenever
$\max(r_1/d_1,\,r_2/d_2)\le 1/32$ and $\min(d_1,d_2)\ge 36$ (satisfied with
in all our experiments, where $r\le 64$ and $d_\ell \ge 2048$),
Lemma~\ref{lem:spectral_conc} below shows that $\mathcal{E}_k$ fails with
probability at most $(r_1{+}r_2{+}2)\,e^{-\min(d_1,d_2)/72}$, so the
conditioned ensemble is statistically indistinguishable from the unconditioned
one; moreover, the conditioning preserves the per-column isotropy
$\E[u_{k,a}u_{k,a}^{\top}]=I_{d_1}$, $\E[v_{k,b}v_{k,b}^{\top}]=I_{d_2}$
(Lemma~\ref{lem:cond_ensemble}).}
% ======================= [E1'] END =========================================

% ======================= [E12] BEGIN (NEW LEMMA) ===========================
{%
\begin{lemma}[Spectral concentration of the sphere-normalized ensemble]
\label{lem:spectral_conc}
Let $U\in\R^{d\times r}$ have columns drawn i.i.d.\ from
$\mathrm{Unif}(\sqrt d\,\mathbb{S}^{d-1})$, with $r/d\le 1/32$ and $d\ge 36$.
Then
\[
\Pr\bigl[\norm{U}_2 > 2\sqrt d\,\bigr]
\;\le\;
(r+1)\,e^{-d/72}.
\]
\end{lemma}
\begin{proof}
Write $u_a=\sqrt d\,g_a/\norm{g_a}$ with
$g_a\overset{\mathrm{iid}}{\sim}\mathcal N(0,I_d)$, so that
$U=\sqrt d\,G\,\mathrm{diag}(\norm{g_a}^{-1})$ and hence
$\norm{U}_2\le \sqrt d\,\norm{G}_2/\min_{a}\norm{g_a}$.
Two standard Gaussian facts:
$\Pr[\norm{G}_2>\sqrt d+\sqrt r+s]\le e^{-s^2/2}$, and, since
$g\mapsto\norm g$ is $1$-Lipschitz with $\E\norm{g_a}\ge\sqrt{d-1}$,
$\Pr[\norm{g_a}<\sqrt{d-1}-s]\le e^{-s^2/2}$.
Taking $s=\sqrt d/4$ in the former and $s=\sqrt d/6$ in the latter with a
union bound over $a\le r$, the event
\[
\norm{G}_2\le\bigl(\tfrac54+\sqrt{r/d}\bigr)\sqrt d,
\qquad
\min_{a\le r}\norm{g_a}\ge\tfrac45\sqrt d
\]
holds with probability at least $1-(r+1)e^{-d/72}$, where we used
$\sqrt{d-1}-\sqrt d/6\ge\tfrac45\sqrt d
\iff \sqrt{d-1}\ge\tfrac{29}{30}\sqrt d
\iff 59\,d\ge 900$,
valid for $d\ge 36$ (indeed for all $d\ge 16$).
On this event, since $\sqrt{r/d}\le 1/\sqrt{32}$,
\[
\norm{U}_2
\;\le\;
\frac{5/4+1/\sqrt{32}}{4/5}\,\sqrt d
\;<\;
2\sqrt d.
\]
\end{proof}
}% ===================== [E12] END ==========================================

% ======================= [E6'] BEGIN (NEW LEMMA) ===========================
{%
\begin{lemma}[Properties of the conditioned ensemble]
\label{lem:cond_ensemble}
Under Assumption~\ref{assump:subspace},
(i) every column satisfies
$\E[u_{k,a}u_{k,a}^{\top}]=I_{d_1}$ and
$\E[v_{k,b}v_{k,b}^{\top}]=I_{d_2}$, and
(ii) $U_k$ and $V_k$ are independent.
\end{lemma}
\begin{proof}
(ii) The event \eqref{eq:spectral_event} factorizes over $U_k$ and $V_k$, and
$U_k$ and $V_k$ are independent before conditioning, so they remain
independent after conditioning on the product event.
(i) The event \eqref{eq:spectral_event} is invariant under the map
$U_k\mapsto R\,U_k$ for any $R\in O(d_1)$, and so is the unconditioned law of
$U_k$; hence the conditional law of $U_k$ is rotation-invariant.
Consequently $\E[u_{k,a}u_{k,a}^{\top}]$ commutes with every orthogonal matrix
and equals $cI_{d_1}$ for some $c\ge 0$; since $\norm{u_{k,a}}^2=d_1$ almost
surely, taking traces gives $c=1$.
The same argument applies to $V_k$.
\end{proof}
}% ===================== [E6'] END ===========================================

% ======================= [E2'] BEGIN =======================================
\begin{remark}

\add{In the actual implementation, the entries of $U_k$ and $V_k$ are sampled
independently from $\mathcal N(0,1)$, as in \cite{LOZO}; since column norms
concentrate at $\sqrt{d_\ell}$, the sphere normalization above is the standard
idealization of this ensemble \cite{LOZO}.
The conditioning on \eqref{eq:spectral_event} is vacuous in practice: by
Lemma~\ref{lem:spectral_conc} its failure probability is exponentially small
in $d_\ell$ (negligible even after a union bound over all periods and layers),
and it can be realized by \emph{deterministic} rejection
sampling---increment the seed and redraw on failure---which preserves
seed-reconstructibility and never triggers in practice.
Since \eqref{eq:spectral_event} is rotation-invariant, the conditioning leaves
the per-column isotropy used in the analysis intact
(Lemma~\ref{lem:cond_ensemble}).}
\end{remark}
% ======================= [E2'] END =========================================

\begin{lemma}[Smoothness of the subspace objective]
\label{lem:smooth_g}
Under Assumptions~\ref{assump:smoothness} and \ref{assump:subspace}, the function $g_k$ is $\widetilde L$-smooth with
\begin{equation}
\widetilde L := L \rho_U^2 \rho_V^2.
\label{eq:L_tilde}
\end{equation}
% ======================= [E3] BEGIN ========================================
Under Assumption~\ref{assump:subspace}, this gives
\add{$\widetilde L = 16\,L d_1 d_2$}.
% ======================= [E3] END ==========================================
That is, for all $B_1,B_2 \in \R^{r_1 \times r_2}$,
\begin{equation}
\fnorm{\nabla g_k(B_1) - \nabla g_k(B_2)} \le \widetilde L \fnorm{B_1-B_2}.
\label{eq:smooth_g}
\end{equation}
\end{lemma}

\begin{proof}
Let $\Delta_{\nabla f} := \nabla f(X^{(k)}+U_k B_1 V_k^{\top}) - \nabla f(X^{(k)}+U_k B_2 V_k^{\top})$.
Using Lemma~\ref{lem:grad_g} and $\|AXB\|_F \le \|A\|_2\|X\|_F\|B\|_2$,
\begin{align*}
\fnorm{\nabla g_k(B_1) - \nabla g_k(B_2)}
&= \fnorm{U_k^{\top}\Delta_{\nabla f}\,V_k}
\le \rho_U \fnorm{\Delta_{\nabla f}} \rho_V
\\
&\le \rho_U \cdot L\fnorm{U_k(B_1-B_2)V_k^{\top}} \cdot \rho_V
\\
&\le L\rho_U^2\rho_V^2 \fnorm{B_1-B_2},
\end{align*}
where the last step uses $L$-smoothness of $f$ and $\fnorm{UAV^\top}\le\norm{U}_2\fnorm{A}\norm{V}_2$.
This proves \eqref{eq:smooth_g}.
\end{proof}

\subsection{Coefficient-space ZO estimator}

Define the lifted stochastic objective
\begin{equation}
G_k(B;\xi) := F(X^{(k)} + U_k B V_k^{\top};\xi).
\label{eq:Gk}
\end{equation}
Using Proposition~\ref{prop:equivalence}, the two-point coefficient \eqref{eq:coefficient} can be rewritten as
\begin{equation}
a_{k,s}
=
\frac{G_k(B_{k,s}+\epsilon C_{k,s};\xi_{k,s})
- G_k(B_{k,s}-\epsilon C_{k,s};\xi_{k,s})}{2\epsilon}.
\label{eq:coefficient_subspace}
\end{equation}
Therefore, the coefficient-space estimator is
\begin{equation}
\widehat{\nabla} g_k(B_{k,s};\xi_{k,s}) := a_{k,s} C_{k,s}.
\label{eq:estimator}
\end{equation}
In other words, the coefficient update \eqref{eq:B_update} is simply
\begin{equation}
B_{k,s+1} = B_{k,s} - \eta \, \widehat{\nabla} g_k(B_{k,s};\xi_{k,s}).
\label{eq:B_update_estimator}
\end{equation}

We now specialize the estimator to the rule we actually use: at each inner step, a single coordinate is sampled uniformly at random, and only that coordinate's two-point finite difference is computed.
This makes our method a \emph{randomized coordinate descent with a one-coordinate finite-difference estimator}, not full coordinate finite-difference gradient descent.

\begin{restatable}[Uniform coordinate sampling]{assumption}{AssumpCoordinate}
\label{assump:coordinate}
$J_{k,s}\sim\mathrm{Unif}\{1,\ldots,q\}$ independently of $\xi_{k,s}$; both
function evaluations in \eqref{eq:seedflood-update} use the same minibatch $\xi_{k,s}$.
\end{restatable}

We work in the period-wise coefficient space introduced above.
Let $q := r_1 r_2$, and let $\mathcal{E} = \{E_1,\ldots,E_q\}$ be the standard
coordinate basis of $\R^{r_1\times r_2}$.
The sampling rule of Assumption~\ref{assump:coordinate} can be written
\begin{equation}
C_{k,s} = E_{J_{k,s}},
\qquad
J_{k,s} \sim \mathrm{Unif}\{1,\ldots,q\},
\label{eq:coord_sampling}
\end{equation}
with $J_{k,s}$ independent of the mini-batch $\xi_{k,s}$ and the two function
values in the central difference sharing the same mini-batch.
For each coordinate $j \in \{1,\ldots,q\}$, we further define
\[
a_j(B;\xi)
:=
\frac{G_k(B+\epsilon E_j;\xi)-G_k(B-\epsilon E_j;\xi)}{2\epsilon},
\qquad
\bar a_j(B) := \E_{\xi}[a_j(B;\xi)],
\]
and the population coordinate finite-difference operator
\[
\nabla_{\epsilon} g_k(B) := \sum_{j=1}^{q} \bar a_j(B)\,E_j.
\]

\begin{lemma}[Standard CGE finite-difference property]
\label{lem:coord_bias}
Under Assumptions~\ref{assump:smoothness}--\ref{assump:coordinate}, the population
coordinate finite-difference operator satisfies
\begin{equation}
\nabla_{\epsilon} g_k(B) = \nabla g_k(B) + \delta_k(B),
\qquad
\fnorm{\delta_k(B)} \le c_\delta \epsilon,
\label{eq:coord_bias}
\end{equation}
with $c_\delta = \sqrt q\,\widetilde L/2$.
\end{lemma}

\begin{proof}
This is the standard central coordinate finite-difference property used in CGE
analyses.
Indeed, for each coordinate $j$, define $h_j(t):=g_k(B+tE_j)$.
Since $g_k$ is $\widetilde L$-smooth by Lemma~\ref{lem:smooth_g}, the derivative
$h_j'(t)=\ip{\nabla g_k(B+tE_j)}{E_j}$ is $\widetilde L$-Lipschitz.
Therefore,
\[
\left|
\frac{g_k(B+\epsilon E_j)-g_k(B-\epsilon E_j)}{2\epsilon}
-\ip{\nabla g_k(B)}{E_j}
\right|
\le
\frac{1}{2\epsilon}\int_{-\epsilon}^{\epsilon}
\widetilde L |t|\,dt
=
\frac{\widetilde L\epsilon}{2}.
\]
Since $\bar a_j(B)$ is the corresponding population central difference, summing
the squared coordinate errors over $j=1,\ldots,q$ gives
$\fnorm{\delta_k(B)}\le \sqrt q\,\widetilde L\epsilon/2$.
\end{proof}

{%
{The next lemma bounds the coordinate finite-difference noise.
Bounding the noise pathwise at the current iterate would pick up the
dimension factor $\rho_U^2\rho_V^2$ through the subspace-lifted variance;
anchoring the noise at the period start $X^{(k)}$, where the minibatch noise
is independent of the subspace draw, removes this factor from the
stochastic-noise constant.}
\begin{lemma}[Anchored coordinate finite-difference noise]
\label{lem:coord_noise_anchor}
Suppose Assumptions~\ref{assump:smoothness}--\ref{assump:coordinate} hold.
Fix a period $k$ and an inner step $s$, let
\[
\mathcal{F}_{k,s}
:=
\sigma\bigl(X^{(k)},U_k,V_k,\xi_{k,0},J_{k,0},\ldots,\xi_{k,s-1},J_{k,s-1}\bigr),
\]
let $B$ be an $\mathcal{F}_{k,s}$-measurable random matrix in
$\R^{r_1\times r_2}$, and let $\xi=\xi_{k,s}$ be the fresh minibatch of step
$s$, independent of $\mathcal{F}_{k,s}$.
Then
\begin{equation}
\frac1q\sum_{j=1}^{q}
\E\bigl[\bigl(a_j(B;\xi)-\bar a_j(B)\bigr)^2 \,\big|\, X^{(k)}\bigr]
\;\le\;
\frac{3\sigma^2}{b}
+\frac{12\widetilde L^2}{q}\,\E\bigl[\fnorm{B}^2 \,\big|\, X^{(k)}\bigr]
+12\,\widetilde L^2\epsilon^2,
\label{eq:coord_noise_anchor}
\end{equation}
where $\bar a_j(B):=\E[a_j(B;\xi)\mid\mathcal{F}_{k,s}]$.
\end{lemma}

\begin{proof}
Write $\Delta(X;\xi):=\nabla F(X;\xi)-\nabla f(X)$,
let $E_j=e_ae_b^{\top}$ denote the coordinate pair $(a,b)$ corresponding to
$j$, so that $\ip{U_k^{\top}MV_k}{E_j}=u_{k,a}^{\top}Mv_{k,b}$ for any
$M\in\R^{d_1\times d_2}$, and set
$X^{(j)}(t):=X^{(k)}+U_k(B+tE_j)V_k^{\top}$ for $t\in[-\epsilon,\epsilon]$.
Since $B$ is $\mathcal{F}_{k,s}$-measurable and $\xi$ is independent of
$\mathcal{F}_{k,s}$, we have, pathwise on $\mathcal{F}_{k,s}$,
$\bar a_j(B)=\bigl(g_k(B+\epsilon E_j)-g_k(B-\epsilon E_j)\bigr)/(2\epsilon)$.
Applying the fundamental theorem of calculus to
$t\mapsto G_k(B+tE_j;\xi)$ and $t\mapsto g_k(B+tE_j)$, whose derivatives are
$\ip{U_k^{\top}\nabla F(X^{(j)}(t);\xi)V_k}{E_j}$ and
$\ip{U_k^{\top}\nabla f(X^{(j)}(t))V_k}{E_j}$ by the chain rule of
Lemma~\ref{lem:grad_g}

\begin{equation}
a_j(B;\xi)-\bar a_j(B)
=
\frac{1}{2\epsilon}
\int_{-\epsilon}^{\epsilon}
u_{k,a}^{\top}\,\Delta\bigl(X^{(j)}(t);\xi\bigr)\,v_{k,b}\,dt.
\label{eq:anchor_ftc}
\end{equation}

\emph{Step 1: three-term anchor decomposition.}
Let $X_B:=X^{(k)}+U_kBV_k^{\top}$ and split the noise along the integration
path into an anchor term at the period start, a drift term \emph{common to all
coordinates}, and a coordinate-specific perturbation residual:
\[
\Delta\bigl(X^{(j)}(t);\xi\bigr)
=\underbrace{\Delta(X^{(k)};\xi)}_{=:\,\Delta_0}
+\underbrace{\Delta(X_B;\xi)-\Delta(X^{(k)};\xi)}_{=:\,\Gamma}
+\underbrace{\Delta\bigl(X^{(j)}(t);\xi\bigr)-\Delta(X_B;\xi)}_{=:\,S^{(j)}_t}.
\]
Only $S^{(j)}_t$ depends on the coordinate $j$ and on $t$.
By per-realization smoothness \eqref{eq:per_realization_smooth} and
$L$-smoothness of $f$ \eqref{eq:L_smooth},
\[
\fnorm{\Gamma}\le 2L\fnorm{X_B-X^{(k)}}\le 2L\rho_U\rho_V\fnorm{B},
\qquad
\fnorm{S^{(j)}_t}\le 2L\fnorm{U_k\,tE_j\,V_k^{\top}}\le 2L\rho_U\rho_V\,\epsilon
\]
for all $|t|\le\epsilon$.
Substituting into \eqref{eq:anchor_ftc},
\[
a_j(B;\xi)-\bar a_j(B)=A_j+D_j+C_j,
\quad
A_j:=u_{k,a}^{\top}\Delta_0v_{k,b},
\quad
D_j:=u_{k,a}^{\top}\Gamma\,v_{k,b},
\quad
C_j:=\frac{1}{2\epsilon}\int_{-\epsilon}^{\epsilon}
u_{k,a}^{\top}S^{(j)}_tv_{k,b}\,dt,
\]
and $(A_j+D_j+C_j)^2\le 3A_j^2+3D_j^2+3C_j^2$.

\emph{Step 2: anchor term (isotropy).}
The term $A_j$ involves neither $B$ nor any past inner-loop randomness:
$\Delta_0$ is a function of $(X^{(k)},\xi)$ only, and $(U_k,V_k)$ is
independent of $(X^{(k)},\xi)$ by Assumption~\ref{assump:subspace}.
Hence, conditionally on $(X^{(k)},\xi)$, the pair $(U_k,V_k)$ retains the law
of Assumption~\ref{assump:subspace}, and Lemma~\ref{lem:cond_ensemble}
(per-column isotropy and $U_k\perp V_k$) gives, exactly as in the single-pair
computation of Lemma~\ref{lem:subspace_coverage},
\[
\E_{U_k,V_k}\bigl[(u_{k,a}^{\top}\Delta_0 v_{k,b})^2\bigr]
=
\E\bigl[u_{k,a}^{\top}\Delta_0\,\E[v_{k,b}v_{k,b}^{\top}]\,\Delta_0^{\top}u_{k,a}\bigr]
=
\mathrm{tr}\bigl(\Delta_0\Delta_0^{\top}\bigr)
=
\fnorm{\Delta_0}^2 .
\]
Taking the expectation over $\xi$ and using \eqref{eq:variance_assump} at
$X=X^{(k)}$,
\begin{equation}
\E\bigl[A_j^2 \,\big|\, X^{(k)}\bigr]
=
\E_{\xi}\bigl[\fnorm{\Delta_0}^2 \,\big|\, X^{(k)}\bigr]
\le
\frac{\sigma^2}{b}
\qquad\text{for every } j .
\label{eq:anchor_isotropy}
\end{equation}

\emph{Step 3: drift term (Frobenius aggregation over coordinates).}
Since $\Gamma$ is common to all coordinates, summing the squared projections over
the full coordinate basis collapses to a Frobenius norm, pointwise in
$(U_k,V_k,B,\xi)$:
\begin{equation}
\frac1q\sum_{j=1}^{q}D_j^2
=\frac1q\fnorm{U_k^{\top}\Gamma V_k}^2
\le\frac{\rho_U^2\rho_V^2}{q}\fnorm{\Gamma}^2
\le\frac{4L^2\rho_U^4\rho_V^4}{q}\fnorm{B}^2
=\frac{4\widetilde L^2}{q}\fnorm{B}^2 .
\label{eq:drift_aggregation}
\end{equation}

\emph{Step 4: perturbation residual.}
For every $j$ and $|t|\le\epsilon$,
$|u_{k,a}^{\top}S^{(j)}_tv_{k,b}|
\le\rho_U\rho_V\fnorm{S^{(j)}_t}
\le 2L\rho_U^2\rho_V^2\,\epsilon
=2\widetilde L\epsilon$,
hence $|C_j|\le 2\widetilde L\epsilon$ and
$\frac1q\sum_{j}C_j^2\le 4\widetilde L^2\epsilon^2$.

\emph{Step 5: combining.}
Averaging $(A_j+D_j+C_j)^2\le 3A_j^2+3D_j^2+3C_j^2$ over
$j\in\{1,\ldots,q\}$, taking $\E[\,\cdot\mid X^{(k)}]$, and inserting
\eqref{eq:anchor_isotropy}, \eqref{eq:drift_aggregation}, and Step~4 yields
\eqref{eq:coord_noise_anchor}.
\end{proof}

\begin{remark}
The coordinate-averaged form of \eqref{eq:coord_noise_anchor} is essential:
it is the aggregation over the full coordinate basis in
\eqref{eq:drift_aggregation} that yields the drift correction
$\widetilde L^2\,\E\fnorm{B}^2/q$, placing it at the same order in $q$ as the
signal terms of the inner-loop recursion (Lemma~\ref{lem:drift}), where it is
absorbed by the Gr\"onwall argument with only a change of absolute constants.
The discretization correction $12\widetilde L^2\epsilon^2$ is of the same
scale as the finite-difference bias term
$(2c_\delta^2/q)\epsilon^2=(\widetilde L^2/2)\epsilon^2$ already present in
Lemma~\ref{lem:coord_estimator_moments}.
\end{remark}
}% ===================== [B1] END ===========================================

\begin{lemma}[Mean and second moment of the coordinate estimator]
\label{lem:coord_estimator_moments}
Under Assumptions~\ref{assump:smoothness}--\ref{assump:coordinate}, the actual one-step
estimator we use,
\begin{equation}
\widehat{\nabla} g_k(B;\xi,J) := a_J(B;\xi)\,E_J,
\label{eq:coord_estimator}
\end{equation}
satisfies
\begin{align}
\E\bigl[\widehat{\nabla} g_k(B;\xi,J) \mid X^{(k)},U_k,V_k,B\bigr]
&=
\frac{1}{q}\bigl(\nabla g_k(B)+\delta_k(B)\bigr).
\label{eq:coord_est_bias}
\end{align}
\addb{Moreover, if $B$ is $\mathcal{F}_{k,s}$-measurable and
$(\xi,J)=(\xi_{k,s},J_{k,s})$ is the fresh randomness of step $s$, as in
Lemma~\ref{lem:coord_noise_anchor}, then}
\begin{align}
\E\bigl[\fnorm{\widehat{\nabla} g_k(B;\xi,J)}^2 \mid X^{(k)}\bigr]
&\le
\frac{2}{q}\,\E\bigl[\fnorm{\nabla g_k(B)}^2 \mid X^{(k)}\bigr]
+\frac{2c_\delta^2}{q}\epsilon^2
+\frac{3\sigma^2}{b}
+\frac{12\widetilde L^2}{q}\,\E\bigl[\fnorm{B}^2 \mid X^{(k)}\bigr]
+12\widetilde L^2\epsilon^2 .
\label{eq:coord_second_moment}
\end{align}
\end{lemma}

\begin{proof}
Conditioning on $X^{(k)},U_k,V_k,B$ and averaging over the uniform coordinate
$J$,
\[
\E[\widehat{\nabla} g_k(B;\xi,J)]
=
\frac{1}{q}\sum_{j=1}^q \bar a_j(B)E_j
=
\frac{1}{q}\nabla_\epsilon g_k(B).
\]
Applying Lemma~\ref{lem:coord_bias} gives \eqref{eq:coord_est_bias}.
{%
For the second moment, conditionally on $\mathcal{F}_{k,s}$ (averaging over
the fresh $(\xi,J)$ only),
\[
\E\bigl[\fnorm{\widehat{\nabla} g_k(B;\xi,J)}^2 \bigm| \mathcal{F}_{k,s}\bigr]
=
\frac{1}{q}\sum_{j=1}^q \E_\xi[a_j(B;\xi)^2]
=
\frac{1}{q}\fnorm{\nabla_\epsilon g_k(B)}^2
+
\frac{1}{q}\sum_{j=1}^q
\E_\xi[(a_j(B;\xi)-\bar a_j(B))^2].
\]
Using $\nabla_\epsilon g_k(B)=\nabla g_k(B)+\delta_k(B)$,
$\fnorm{M+M'}^2\le 2\fnorm{M}^2+2\fnorm{M'}^2$, and
$\fnorm{\delta_k(B)}\le c_\delta\epsilon$, the first term is at most
$\frac{2}{q}\fnorm{\nabla g_k(B)}^2+\frac{2c_\delta^2}{q}\epsilon^2$,
pointwise on $\mathcal{F}_{k,s}$.
Taking $\E[\,\cdot\mid X^{(k)}]$ of both sides via the tower property and
applying Lemma~\ref{lem:coord_noise_anchor} to the averaged noise term
yields \eqref{eq:coord_second_moment}.}
\end{proof}

\subsection{Inner-loop drift bound}

We now control how far the coefficient iterate $B_{k,s}$ moves from the period start.
Compared with a full-gradient estimator, the one-coordinate estimator only touches one of the $q$ coordinates at each step, contributing a $1/q$ factor to both the conditional mean and the second moment.
This is reflected in the bound through the quantity
\begin{equation}
\Lambda_q(\tau)
:=
\frac{\tau}{q}\!\left(1+\frac{\tau}{q}\right).
\label{eq:lambda_q}
\end{equation}

\begin{lemma}[Inner-loop drift for the one-coordinate estimator]
\label{lem:drift}
Under Assumptions~\ref{assump:smoothness}--\ref{assump:coordinate}, \addb{there exist absolute constants $C_1,C_2,C_3>0$ such that, for every $s \le \tau$,}
\begin{equation}
{
\E\bigl[\fnorm{B_{k,s}}^2 \mid X^{(k)}\bigr]
\le
C_1 \eta^2 q\,\Lambda_q(\tau) \fnorm{\nabla f(X^{(k)})}^2
+
C_2 \eta^2 \frac{\tau}{b} \sigma^2
+
C_3 \eta^2 \bigl(c_\delta^2\,\Lambda_q(\tau) + 6\widetilde L^2\tau\bigr) \epsilon^2,}
\label{eq:drift_bound}
\end{equation}
provided
\begin{equation}
{
\eta \;\le\; \frac{1}{\widetilde L\,\sqrt{32\,\Lambda_q(\tau)}}.}
\label{eq:eta_gronwall}
\end{equation}
\end{lemma}

\begin{proof}
Starting from \eqref{eq:B_update_estimator} and $B_{k,0}=0$,
\begin{equation}
B_{k,s} = -\eta \sum_{j=0}^{s-1} \widehat{\nabla} g_k(B_{k,j};\xi_{k,j},J_{k,j}).
\label{eq:B_expand}
\end{equation}
Let $\mathcal{F}_j$ be the $\sigma$-algebra generated by $(X^{(k)},U_k,V_k,\xi_{k,0},J_{k,0},\dots,\xi_{k,j-1},J_{k,j-1})$.
Decompose each estimator as
\[
\widehat{\nabla} g_k(B_{k,j};\xi_{k,j},J_{k,j}) = \bar{Z}_j + N_j,
\qquad
\bar{Z}_j := \frac{1}{q}\bigl(\nabla g_k(B_{k,j}) + \delta_k(B_{k,j})\bigr),
\]
where $\bar{Z}_j$ is the $\mathcal{F}_j$-conditional mean (by \eqref{eq:coord_est_bias}) and $N_j$ is the centered noise, which is a martingale-difference sequence with respect to $\{\mathcal{F}_j\}$.
\addb{Throughout the rest of the proof, plain expectation signs abbreviate
$\E[\,\cdot \mid X^{(k)}]$.}
Using \addb{$\fnorm{M+M'}^2 \le 2\fnorm{M}^2+2\fnorm{M'}^2$},
\begin{equation}
{\E\bigl[\fnorm{B_{k,s}}^2 \mid X^{(k)}\bigr]}
\le
2\eta^2 \underbrace{\E\!\left[\fnorm{\sum_{j=0}^{s-1}\bar{Z}_j}^2\right]}_{\text{signal}}
+
2\eta^2 \underbrace{\E\!\left[\fnorm{\sum_{j=0}^{s-1}N_j}^2\right]}_{\text{noise}}.
\label{eq:bv_split}
\end{equation}

\textbf{Signal part.}
By Cauchy--Schwarz and $\fnorm{\bar{Z}_j} \le \tfrac{1}{q}\bigl(\fnorm{\nabla g_k(B_{k,j})}+c_\delta\epsilon\bigr)$,
\begin{equation}
\fnorm{\sum_{j=0}^{s-1}\bar{Z}_j}^2
\le
\frac{2s}{q^2}\sum_{j=0}^{s-1}\bigl(\fnorm{\nabla g_k(B_{k,j})}^2 + c_\delta^2\epsilon^2\bigr).
\label{eq:signal_cs}
\end{equation}
By $\widetilde{L}$-smoothness of $g_k$ (Lemma~\ref{lem:smooth_g}),
\begin{equation}
\fnorm{\nabla g_k(B_{k,j})}^2
\le
2\fnorm{\nabla g_k(0)}^2 + 2\widetilde L^2 \fnorm{B_{k,j}}^2.
\label{eq:grad_expand}
\end{equation}
\addb{Both \eqref{eq:signal_cs} and \eqref{eq:grad_expand} hold pointwise.
Substituting, using $s\le\tau$, taking $\E[\,\cdot\mid X^{(k)}]$, and applying
Lemma~\ref{lem:subspace_coverage} to the $X^{(k)}$-measurable matrix
$\nabla f(X^{(k)})$, i.e.\
$\E\bigl[\fnorm{\nabla g_k(0)}^2 \mid X^{(k)}\bigr]
=\E\bigl[\fnorm{U_k^{\top}\nabla f(X^{(k)})V_k}^2 \mid X^{(k)}\bigr]
=q\fnorm{\nabla f(X^{(k)})}^2$,}
\begin{align}
2\eta^2\E\!\left[\fnorm{\sum_{j<s}\bar{Z}_j}^2\right]
&\le
\frac{8\tau^2\eta^2}{q}\fnorm{\nabla f(X^{(k)})}^2
+ \frac{8\tau\eta^2\widetilde L^2}{q^2}\sum_{j=0}^{s-1}\E\bigl[\fnorm{B_{k,j}}^2\bigr]
\nonumber\\
&\quad
+ \frac{4c_\delta^2\tau^2\eta^2}{q^2}\epsilon^2.
\label{eq:signal_bound}
\end{align}

\textbf{Noise part.}
Since $\{N_j\}$ is a martingale-difference sequence, the cross terms
vanish\addb{; indeed, for $i<j$, towering through
$\mathcal{F}_j\supset\sigma(X^{(k)})$ gives
$\E\bigl[\ip{N_i}{N_j}\bigm| X^{(k)}\bigr]
=\E\bigl[\ip{N_i}{\E[N_j\mid\mathcal{F}_j]}\bigm| X^{(k)}\bigr]=0$, so the
orthogonality survives the coarser conditioning}:
\begin{equation}
\E\!\left[\fnorm{\sum_{j=0}^{s-1}N_j}^2\right]
= \sum_{j=0}^{s-1}\E\bigl[\fnorm{N_j}^2\bigr]
\le \sum_{j=0}^{s-1}\E\bigl[\fnorm{\widehat{\nabla} g_k(B_{k,j};\xi_{k,j},J_{k,j})}^2\bigr].
\label{eq:noise_isometry}
\end{equation}
\addb{Applying \eqref{eq:coord_second_moment} to each term (with $B=B_{k,j}$,
which is $\mathcal{F}_{k,j}$-measurable and faces the fresh randomness
$(\xi_{k,j},J_{k,j})$), then \eqref{eq:grad_expand} and
Lemma~\ref{lem:subspace_coverage} as in the signal part,}
\begin{align}
2\eta^2\E\!\left[\fnorm{\sum_{j<s}N_j}^2\right]
&\le
8\tau\eta^2\,\fnorm{\nabla f(X^{(k)})}^2
+ \frac{32\,\eta^2\widetilde L^2}{q}\sum_{j=0}^{s-1}\E\bigl[\fnorm{B_{k,j}}^2\bigr]
\nonumber\\
&\quad
+ \frac{4c_\delta^2\tau\eta^2}{q}\epsilon^2
+ \frac{6\tau\eta^2}{b}\sigma^2
+ 24\,\tau\eta^2\widetilde L^2\epsilon^2 .
\label{eq:noise_bound}
\end{align}

\textbf{Combining.}
Adding \eqref{eq:signal_bound} and \eqref{eq:noise_bound},
\begin{align}
\E\bigl[\fnorm{B_{k,s}}^2 \mid X^{(k)}\bigr]
&\le
8\eta^2\!\left(\frac{\tau^2}{q}+\tau\right)\!\fnorm{\nabla f(X^{(k)})}^2
+ \frac{6\tau\eta^2}{b}\sigma^2
\nonumber\\
&\quad
+ 4c_\delta^2\eta^2\!\left(\frac{\tau^2}{q^2}+\frac{\tau}{q}\right)\!\epsilon^2
+ 24\,\tau\eta^2\widetilde L^2\epsilon^2
\nonumber\\
&\quad
+ 8\eta^2\widetilde L^2\!\left(\frac{\tau}{q^2}+\frac{4}{q}\right)\!\sum_{j=0}^{s-1}\E\bigl[\fnorm{B_{k,j}}^2\bigr].
\label{eq:drift_recursion}
\end{align}
\addb{Note $\tfrac{\tau^2}{q}+\tau = q\,\Lambda_q(\tau)$ and
$\tfrac{\tau^2}{q^2}+\tfrac{\tau}{q} = \Lambda_q(\tau)$.
The stochastic noise now enters at the dimension-free level $\sigma^2/b$ and
accumulates linearly in $\tau$ because the centered noise terms form a
martingale-difference sequence; the drift correction inherited from
Lemma~\ref{lem:coord_noise_anchor} enters the recursive coefficient at the
same $1/q$ order as the signal terms.}

\medskip\noindent\textbf{Gr\"{o}nwall step.}
\addb{Write $\phi_s := \E\bigl[\fnorm{B_{k,s}}^2 \mid X^{(k)}\bigr]$.}
The recursion \eqref{eq:drift_recursion} takes the form
\[
\phi_s \;\le\; \alpha + \beta\sum_{j=0}^{s-1}\phi_j,
\]
where
\begin{align*}
\alpha &:= 8\eta^2 q\,\Lambda_q(\tau)\fnorm{\nabla f(X^{(k)})}^2
         + \frac{6\tau\eta^2}{b}\sigma^2
         + 4\eta^2\bigl(c_\delta^2\,\Lambda_q(\tau)+6\widetilde L^2\tau\bigr)\epsilon^2,\\
\beta  &:= 8\eta^2\widetilde{L}^2\!\left(\frac{\tau}{q^2}+\frac{4}{q}\right).
\end{align*}
\addb{Since
$\beta\tau = 8\eta^2\widetilde L^2\bigl(\tfrac{\tau^2}{q^2}+\tfrac{4\tau}{q}\bigr)
\le 32\,\eta^2\widetilde L^2\Lambda_q(\tau)$,
condition \eqref{eq:eta_gronwall} implies $\beta\tau \le 1$.}
By the discrete Gr\"{o}nwall inequality, for all $s\le\tau$,
\[
\phi_s \;\le\; \alpha(1+\beta)^s \;\le\; \alpha(1+\beta)^\tau \;\le\; e^{\,\beta\tau}\alpha \;\le\; e\,\alpha.
\]
Setting
\begin{equation}
C_1 := 8e,\qquad C_2 := 6e,\qquad C_3 := 4e,
\label{eq:drift_constants}
\end{equation}
we obtain \eqref{eq:drift_bound}.
\end{proof}

\subsection{One-period descent}

The next lemma proves that one full period decreases the objective by a quantity proportional to the projected gradient norm, up to higher-order drift and stochastic error terms.

\begin{lemma}[One-period descent]
\label{lem:one_period}
Under Assumptions~\ref{assump:smoothness}--\ref{assump:coordinate}, there exist absolute constants $c_1,c_2,c_3>0$ and a constant $c_4>0$ depending only on $c_\delta$ such that
\begin{align}
\E\bigl[f(X^{(k+1)}) - f(X^{(k)}) \mid X^{(k)}\bigr]
&\le
-\!\left(c_1 - c_2\eta\widetilde L\!\left(1+\frac{\tau}{q}\right)\!\right)\!\eta\tau\,\fnorm{\nabla f(X^{(k)})}^2
\nonumber\\
&\quad
+ c_3 \widetilde L \,\frac{\eta^2 \tau}{b}\,\sigma^2
+ c_4\!\left(\frac{\eta\tau}{q} + \widetilde L\eta^2\Lambda_q(\tau)\right)\!\epsilon^2.
\label{eq:one_period_descent}
\end{align}
provided \eqref{eq:eta_gronwall} holds.
\end{lemma}

\begin{proof}
By Proposition~\ref{prop:equivalence}, the final point of period $k$ satisfies
\[
X^{(k+1)} = X^{(k)} + U_k B_{k,\tau} V_k^{\top}.
\]
Therefore,
\begin{equation}
f(X^{(k+1)}) - f(X^{(k)}) = g_k(B_{k,\tau}) - g_k(0).
\label{eq:objective_subspace}
\end{equation}
By the $\widetilde L$-smoothness of $g_k$,
\begin{equation}
g_k(B_{k,\tau}) - g_k(0)
\le
\ip{\nabla g_k(0)}{B_{k,\tau}} + \frac{\widetilde L}{2}\fnorm{B_{k,\tau}}^2.
\label{eq:smooth_descent_g}
\end{equation}
We now analyze the two terms on the right-hand side.

\paragraph{Step 1: the linear term.}
Using \eqref{eq:B_expand},
\begin{equation}
\ip{\nabla g_k(0)}{B_{k,\tau}}
=
-\eta \sum_{s=0}^{\tau-1} \ip{\nabla g_k(0)}{\widehat{\nabla} g_k(B_{k,s};\xi_{k,s},J_{k,s})}.
\label{eq:linear_term_expand}
\end{equation}
\addb{Write $\E_k[\cdot] := \E[\cdot \mid X^{(k)}]$ for brevity.
Taking conditional expectation, towering through $\mathcal{F}_{k,s}$ (with
respect to which $\nabla g_k(0)$ is measurable), and using
\eqref{eq:coord_est_bias},}
\begin{align}
\E_k\bigl[\ip{\nabla g_k(0)}{B_{k,\tau}}\bigr]
&=
-\frac{\eta}{q} \sum_{s=0}^{\tau-1}
\E_k\bigl[\ip{\nabla g_k(0)}{\nabla g_k(B_{k,s}) + \delta_k(B_{k,s})}\bigr]
\nonumber\\
&=
-\frac{\eta\tau}{q}{\,\E_k\bigl[\fnorm{\nabla g_k(0)}^2\bigr]}
\nonumber\\
&\quad
-\frac{\eta}{q}\sum_{s=0}^{\tau-1}
\E_k\bigl[\ip{\nabla g_k(0)}{\nabla g_k(B_{k,s}) - \nabla g_k(0)}\bigr]
\nonumber\\
&\quad
-\frac{\eta}{q}\sum_{s=0}^{\tau-1}
\E_k\bigl[\ip{\nabla g_k(0)}{\delta_k(B_{k,s})}\bigr].
\label{eq:linear_term_split}
\end{align}
We bound the drift term using Cauchy--Schwarz and Lemma~\ref{lem:smooth_g}:
\begin{align}
\bigl|\ip{\nabla g_k(0)}{\nabla g_k(B_{k,s}) - \nabla g_k(0)}\bigr|
&\le
\widetilde L \, \fnorm{\nabla g_k(0)} \, \fnorm{B_{k,s}}.
\label{eq:drift_linear_bound}
\end{align}
Applying Young's inequality,
\begin{equation}
\widetilde L \, \fnorm{\nabla g_k(0)} \, \fnorm{B_{k,s}}
\le
\frac{1}{4}\fnorm{\nabla g_k(0)}^2 + \widetilde L^2 \fnorm{B_{k,s}}^2.
\label{eq:young_drift}
\end{equation}
Similarly, the bias term is bounded by
\begin{equation}
\bigl|\ip{\nabla g_k(0)}{\delta_k(B_{k,s})}\bigr|
\le
\frac{1}{4}\fnorm{\nabla g_k(0)}^2 + c_\delta^2 \epsilon^2.
\label{eq:bias_linear_bound}
\end{equation}
\addb{Bounding each inner-product term in \eqref{eq:linear_term_split}
pointwise by \eqref{eq:young_drift} and \eqref{eq:bias_linear_bound}, summing
over $s = 0,\dots,\tau-1$, and applying Lemma~\ref{lem:coverage} in the form
$\E_k\fnorm{\nabla g_k(0)}^2 = q\fnorm{\nabla f(X^{(k)})}^2$,}
\begin{align}
\E_k\bigl[\ip{\nabla g_k(0)}{B_{k,\tau}}\bigr]
&\le
-\eta\tau\fnorm{\nabla f(X^{(k)})}^2
+\frac{\eta\tau}{2}\fnorm{\nabla f(X^{(k)})}^2
+\frac{\eta\widetilde L^2}{q}\sum_{s=0}^{\tau-1}\E_k\bigl[\fnorm{B_{k,s}}^2\bigr]
\nonumber\\
&\quad
+\frac{\eta\tau}{q} c_\delta^2\epsilon^2
\nonumber\\
&=
-\frac{\eta\tau}{2}\fnorm{\nabla f(X^{(k)})}^2
+\frac{\eta\widetilde L^2}{q}\sum_{s=0}^{\tau-1}\E_k\bigl[\fnorm{B_{k,s}}^2\bigr]
+\frac{\eta\tau}{q} c_\delta^2\epsilon^2.
\label{eq:linear_before_drift}
\end{align}
Applying Lemma~\ref{lem:drift} to each index $s = 0,\dots,\tau-1$ and summing,
\begin{equation}
\sum_{s=0}^{\tau-1}\E_k\bigl[\fnorm{B_{k,s}}^2\bigr]
\;\le\;
\tau\!\left(C_1\eta^2 q\,\Lambda_q(\tau)\fnorm{\nabla f(X^{(k)})}^2 + C_2\eta^2\frac{\tau}{b}\sigma^2 + C_3\eta^2\bigl(c_\delta^2\Lambda_q(\tau)+6\widetilde L^2\tau\bigr)\epsilon^2\right)\!.
\label{eq:sum_drift}
\end{equation}
Substituting \eqref{eq:sum_drift} into \eqref{eq:linear_before_drift}, we obtain the Step~1 result
\begin{align}
\E_k\bigl[\ip{\nabla g_k(0)}{B_{k,\tau}}\bigr]
&\le
-\frac{\eta\tau}{2}\fnorm{\nabla f(X^{(k)})}^2
+ C_1\widetilde L^2\tau\eta^3\Lambda_q(\tau)\fnorm{\nabla f(X^{(k)})}^2
\nonumber\\
&\quad
+ C_2\widetilde L^2\frac{\tau^2\eta^3}{q b}\sigma^2
+ \!\left(C_3\widetilde L^2\frac{\tau\eta^3}{q}\bigl(c_\delta^2\Lambda_q(\tau)+6\widetilde L^2\tau\bigr) + c_\delta^2\frac{\eta\tau}{q}\right)\!\epsilon^2.
\label{eq:linear_term_expanded}
\end{align}

\paragraph{Step 2: the quadratic smoothness remainder.}
Applying Lemma~\ref{lem:drift} with $s=\tau$,
\begin{align}
\E_k\!\left[\frac{\widetilde L}{2}\fnorm{B_{k,\tau}}^2\right]
&\le
\frac{\widetilde L}{2}
\!\left(
C_1 \eta^2 q\,\Lambda_q(\tau) \fnorm{\nabla f(X^{(k)})}^2
+
C_2 \eta^2 \frac{\tau}{b}\sigma^2
+
C_3 \eta^2\bigl(c_\delta^2\Lambda_q(\tau)+6\widetilde L^2\tau\bigr) \epsilon^2
\right)\!.
\label{eq:quadratic_term_final}
\end{align}

\paragraph{Combining Steps 1 and 2.}
\addb{Adding \eqref{eq:linear_term_expanded} and \eqref{eq:quadratic_term_final}
via \eqref{eq:smooth_descent_g}, the Gr\"{o}nwall condition
\eqref{eq:eta_gronwall} gives $32\widetilde L^2\eta^2\Lambda_q(\tau)\le 1$,
and in particular
$\widetilde L\eta\tau/q
=\sqrt{\tau/q}\,\bigl(\widetilde L\eta\sqrt{\Lambda_q(\tau)}\bigr)\big/\sqrt{1+\tau/q}
\le 1/\sqrt{32}$.
Each cubic-in-$\eta$ correction in \eqref{eq:linear_term_expanded} equals the
corresponding $\widetilde L\eta^2$ term of \eqref{eq:quadratic_term_final}
multiplied by $2\widetilde L\eta\tau/q\le 1/(2\sqrt2)$, and is therefore
dominated by it.
Moreover, since $c_\delta^2=q\widetilde L^2/4$ (Lemma~\ref{lem:coord_bias}),
$6\widetilde L^2\tau = 24\,c_\delta^2\,\tfrac{\tau}{q}
\le 24\,c_\delta^2\,\Lambda_q(\tau)$.
Consequently, there exist absolute constants $\bar C_1,\bar C_2,\bar C_3>0$
such that}
\begin{align}
\E_k\bigl[f(X^{(k+1)})-f(X^{(k)})\bigr]
&\le
-\frac{\eta\tau}{2}\fnorm{\nabla f(X^{(k)})}^2
+ \bar C_1\widetilde L\eta^2 q\,\Lambda_q(\tau)\fnorm{\nabla f(X^{(k)})}^2
\nonumber\\
&\quad
+ \bar C_2\widetilde L\,\frac{\eta^2\tau}{b}\sigma^2
+ \!\left(c_\delta^2\frac{\eta\tau}{q} + \bar C_3\,c_\delta^2\,\widetilde L\eta^2\Lambda_q(\tau)\right)\!\epsilon^2.
\label{eq:combined_simplified}
\end{align}
\addb{Since $q\,\Lambda_q(\tau) = \tau(1+\tfrac{\tau}{q})$, the second term
on the right-hand side equals
$\bar C_1\widetilde L\eta(1+\tfrac{\tau}{q})\cdot\eta\tau\fnorm{\nabla f(X^{(k)})}^2$.
Identifying with \eqref{eq:one_period_descent},}
\[
c_1 := \tfrac{1}{2},\qquad
c_2 := \bar C_1,\qquad
c_3 := \bar C_2,\qquad
c_4 := \bar C_3\,c_\delta^2
\]
yields \eqref{eq:one_period_descent}.
\end{proof}

\subsection{Recovery of the full gradient in expectation}

We now relate the subspace gradient $\nabla g_k(0)=U_k^{\top}\nabla f(X^{(k)})V_k$ to the full gradient $\nabla f(X^{(k)})$.

\begin{lemma}[Coverage of sphere-normalized subspaces]
\label{lem:subspace_coverage}
Under Assumption~\ref{assump:subspace}, for every deterministic matrix
$G \in \R^{d_1 \times d_2}$,
\begin{equation}
\E_{U_k,V_k}\fnorm{U_k^{\top} G V_k}^2
=
r_1r_2\fnorm{G}^2.
\label{eq:coverage}
\end{equation}
\end{lemma}

\begin{proof}
\add{By Lemma~\ref{lem:cond_ensemble}, each column satisfies
$\E[u_{k,a}u_{k,a}^{\top}]=I_{d_1}$ and
$\E[v_{k,b}v_{k,b}^{\top}]=I_{d_2}$, and $U_k$ and $V_k$ are independent;
the computation below uses only these facts and requires no independence
across columns.}
Therefore,
\begin{align*}
\E_{U_k,V_k}\fnorm{U_k^{\top} G V_k}^2
&=
\sum_{a=1}^{r_1}\sum_{b=1}^{r_2}
\E\bigl[(u_{k,a}^{\top}Gv_{k,b})^2\bigr]
\\
&=
\sum_{a=1}^{r_1}\sum_{b=1}^{r_2}
\E\bigl[u_{k,a}^{\top}G\,\E[v_{k,b}v_{k,b}^{\top}]\,G^{\top}u_{k,a}\bigr]
\\
&=
\sum_{a=1}^{r_1}\sum_{b=1}^{r_2}
\E\bigl[u_{k,a}^{\top}GG^{\top}u_{k,a}\bigr]
=
r_1r_2\fnorm{G}^2.
\end{align*}
\end{proof}

\begin{lemma}[Projected gradient recovers the full gradient in expectation]
\label{lem:coverage}
Under Assumption~\ref{assump:subspace},
\begin{equation}
\E_{U_k,V_k}\fnorm{\nabla g_k(0)}^2
=
\E_{U_k,V_k}\fnorm{U_k^{\top} \nabla f(X^{(k)}) V_k}^2
=
r_1r_2 \fnorm{\nabla f(X^{(k)})}^2.
\label{eq:coverage_grad}
\end{equation}
\end{lemma}

\begin{proof}
The equality is just Lemma~\ref{lem:grad_g} evaluated at $B=0$.
Since $U_k$ and $V_k$ are sampled independently of $X^{(k)}$,
Lemma~\ref{lem:subspace_coverage} applies conditionally on $X^{(k)}$ with
$G = \nabla f(X^{(k)})$.
\end{proof}

\subsection{Global convergence theorem}

We are now ready to derive the global convergence result.

\begin{restatable}[Convergence of the subspace coordinate-wise gradient estimator]{proposition}{ThmLowrankEstimator}
  \label{thm:lowrank-estimator}
  Under Assumptions~\ref{assump:smoothness}--\ref{assump:coordinate}, let
  $q:=r_1r_2$, $T:=K\tau$, $\widetilde{L}:=L\rho_U^2\rho_V^2$ and
  $\Lambda_q(\tau):=\tfrac{\tau}{q}(1+\tfrac{\tau}{q})$. If
  $\eta\le\bigl(\widetilde{L}{\sqrt{32\,\Lambda_q(\tau)}}\bigr)^{-1}$ and
  $c_2\eta\widetilde{L}(1+\tau/q)\le c_1/2$, then
  \[
      \frac{1}{K}\sum_{k=0}^{K-1}\mathbb{E}\|\nabla f(X_{k\tau})\|_F^2
      \;\le\;
      \underbrace{\frac{2\Delta_0}{c_1\eta T}}_{\mathrm{optimization}}
      +\underbrace{\frac{2c_3\widetilde{L}\eta{\sigma^2}}{c_1 b}}_{\mathrm{stochastic\ noise}}
      +\underbrace{\frac{2c_4}{c_1 q}\!\left[1+\widetilde{L}\eta\!\left(1+\frac{\tau}{q}\right)\right]\!\epsilon^2}_{\mathrm{finite\ difference\ bias}},
  \]
  where $\Delta_0:=f(X_0)-f^\star$, $c_1,c_2,c_3>0$ are universal constants, and $c_4>0$ is a problem-dependent constant that may depend on the subspace dimension $q$ and the effective smoothness parameter $\widetilde{L}$ (specifically, $c_4=O(c_\delta^2)$ with $c_\delta=\sqrt{q}\,\widetilde{L}/2$).
  \end{restatable}

\begin{remark}[Constants and scaling in \Cref{thm:lowrank-estimator}]
The quantities appearing in the bound of \Cref{thm:lowrank-estimator} are
\[
q=r_1r_2,\qquad
\Lambda_q(\tau)=\frac{\tau}{q}\left(1+\frac{\tau}{q}\right),
\qquad
\widetilde L=L\rho_U^2\rho_V^2.
\]
% ======================= [E8] BEGIN ========================================

\add{Under the conditioned sphere-normalized sampling of
Assumption~\ref{assump:subspace},}
\begin{equation*}
\rho_U^2\rho_V^2=16\,d_1d_2,
\qquad
\widetilde L=16\,Ld_1d_2,
\end{equation*}
\addb{so that $\widetilde L$ is independent of the subspace dimension $q$,
and the stochastic-noise coefficient of \Cref{thm:lowrank-estimator}
is the dimension-free minibatch variance $\sigma^2$ itself.}
The finite-difference bias constant is
\[
c_\delta=\frac{\sqrt q\,\widetilde L}{2},
\qquad
c_\delta^2=O(q\widetilde L^2)
\;=O\bigl(qL^2d_1^2d_2^2\bigr).
\]
% ======================= [E8] END ==========================================
The descent constants satisfy \(c_1=1/2\), \(c_2,c_3=O(1)\), and
\(c_4=O(c_\delta^2)\).  Hence \Cref{thm:lowrank-estimator} can be read as
\[
\frac{1}{K}\sum_{k=0}^{K-1}\E\fnorm{\nabla f(X_{k\tau})}^2
\le
O\!\left(
\frac{\Delta_0}{\eta T}
+
\frac{\widetilde L\eta\sigma^2}{b}
+
\frac{c_\delta^2}{q}
\left[1+\widetilde L\eta\left(1+\frac{\tau}{q}\right)\right]\epsilon^2
\right).
\]
Equivalently, since \(c_\delta^2/q=O(\widetilde L^2)\), the finite-difference
term is \(O(\widetilde L^2[1+\widetilde L\eta(1+\tau/q)]\epsilon^2)\).
The stepsize condition is
\[
\eta
=
O\!\left(
\min\left\{
\frac{1}{\widetilde L\sqrt{\Lambda_q(\tau)}},
\frac{1}{\widetilde L(1+\tau/q)}
\right\}
\right).
\]
\end{remark}

\begin{proof}

By the tower property,
\[
\E\bigl[f(X^{(k+1)})-f(X^{(k)})\bigr]
= \E\Bigl[\,\E\bigl[f(X^{(k+1)})-f(X^{(k)}) \bigm| X^{(k)}\bigr]\Bigr].
\]
The first stepsize condition of \Cref{thm:lowrank-estimator},
$\eta \le 1/(\widetilde L\sqrt{32\,\Lambda_q(\tau)})$, ensures that the
one-period descent lemma can be applied.
By Lemma~\ref{lem:one_period},
\begin{align}
\E\bigl[f(X^{(k+1)}) - f(X^{(k)})\bigr]
&\le
-\!\left(c_1 - c_2\eta\widetilde L\!\left(1+\frac{\tau}{q}\right)\!\right)\!\eta\tau\,\E\fnorm{\nabla f(X^{(k)})}^2
\nonumber\\
&\quad
+ c_3\widetilde L\,\frac{\eta^2\tau}{b}\,\sigma^2
+ c_4\!\left(\frac{\eta\tau}{q} + \widetilde L\eta^2\Lambda_q(\tau)\right)\!\epsilon^2.
\label{eq:global_step1}
\end{align}
By the second stepsize condition of \Cref{thm:lowrank-estimator},
$c_2\eta\widetilde L(1+\tau/q)\le c_1/2$,
\[
c_1 - c_2\eta\widetilde L\!\left(1+\frac{\tau}{q}\right) \;\ge\; \frac{c_1}{2},
\]
so that
\begin{align}
\E\bigl[f(X^{(k+1)}) - f(X^{(k)})\bigr]
&\le
-\frac{c_1}{2}\,\eta\tau\,\E\fnorm{\nabla f(X^{(k)})}^2
+ c_3\widetilde L\,\frac{\eta^2\tau}{b}\,\sigma^2
+ c_4\!\left(\frac{\eta\tau}{q} + \widetilde L\eta^2\Lambda_q(\tau)\right)\!\epsilon^2.
\label{eq:global_step2}
\end{align}
Summing \eqref{eq:global_step2} over $k=0,1,\dots,K-1$ and telescoping,
\begin{align}
\frac{c_1}{2}\,\eta\tau\sum_{k=0}^{K-1} \E\fnorm{\nabla f(X^{(k)})}^2
&\le
\Delta_0
+ K c_3\widetilde L\,\frac{\eta^2\tau}{b}\,\sigma^2
+ K c_4\!\left(\frac{\eta\tau}{q} + \widetilde L\eta^2\Lambda_q(\tau)\right)\!\epsilon^2.
\label{eq:global_step3}
\end{align}
Dividing both sides by $K(c_1/2)\eta\tau$ gives
\begin{align}
\frac{1}{K}\sum_{k=0}^{K-1} \E\fnorm{\nabla f(X^{(k)})}^2
&\le
\frac{2\Delta_0}{c_1 \eta\tau K}
+
\frac{2c_3}{c_1}\,\frac{\widetilde L\eta}{b}\,\sigma^2
+
\frac{2c_4}{c_1q}\!\left[1 + \frac{q\widetilde L\eta\Lambda_q(\tau)}{\tau}\right]\!\epsilon^2.
\label{eq:main_proof_step4}
\end{align}
Since $\frac{q\Lambda_q(\tau)}{\tau} = 1+\frac{\tau}{q}$, $X^{(k)} = X_{k\tau}$, and $T = K\tau$, equation \eqref{eq:main_proof_step4} is exactly the bound stated in \Cref{thm:lowrank-estimator}.
\end{proof}

\section{SeedFlood as Data-Parallel Subspace Zeroth-Order Optimization}
\label{appendix:seedflood-proof}

We now extend the single-client analysis to the idealized decentralized
SeedFlood setting.  The role of flooding is separated from the optimization
analysis: we assume that, at each inner step, flooding disseminates all client-side
zeroth-order estimators exactly before the model update is applied.  Under this
exact-flooding idealization, the optimization dynamics are the same as
centralized data-parallel zeroth-order optimization over the shared low-rank
subspace with \(n\) independent client estimators.

\subsection{Decentralized setup and exact flooding}

Let \(n\) clients have local objectives
\[
f_i(X):=\E_{\xi\sim\mathcal D_i}[F_i(X;\xi)],
\qquad
f(X):=\frac1n\sum_{i=1}^n f_i(X).
\]
For the data-parallel reduction, all clients use the same period-wise subspace
\(U_k,V_k\), generated for instance from a shared random seed, as in
Assumption~\ref{assump:subspace}.  For each period \(k\), define
\[
g_{i,k}(B):=f_i(X^{(k)}+U_kBV_k^\top),
\qquad
g_k(B):=\frac1n\sum_{i=1}^n g_{i,k}(B).
\]
Thus \(g_k(B)=f(X^{(k)}+U_kBV_k^\top)\), so the global subspace objective is
exactly the average of the local subspace objectives.

\begin{restatable}[Per-client smoothness, bounded noise, and heterogeneity]{assumption}{AssumpClient}
\label{assump:client}
\addb{$f$ is bounded below by $f^\star$, For each client $i$, every realization $F_i(\cdot;\xi)$ is $L$-smooth
(hence $f_i$ is $L$-smooth), stochastic gradients are unbiased with
variance $\le\sigma^2/b$, and
$\tfrac{1}{n}\sum_i\|\nabla f_i(X)-\nabla f(X)\|_F^2\le\zeta^2$ for all $X$.}
The global objective $f=\tfrac1n\sum_i f_i$ is bounded below by
$f^\star$.
\end{restatable}

Equivalently, for each client \(i\) we assume that \addb{every realization
\(F_i(\cdot;\xi)\) is \(L\)-smooth,
\begin{equation}
\fnorm{\nabla F_i(X;\xi)-\nabla F_i(Y;\xi)} \le L\fnorm{X-Y},
\qquad \forall X,Y,\ \forall\xi,
\label{eq:per_realization_smooth_client}
\end{equation}
so that \(f_i\) is \(L\)-smooth by Jensen's inequality,} and mini-batch
stochastic gradients are unbiased with bounded variance,
\[
\E_{\xi\sim\mathcal D_i}[\nabla F_i(X;\xi)] = \nabla f_i(X),
\qquad
\E_{\xi\sim\mathcal D_i}\fnorm{\nabla F_i(X;\xi)-\nabla f_i(X)}^2
\le \frac{\sigma^2}{b},
\]
and that the local gradients have bounded dissimilarity,
\begin{equation}
\frac1n\sum_{i=1}^n
\fnorm{\nabla f_i(X)-\nabla f(X)}^2
\le
\zeta^2,
\qquad \forall X.
\label{eq:gradient_dissimilarity}
\end{equation}

By Lemma~\ref{lem:grad_g} applied to each local objective and the operator norm
bound in Assumption~\ref{assump:subspace}, Assumption~\ref{assump:client}
implies the coefficient-space heterogeneity bound
\begin{equation}
\frac1n\sum_{i=1}^n
\fnorm{\nabla g_{i,k}(B)-\nabla g_k(B)}^2
\le
\widetilde\zeta^2,
\qquad
\widetilde\zeta^2:=\rho_U^2\rho_V^2\zeta^2.
\label{eq:subspace_heterogeneity}
\end{equation}

At inner step \(s\), client \(i\) samples an independent coordinate
\[
J_{i,k,s}\sim\mathrm{Unif}\{1,\ldots,q\},
\qquad C_{i,k,s}=E_{J_{i,k,s}},
\]
and computes the local coefficient
\[
a_{i,k,s}
:=
\frac{
G_{i,k}(B_{k,s}+\epsilon C_{i,k,s};\xi_{i,k,s})
-
G_{i,k}(B_{k,s}-\epsilon C_{i,k,s};\xi_{i,k,s})
}{2\epsilon},
\]
where
\[
G_{i,k}(B;\xi):=F_i(X^{(k)}+U_kBV_k^\top;\xi).
\]
The local coefficient-space estimator is
\[
\widehat{\nabla}g_{i,k}(B_{k,s};\xi_{i,k,s},J_{i,k,s})
:=
a_{i,k,s}E_{J_{i,k,s}}.
\]
Exact flooding averages these estimators:
\begin{equation}
\widehat{\nabla}_{\mathrm{SF}}g_k(B_{k,s})
:=
\frac1n\sum_{i=1}^n
\widehat{\nabla}g_{i,k}(B_{k,s};\xi_{i,k,s},J_{i,k,s}).
\label{eq:sf_avg_estimator}
\end{equation}
The common SeedFlood iterate then satisfies
\begin{equation}
B_{k,s+1}
=
B_{k,s}-\eta\,\widehat{\nabla}_{\mathrm{SF}}g_k(B_{k,s}),
\qquad
X_{k,s}=X^{(k)}+U_kB_{k,s}V_k^\top.
\label{eq:sf_B_update}
\end{equation}

\begin{proposition}[Exact flooding yields data-parallel subspace zeroth-order updates]
\label{prop:sf_equivalence}
Suppose all clients start inner step \(s\) from the same model \(X_{k,s}\), and
flooding returns the exact average \eqref{eq:sf_avg_estimator} before the update.
Then all clients remain synchronized, and the common model follows
\eqref{eq:sf_B_update}.  Hence the decentralized optimization dynamics are
identical to centralized data-parallel zeroth-order optimization over the
shared low-rank subspace.
\end{proposition}

\begin{proof}
Each local update direction has the form
\(U_k\widehat{\nabla}g_{i,k}(B_{k,s};\xi_{i,k,s},J_{i,k,s})V_k^\top\).
Exact flooding replaces the local direction by their average.  Therefore the
full-space update is
\[
X_{k,s+1}
=
X_{k,s}
-
\eta U_k
\left(
\frac1n\sum_{i=1}^n
\widehat{\nabla}g_{i,k}(B_{k,s};\xi_{i,k,s},J_{i,k,s})
\right)
V_k^\top,
\]
which is exactly \eqref{eq:sf_B_update} under the representation
\(X_{k,s}=X^{(k)}+U_kB_{k,s}V_k^\top\).
\end{proof}

\subsection{Averaged estimator moments}

For the exact-flooding analysis, the effective inner-loop drift scale is
\begin{equation}
\Lambda_{q,n}(\tau)
:=
\frac{\tau^2}{q^2}+\frac{\tau}{nq}.
\label{eq:lambda_qn}
\end{equation}
The first term is the deterministic drift caused by repeatedly moving in the
mean direction \(\nabla g_k/q\) inside a fixed subspace.  The second term comes
from the averaged coordinate-sampling fluctuation and is reduced by \(n\).

\begin{lemma}[Mean and second moment under exact flooding]
\label{lem:sf_moments}
Under Assumptions~\ref{assump:subspace} and
\ref{assump:client}, and the coordinate finite-difference construction
above, the averaged SeedFlood estimator satisfies
\begin{align}
\E\!\left[
\widehat{\nabla}_{\mathrm{SF}}g_k(B)
\mid X^{(k)},U_k,V_k,B
\right]
&=
\frac1q\left(\nabla g_k(B)+\delta_k(B)\right),
\label{eq:sf_mean}
\end{align}
where \(\fnorm{\delta_k(B)}\le c_\delta\epsilon\).
\addb{Moreover, let
$\mathcal{F}_{k,s}
:=\sigma\bigl(X^{(k)},U_k,V_k,
\{\xi_{i,k,\ell},J_{i,k,\ell}\}_{i\in[n],\,\ell<s}\bigr)$;
if $B$ is $\mathcal{F}_{k,s}$-measurable and
$(\xi_{i,k,s},J_{i,k,s})_{i\in[n]}$ is the fresh per-client randomness of
step $s$, then}
\begin{align}
\E\!\left[
\fnorm{\widehat{\nabla}_{\mathrm{SF}}g_k(B)}^2
\mid X^{(k)}
\right]
&\le
\left(\frac{2}{q^2}+\frac{2}{nq}\right)\E\bigl[\fnorm{\nabla g_k(B)}^2 \mid X^{(k)}\bigr]
+
\frac{2}{nq}\widetilde\zeta^2
\nonumber\\
&\quad
+
2c_\delta^2\left(\frac{1}{q^2}+\frac{1}{nq}\right)\epsilon^2
+
\frac{3\sigma^2}{nb}
+
\frac{12\widetilde L^2}{nq}\,\E\bigl[\fnorm{B}^2 \mid X^{(k)}\bigr]
+
\frac{12\widetilde L^2}{n}\epsilon^2 .
\label{eq:sf_second_moment}
\end{align}
\end{lemma}

\begin{proof}
Write
\[
h_i:=\nabla g_{i,k}(B),\qquad
h:=\nabla g_k(B)=\frac1n\sum_{i=1}^n h_i,
\]
and let
\[
Z_i:=\widehat{\nabla}g_{i,k}(B;\xi_i,J_i),
\qquad
Z:=\widehat{\nabla}_{\mathrm{SF}}g_k(B)=\frac1n\sum_{i=1}^n Z_i.
\]
For each client, the standard CGE finite-difference property gives
\[
\mu_i:=\E[Z_i\mid X^{(k)},U_k,V_k,B]
=
\frac1q(h_i+\delta_{i,k}(B)),
\qquad
\fnorm{\delta_{i,k}(B)}\le c_\delta\epsilon.
\]
Averaging over clients yields \eqref{eq:sf_mean} with
\(\delta_k(B)=\frac1n\sum_i\delta_{i,k}(B)\).

For the second moment, decompose the average estimator into its
$\mathcal{F}_{k,s}$-conditional mean and centered fluctuations:
\[
\bar\mu:=\E[Z\mid \mathcal{F}_{k,s}]
=
\frac1n\sum_{i=1}^n\mu_i
=
\frac1q(h+\delta_k(B)),
\qquad
\E[\fnorm{Z}^2\mid \mathcal{F}_{k,s}]
=
\fnorm{\bar\mu}^2
+
\E[\fnorm{Z-\bar\mu}^2\mid \mathcal{F}_{k,s}].
\]
The mean term satisfies, pointwise on $\mathcal{F}_{k,s}$,
\begin{equation}
\fnorm{\bar\mu}^2
=
\frac1{q^2}\fnorm{h+\delta_k(B)}^2
\le
\frac{2}{q^2}\fnorm h^2
+
\frac{2c_\delta^2}{q^2}\epsilon^2.
\label{eq:sf_mean_square}
\end{equation}
Because the clients use independent fresh mini-batches and coordinates, the
centered variables \(Z_i-\mu_i\) are conditionally independent given
$\mathcal{F}_{k,s}$ and have zero mean, so the cross terms vanish:
\begin{align}
\E[\fnorm{Z-\bar\mu}^2\mid \mathcal{F}_{k,s}]
&=
\frac1{n^2}\sum_{i=1}^n
\E[\fnorm{Z_i-\mu_i}^2\mid \mathcal{F}_{k,s}]
\le
\frac1{n^2}\sum_{i=1}^n
\E[\fnorm{Z_i}^2\mid \mathcal{F}_{k,s}].
\label{eq:sf_variance_split}
\end{align}
By Assumption~\ref{assump:client}, each pair \((g_{i,k},G_{i,k})\) satisfies
the hypotheses of Lemma~\ref{lem:coord_noise_anchor} and
Lemma~\ref{lem:coord_estimator_moments}: every realization
\(F_i(\cdot;\xi)\) is \(L\)-smooth \eqref{eq:per_realization_smooth_client},
mini-batch gradients are unbiased with variance at most \(\sigma^2/b\), and
the anchor
\(\Delta_0^{(i)}:=\nabla F_i(X^{(k)};\xi_{i,k,s})-\nabla f_i(X^{(k)})\)
is independent of \((U_k,V_k)\).
Hence, taking \(\E[\,\cdot\mid X^{(k)}]\) of \eqref{eq:sf_variance_split}
via the tower property and applying \eqref{eq:coord_second_moment} verbatim
to each client,
\begin{align}
\E[\fnorm{Z-\bar\mu}^2\mid X^{(k)}]
&\le
\frac{2}{nq}
\left(\frac1n\sum_{i=1}^n\E\bigl[\fnorm{h_i}^2 \mid X^{(k)}\bigr]\right)
+
\frac{2c_\delta^2}{nq}\epsilon^2
+
\frac{3\sigma^2}{nb}
+
\frac{12\widetilde L^2}{nq}\,\E\bigl[\fnorm{B}^2 \mid X^{(k)}\bigr]
+
\frac{12\widetilde L^2}{n}\epsilon^2 .
\label{eq:sf_variance_bound}
\end{align}
Finally, since \(h=\frac1n\sum_i h_i\), pointwise
\[
\frac1n\sum_{i=1}^n\fnorm{h_i}^2
=
\fnorm h^2
+
\frac1n\sum_{i=1}^n\fnorm{h_i-h}^2
\le
\fnorm h^2+\widetilde\zeta^2,
\]
where the last step uses \eqref{eq:subspace_heterogeneity}.  Taking
\(\E[\,\cdot\mid X^{(k)}]\) of \eqref{eq:sf_mean_square} and combining with
\eqref{eq:sf_variance_bound} proves \eqref{eq:sf_second_moment}.
\end{proof}

\subsection{\methodname{} convergence}

\ThmSeedflood*

\begin{remark}[Constants and scaling in \Cref{thm:seedflood}]
The constants \(q,\widetilde L\), and \(c_\delta\) are the same as in
\Cref{thm:lowrank-estimator}.  The effective SeedFlood drift scale is
\[
\Lambda_{q,n}(\tau)=\frac{\tau^2}{q^2}+\frac{\tau}{nq}.
\]
The additional heterogeneity constant is
\[
\widetilde\zeta^2:=\rho_U^2\rho_V^2\zeta^2,
\]
% ======================= [E9] BEGIN ========================================
so under \add{the conditioned
sphere-normalized sampling of Assumption~\ref{assump:subspace}},

\begin{equation*}
\widetilde\zeta^2=16\,d_1d_2\,\zeta^2.
\end{equation*}
\add{In particular, the heterogeneity contribution
$\widetilde\zeta^2/(qn)=16\,d_1d_2\zeta^2/(qn)$ in
\Cref{thm:seedflood} decreases with the subspace dimension $q$.}
% ======================= [E9] END ==========================================
Thus \Cref{thm:seedflood} can be written in big-\(O\) form as
\begin{align*}
&\frac1K\sum_{k=0}^{K-1}\E\fnorm{\nabla f(X_{k\tau})}^2
\\
&\le
O\!\left(
\frac{\Delta_0}{\eta T}
+
\widetilde L\eta
\left(
\frac{\sigma^2}{nb}
+
\frac{\widetilde\zeta^2}{qn}
\right)
+
\frac{c_\delta^2}{q}
\left[1+\widetilde L\eta\left(\frac{\tau}{q}+\frac1n\right)\right]\epsilon^2
\right).
\end{align*}
The admissible stepsize conditions now improve with \(n\), because averaging
independent client estimators also reduces the coordinate-sampling fluctuation.
Equivalently,
\[
\eta
=
O\!\left(
\min\left\{
\frac{1}{\widetilde L\sqrt{\tau^2/q^2+\tau/(nq)}},
\frac{1}{\widetilde L(\tau/q+1/n)}
\right\}
\right).
\]
Thus increasing \(n\) can reduce the first optimization-transient term
\(\Delta_0/(\eta T)\) indirectly by allowing a larger stable stepsize.  This
improvement saturates at the deterministic fixed-subspace drift scale
\(\eta=O(q/(\widetilde L\tau))\).  In addition, larger \(n\) directly reduces
the stochastic variance and client-heterogeneity terms by \(1/n\).  The
deterministic finite-difference bias itself is not reduced by \(n\), although
its drift amplification is affected through the larger stable stepsize range.
\end{remark}

\begin{proof}

The proof repeats the single-client proof of \Cref{thm:lowrank-estimator}
with \(\widehat{\nabla}g_k\) replaced by the averaged estimator
\(\widehat{\nabla}_{\mathrm{SF}}g_k\), all conditional expectations taken
with respect to \(X^{(k)}\) alone, and Lemma~\ref{lem:coverage} applied inside
the drift and descent arguments exactly as in Lemmas~\ref{lem:drift}
and~\ref{lem:one_period}.
The conditional mean \eqref{eq:sf_mean} is identical to the single-client
mean with \(g_k\) now equal to the global subspace objective, so the descent
term becomes \(-\eta\tau\fnorm{\nabla f(X^{(k)})}^2\) up to the same
corrections as in the single-client case.

For the stochastic part we use, as in the single-client analysis, the
orthogonal decomposition
\(\E[\fnorm{Z}^2\mid\mathcal F_{k,s}]
=\fnorm{\bar\mu}^2+\E[\fnorm{Z-\bar\mu}^2\mid\mathcal F_{k,s}]\)
established in the proof of Lemma~\ref{lem:sf_moments}, rather than the
uncentered bound \eqref{eq:sf_second_moment}, so that the deterministic mean
is not conflated with the zero-mean fluctuation.
The centered fluctuations across inner steps form a martingale-difference
sequence with respect to \(\{\mathcal F_{k,j}\}\), and the orthogonality
survives the coarser conditioning on \(X^{(k)}\) exactly as in
Lemma~\ref{lem:drift}.
The mean component \eqref{eq:sf_mean_square} contributes the deterministic
signal/drift terms; the anchored centered-variance bound
\eqref{eq:sf_variance_bound} contributes the dimension-free noise floor
\(3\sigma^2/(nb)\), the heterogeneity term \(\widetilde\zeta^2\!/(nq)\), and
a drift correction \(12\widetilde L^2\,\E\fnorm{B}^2/(nq)\) that enters the
Gr\"onwall recursion at the same order in \((q,n)\) as the signal terms,
changing only absolute constants; the induced Gr\"onwall coefficient
satisfies \(\beta\tau\le 32\,\eta^2\widetilde L^2\Lambda_{q,n}(\tau)\), which
is the origin of the stepsize condition
\(\eta\le 1/(\widetilde L\sqrt{32\,\Lambda_{q,n}(\tau)})\).
The remaining \(\widetilde L^2\epsilon^2/n\) correction is absorbed into the
finite-difference bracket via \(c_\delta^2=q\widetilde L^2/4\), as in the
proof of Lemma~\ref{lem:one_period}.
Propagating these two components through the arguments of
Lemmas~\ref{lem:drift} and~\ref{lem:one_period} gives, for universal
numerical constants \(\bar a_\sigma,\bar a_\zeta,\bar a_\epsilon>0\), the
one-period bound
\[
\E[f(X^{(k+1)})-f(X^{(k)})\mid X^{(k)}]
\le
-\left(c_1-c_2\eta\widetilde L\left(\frac{\tau}{q}+\frac1n\right)\right)
\eta\tau\,\fnorm{\nabla f(X^{(k)})}^2
\]
\[
\qquad
+
\bar a_\sigma\widetilde L\eta^2\tau\,\frac{\sigma^2}{nb}
+
\bar a_\zeta\widetilde L\eta^2\tau\,\frac{\widetilde\zeta^2}{qn}
+
\bar a_\epsilon c_\delta^2\left(\frac{\eta\tau}{q}+\widetilde L\eta^2\Lambda_{q,n}(\tau)\right)\epsilon^2.
\]
The drift amplification factor is the SeedFlood drift scale
\(\Lambda_{q,n}(\tau)=\tau^2/q^2+\tau/(nq)\): the \(\tau^2/q^2\) term comes
from the deterministic mean component \eqref{eq:sf_mean_square} (which is
not reduced by averaging over clients), while the \(\tau/(nq)\) term comes
from the centered variance \eqref{eq:sf_variance_bound} (whose
client-averaged contribution is reduced by \(1/n\)).

The stated result follows by summing the one-period bound over
\(k=0,\ldots,K-1\), using \(f(X)\ge f^\star\), and dividing by
\(K(c_1/2)\eta\tau\), with
\[
a_\sigma:=\frac{2\bar a_\sigma}{c_1},\qquad
a_\zeta:=\frac{2\bar a_\zeta}{c_1},\qquad
a_\epsilon:=\frac{2\bar a_\epsilon}{c_1}.
\]
\end{proof}

%% file: sections/appendix-expdetail.tex
\section{Experiment setup}
\label{app:exp_setup}

\subsection{Dataset}
Due to computational constraints, we evaluate our method on a subset of SuperGLUE tasks \citep{wang2019superglue}.
For experiments with OPT-1.3B using 16 clients, we consider BoolQ \citep{clark2019boolq}, MultiRC \cite{MultiRC}, ReCoRD \citep{zhang2018record}, RTE \citep{RTE1, RTE2, RTE3}, and WiC\citep{pilehvar2019wic}, and additionally include SST-2 \citep{socher2013recursive}.

For experiments with OPT-125M and larger client counts \{16, 32, 64, 128 \}, we restrict evaluation to SST-2 \cite{socher2013recursive}, BoolQ \cite{clark2019boolq}, and RTE \citep{RTE1, RTE2, RTE3} to ensure feasibility.

For each task, we randomly sample 1,024 training examples and distribute them uniformly across clients. We use a fixed global validation set of 500 examples and a test set of 1,000 examples.

We use the same prompts as \citet{MeZO} for all tasks, and apply them consistently across both zeroth-order and first-order methods.
\subsection{Hyperparameters}
\begin{table}[!t]
\centering
\footnotesize
\setlength{\tabcolsep}{6pt}
\renewcommand{\arraystretch}{1.05}
\caption{Hyperparameter setup for experiment \Cref{sec:main-results} and \Cref{sec:scaling} Validation loss is evaluated every one-tenth of the total training iterations, and the model achieving the best validation loss is selected for evaluation on a held-out test set.
}
\label{tab:hparams}
\begin{tabular}{l l c}
\toprule
\textbf{Experiment} & \textbf{Hyperparameter} & \textbf{Value} \\
\midrule

% -------------------- DSGD (FT) --------------------
\multirow{3}{*}{DSGD}
& batch size & 8 \\
& learning rate & \{\,1e-4, 1e-5, 1e-6\,\} \\
& local iteration & 5 \\
\midrule

% -------------------- DSGD (LoRA) --------------------
\multirow{3}{*}{DSGD (LoRA)}
& batch size & 8 \\
& learning rate & \{\,1e-2, 1e-3, 1e-4\,\} \\
& local iteration & 5 \\
\midrule

% -------------------- ChocoSGD (FT) --------------------
\multirow{4}{*}{ChocoSGD}
& batch size & 8 \\
& learning rate & \{\,1e-4, 1e-5, 1e-6\,\} \\
& local iteration & 5 \\
& consensus step size & 1 \\
& compression & Top-K 99\% sparsification \\
\midrule

% -------------------- ChocoSGD (LoRA) --------------------
\multirow{4}{*}{ChocoSGD (LoRA)}
& batch size & 8 \\
& learning rate & \{\,1e-2, 1e-3, 1e-4\,\} \\
& local iteration & 5 \\
& consensus step size & 1 \\
& compression & Top-K 99\% sparsification\\
\midrule

% -------------------- DZSGD (FT, ZO) --------------------
\multirow{5}{*}{DZSGD}
& batch size & 16 \\
& learning rate & \{ 1e-5, 1e-6 \} \\
& local iteration & 5 \\
& perturbation scale $\epsilon$ & 1e-3 \\
\midrule

% -------------------- DZSGD (LoRA, ZO) --------------------
\multirow{5}{*}{DZSGD (LoRA)}
& batch size & 16 \\
& learning rate & \{ 1e-3, 1e-4, 1e-5 \} \\
& local iteration & 5 \\
& perturbation scale $\epsilon$ & 1e-3 \\
\midrule

% -------------------- SeedFlood (ZO + flooding) --------------------
\multirow{6}{*}{SeedFlood (\Cref{sec:main-results})}
& batch size & 16 \\
& learning rate & \{ 1e-5, 1e-6 \} \\
& flooding steps $k$ & same as Network Diameter \\
& perturbation scale $\epsilon$ & 1e-3 \\
& subspace rank $r$ & 32 \\
& subspace refresh period $\tau$ & 1000 \\
\midrule

\multirow{6}{*}{SeedFlood (\Cref{sec:scaling})}
& batch size & 16 \\
& learning rate & \{ 1e-5, 1e-6 \} \\
& flooding steps $k$ & same as Network Diameter \\
& perturbation scale $\epsilon$ & 1e-3 \\
& subspace rank $r$ & 64 \\
& subspace refresh period $\tau$ & 5000 (no change) \\

\bottomrule
\end{tabular}
\end{table}

We report the hyperparameter settings used in \Cref{sec:main-results,sec:scaling}. Following prior work \citep{MeZO, yu2025zeroth}, we do not apply momentum and weight decay to local SGD updates. To ensure feasibility under limited computational resources, all baseline methods are configured to exchange communication every five local iterations. While increasing the number of local iterations can reduce communication overhead, overly large local steps are known to degrade performance; therefore, we adopt the smallest local iteration count permitted by our computational budget. All experiments use a constant learning rate.

For full-parameter tuning of ChocoSGD, we initialize surrogate model parameters with pretrained weights, which yields substantial performance improvement.

For SeedFlood, we fix the subspace rank and refresh period without additional search. Specifically, experiments in \Cref{sec:main-results} use a rank of 32 with a refresh period of 1000, while those in \Cref{sec:scaling} use a rank of 64 and a refresh period of 5000. Since the total number of training iterations is 5000, this effectively corresponds to using a fixed subspace throughout training. For zeroth-order methods, the perturbation scale is set following \citet{MeZO}. All other training-related settings not explicitly mentioned follow the configuration of \cite{MeZO}.

\subsection{LoRA configuration}
For all LoRA-based experiments, we adopt a fixed LoRA configuration. Specifically, we set the LoRA rank to 8 and the scaling factor 
$\alpha$ to 16. LoRA adapters are applied to the query and value projection layers (\texttt{q\_proj} and \texttt{v\_proj}) of the transformer. We use a LoRA dropout rate of 0.05 across all experiments.

FFA-LoRA \citep{sun2024improving} was originally proposed for server-based federated learning, where freezing the randomly initialized $A$ matrices removes the aggregation error caused by averaging the products of two trainable low-rank factors.
We adapt it to our decentralized setting as follows: all clients initialize the $A$ matrices identically and keep them frozen throughout training, so that only the $B$ matrices are trained and communicated.
All other configurations (rank, scaling factor, target modules, dropout) follow the LoRA setup above.

\subsection{Iteration-Budget Experiment Setup}
\label{app:iter_budget}
Validation loss is evaluated every 5 iterations, and a run stops once it reaches the target loss.
Each run is allocated a maximum of 10,000 iterations, with early stopping if the validation loss does not improve for 1,000 consecutive iterations.
The learning rate is searched over a finer grid than in the previous experiments. Specifically, we consider seven values evenly spaced on a logarithmic scale from $10^{-2}$ to $10^{-4}$ for the LoRA baselines and from $10^{-4}$ to $10^{-6}$ for \methodname{}. All other experimental settings are identical to those in \Cref{sec:scaling}.

\subsection{Compute resource}
All experiments were conducted on GPU servers equipped with either NVIDIA A100-80GB GPUs
or NVIDIA RTX A6000 GPUs. To simulate decentralized training while keeping all client model
instances resident on GPUs, we used either two A100-80GB GPUs or four A6000 GPUs per run,
placing multiple client models on each GPU. For the main experiments in Sections~\ref{sec:main-results} and~\ref{subsec:large_network},
the compute cost per run ranged from approximately 8 GPU-hours on A6000 GPUs for lighter
LoRA-based first-order baselines to approximately 240 GPU-hours for the most expensive
full-parameter zeroth-order baselines. This range mainly reflects differences in model size,
full-parameter versus LoRA tuning, first-order versus zeroth-order iteration budgets, and the
cost of simulating multiple decentralized clients. It should therefore be interpreted as a
reproducibility-oriented resource estimate rather than as a controlled wall-clock efficiency
comparison between methods

%% file: sections/appendix-SupplementaryResults.tex
\section{Supplementary Task-Level Results}
\label{app:supresult}
Tables~\ref{tab:fozo_results},\ref{tab:multirun_scalability} provide the numerical values underlying the results presented in \Cref{sec:main-results} and \Cref{subsec:large_network}.

\begin{table*}[t]
\scriptsize
\renewcommand\arraystretch{0.8}
\setlength{\tabcolsep}{2.5pt}
\centering
\caption{Generalized performance comparison over a 16-client network among baseline decentralized methods and \methodname{}. SST2--ReCoRD are evaluated on OPT-1.3B and SQuAD/DROP on Qwen3-1.7B. FO and ZO denote first-order and zeroth-order optimization, and the -FFA suffix denotes FFA-LoRA fine-tuning. Cost is the total transmitted volume per edge over training, reported separately for each model. $\Delta$\,DSGD is the average relative performance difference across all tasks with respect to DSGD (FFT) under the same network topology, and Rel.\ ZS is the average performance relative to zero-shot. Best results among ZO methods are in bold.}
\label{tab:fozo_results}
\begin{tabularx}{0.995\textwidth}{c l *{8}{Y} c c r r}
\toprule
\multicolumn{14}{c}{\textbf{Ring Network}} \\
\midrule
& & \multicolumn{6}{c}{OPT-1.3B} & \multicolumn{2}{c}{Qwen3-1.7B} & \multicolumn{2}{c}{Cost} & & \\
\cmidrule(lr){3-8} \cmidrule(lr){9-10} \cmidrule(lr){11-12}
Type & Method & SST2 & RTE & BoolQ & WiC & MultiRC & ReCoRD & SQuAD & DROP & OPT & Qwen & \multicolumn{1}{c}{$\Delta$\,DSGD} & \multicolumn{1}{c}{Rel.\ ZS} \\
\midrule
\multirow{1}{*}{-}
& ZeroShot & 53.56 & 53.43 & 45.50 & 56.90 & 45.40 & 70.50 & 57.57 & 41.11 & -- & -- & -24.29\% & 100.00\% \\
\midrule
\multirow{6}{*}{\centering FO}
& DSGD       & 93.69 & 71.84 & 66.10 & 62.33 & 68.50 & 71.50 & 88.59 & 48.86 & 526.3GB & 681.4GB & 0.00\%  & 136.15\% \\
& DSGD-LoRA  & 93.58 & 70.76 & 65.70 & 61.44 & 68.80 & 71.20 & 86.93 & 49.06 & 629.1MB & 635.8MB & -0.64\% & 135.30\% \\
& DSGD-FFA   & 92.78 & 70.04 & 67.10 & 60.19 & 66.50 & 71.80 & 86.27 & 48.31 & 314.6MB & 272.5MB & -1.46\% & 134.16\% \\
& ChocoSGD   & 93.35 & 67.87 & 68.60 & 61.60 & 67.40 & 72.05 & 87.45 & 47.10 & 15.79GB & 17.03GB & -1.13\% & 134.68\% \\
& Choco-LoRA & 83.95 & 59.21 & 63.10 & 59.88 & 55.50 & 72.70 & 79.49 & 47.39 & 18.8MB  & 18.96MB & -8.38\% & 123.77\% \\
& Choco-FFA  & 91.74 & 60.29 & 65.80 & 58.31 & 60.30 & 72.10 & 84.79 & 45.39 & 9.4MB   & 8.12MB  & -5.95\% & 128.00\% \\
\midrule
\multirow{4}{*}{\centering ZO}
& DZSGD      & 91.40 & 61.01 & 64.40 & \textbf{59.56} & 59.70 & 70.50 & 86.80 & 47.11 & 5.26TB & 6.88TB & -5.55\% & 128.49\% \\
& DZSGD-LoRA & 91.17 & 54.51 & 64.40 & 57.99 & 59.95 & 71.00 & 83.18 & 45.22 & 6.29GB & 6.42GB & -7.89\% & 125.37\% \\
& DZSGD-FFA  & 90.71 & \textbf{65.70} & 65.10 & 59.40 & 62.10 & 70.20 & 82.29 & 44.21 & 3.14GB & 2.75GB & -5.72\% & 128.33\% \\
& \textbf{\methodname{}} & \textbf{92.78} & 63.18 & \textbf{67.20} & 57.99 & \textbf{63.70} & \textbf{71.90} & \textbf{87.09} & \textbf{48.63} & \textbf{400KB} & \textbf{400KB} & \textbf{-3.37\%} & \textbf{131.62\%} \\
\bottomrule
\vspace{2pt}
\end{tabularx}
\begin{tabularx}{0.995\textwidth}{c l *{8}{Y} c c r r}
\toprule
\multicolumn{14}{c}{\textbf{Meshgrid Network}} \\
\midrule
& & \multicolumn{6}{c}{OPT-1.3B} & \multicolumn{2}{c}{Qwen3-1.7B} & \multicolumn{2}{c}{Cost} & & \\
\cmidrule(lr){3-8} \cmidrule(lr){9-10} \cmidrule(lr){11-12}
Type & Method & SST2 & RTE & BoolQ & WiC & MultiRC & ReCoRD & SQuAD & DROP & OPT & Qwen & \multicolumn{1}{c}{$\Delta$\,DSGD} & \multicolumn{1}{c}{Rel.\ ZS} \\
\midrule
\multirow{1}{*}{-}
& ZeroShot & 53.56 & 53.43 & 45.50 & 56.90 & 45.40 & 70.50 & 57.57 & 41.11 & -- & -- & -25.21\% & 100.00\% \\
\midrule
\multirow{6}{*}{\centering FO}
& DSGD       & 94.15 & 73.64 & 69.80 & 62.54 & 69.80 & 70.70 & 88.64 & 49.53 & 526.3GB & 681.4GB & 0.00\%  & 138.18\% \\
& DSGD-LoRA  & 93.92 & 71.84 & 67.10 & 61.60 & 68.00 & 71.10 & 87.74 & 50.30 & 629.1MB & 635.8MB & -1.19\% & 136.37\% \\
& DSGD-FFA   & 93.12 & 71.12 & 71.00 & 57.37 & 67.50 & 71.70 & 86.11 & 48.42 & 314.6MB & 272.5MB & -2.25\% & 135.20\% \\
& ChocoSGD   & 92.09 & 70.76 & 68.00 & 61.76 & 68.10 & 71.60 & 88.30 & 47.92 & 15.79GB & 17.03GB & -1.84\% & 135.48\% \\
& Choco-LoRA & 87.62 & 59.93 & 63.10 & 57.84 & 55.40 & 72.90 & 83.09 & 45.04 & 18.8MB  & 18.96MB & -9.44\% & 124.42\% \\
& Choco-FFA  & 91.74 & 61.73 & 65.30 & 60.03 & 57.00 & 72.40 & 84.83 & 45.67 & 9.4MB   & 8.12MB  & -7.15\% & 127.82\% \\
\midrule
\multirow{4}{*}{\centering ZO}
& DZSGD      & 91.28 & \textbf{66.07} & 65.00 & \textbf{59.25} & 61.40 & 71.60 & \textbf{87.19} & 45.94 & 5.26TB & 6.88TB & -5.64\% & 130.13\% \\
& DZSGD-LoRA & 90.14 & 54.87 & 64.70 & 58.46 & 56.30 & 71.20 & 84.53 & 44.53 & 6.29GB & 6.42GB & -9.62\% & 124.51\% \\
& DZSGD-FFA  & 91.17 & 64.98 & 63.30 & 58.93 & \textbf{64.20} & 70.00 & 83.11 & 44.70 & 3.14GB & 2.75GB & -6.88\% & 128.54\% \\
& \textbf{\methodname{}} & \textbf{92.89} & 63.90 & \textbf{66.70} & 57.99 & 61.90 & \textbf{71.90} & 86.07 & \textbf{48.36} & \textbf{400KB} & \textbf{400KB} & \textbf{-5.15\%} & \textbf{130.88\%} \\
\bottomrule
\end{tabularx}
\end{table*}

\begin{table}[ht]
    \footnotesize
    \centering
    \caption{Three-run averaged results for the ring-network scalability experiment in \Cref{sec:scaling}. SST2, RTE, and BoolQ use OPT-125m; SQuAD uses Qwen3-0.6B. Average relative performance is computed with respect to the corresponding reference performance used in the main paper.}
    \label{tab:multirun_scalability}
    \begin{tabular}{llrrrr}
        \toprule
        Task / Metric & Method & 16 & 32 & 64 & 128 \\
        \midrule
        SST2 & DSGD-LoRA   & 86.07 & 85.35 & 73.47 & 55.19 \\
             & DSGD-FFA    & 86.58 & 85.55 & 85.32 & 84.44 \\
             & Choco-LoRA  & 69.53 & 55.04 & 55.16 & 53.59 \\
             & Choco-FFA   & 85.21 & 84.63 & 83.68 & 82.38 \\
             & \methodname{} & 84.57 & 85.20 & 84.93 & 84.82 \\
        \midrule
        RTE  & DSGD-LoRA   & 58.36 & 54.99 & 51.96 & 53.67 \\
             & DSGD-FFA    & 59.21 & 56.32 & 54.63 & 54.39 \\
             & Choco-LoRA  & 53.91 & 53.67 & 53.79 & 53.55 \\
             & Choco-FFA   & 55.72 & 55.96 & 53.91 & 53.31 \\
             & \methodname{} & 54.15 & 55.11 & 56.91 & 60.04 \\
        \midrule
        BoolQ & DSGD-LoRA   & 62.77 & 60.56 & 50.37 & 44.93 \\
              & DSGD-FFA    & 62.50 & 61.97 & 60.73 & 59.40 \\
              & Choco-LoRA  & 59.63 & 52.96 & 52.20 & 47.77 \\
              & Choco-FFA   & 61.07 & 59.63 & 58.60 & 57.77 \\
              & \methodname{} & 60.70 & 61.80 & 60.80 & 62.56 \\
        \midrule
        SQuAD & DSGD-LoRA   & 83.42 & 78.70 & 72.87 & 63.65 \\
              & DSGD-FFA    & 83.66 & 82.45 & 80.75 & 79.82 \\
              & Choco-LoRA  & 74.82 & 73.82 & 63.05 & 59.26 \\
              & Choco-FFA   & 81.95 & 79.98 & 78.45 & 77.12 \\
              & \methodname{} & 82.92 & 82.24 & 84.41 & 84.19 \\
        \bottomrule
    \end{tabular}
\end{table}